\documentclass[lettersize,journal]{IEEEtran}

\usepackage[T1]{fontenc}    
\usepackage[citecolor=blue, colorlinks, pagebackref=true]{hyperref}       
\usepackage{url}            
\usepackage{booktabs}       
\usepackage{amsfonts}       
\usepackage{nicefrac}       
\usepackage{microtype}      
\usepackage{xcolor}         
\usepackage{colortbl}       
\usepackage{graphicx}
\usepackage{subcaption}
\usepackage{enumitem}
\usepackage{multirow}
\usepackage{amsmath}
\usepackage{amssymb}
\usepackage{wrapfig}
\usepackage{stfloats}
\fnbelowfloat

\newif\ifshowchanges
\showchangestrue

\definecolor{rankgold}{HTML}{FFE699}
\definecolor{ranksilver}{HTML}{D9E2F3}
\definecolor{rankbronze}{HTML}{F3E7DC}
\newcommand{\rankfirst}[1]{\cellcolor{rankgold}\textbf{#1}}
\newcommand{\ranksecond}[1]{\cellcolor{ranksilver}\underline{#1}}
\newcommand{\rankthird}[1]{\cellcolor{rankbronze}#1}
\begin{document}

\title{Beyond UV Mapping: Mesh Texture Compression via Surface-Aligned Texture Fields}

\author{Jianqiang Wang,  Junhui Hou~\IEEEmembership{Senior Member,~IEEE}, Siyu Ren, Weiyao Lin, Wenping Wang~\IEEEmembership{Fellow,~IEEE}
        \thanks{This work was supported in part by the Natural Science Foundation of China under Grant 62422118, and in part by the Hong Kong Research Grants Council under Grants 11220426, 11219324, and N\_CityU1114/25.}
\thanks{J. Wang, J. Hou, and S. Ren are with the Department of Computer Science, City University of Hong Kong, Hong Kong SAR, China (Email: wang.jq@cityu.edu.hk; jh.hou@cityu.edu.hk; siyuren2-c@my.cityu.edu.hk)}
\thanks{W. Lin is with the Department of Electrical Engineering, Shanghai Jiao Tong University, Shanghai, China (e-mail: wylin@sjtu.edu.cn)}
\thanks{W. Wang is with the Department of Computer Science and Engineering, Texas A\&M University, TX 77840, US (Email: wenping@tamu.edu).}
}

\markboth{}{Wang \MakeLowercase{\textit{et al.}}: Beyond UV Mapping: Mesh Texture Compression via Surface-Aligned Texture Fields}

\maketitle

\begin{abstract}
Mesh texture compression typically relies on 2D UV atlases, whose chart discontinuities and mapping overhead can limit coding efficiency. To tackle this challenge, we introduce TexF, a surface-aligned texture field that organizes texture attributes in sparse voxels derived from the mesh surface. This representation supports high-resolution textures while preserving local 3D correlations for compression and enabling direct surface queries. 
For bitstream compression, TexF reuses established 3D attribute codecs, with voxel locations reconstructed from the decoded mesh without separate transmission. For GPU-resident compression, we develop 3DNTC, which combines quantized hash features with a lightweight decoder for random-access reconstruction at surface positions. 
Differentiable rendering enables image-space refinement of both voxel attributes and compressed neural fields. 
Experiments on the MPEG and AOM mesh compression benchmarks demonstrate improved average rate–distortion performance over representative UV-based methods for both bitstream and GPU-resident compression. 3DNTC also supports real-time rendering.
The code will be publicly available at \url{https://github.com/yydlmzyz1/TexF}.
\end{abstract}


\section{Introduction}

Textured 3D meshes are widely used to represent 3D assets, enabling the depiction of continuous geometric surfaces with high-fidelity visual appearances. With the rapid expansion of immersive media \cite{tutorialimmersive2023} and spatial intelligence \cite{Yang_2025_CVPR}, the efficient representation and compression of these assets have become essential \cite{vdmc21cfp,AOM_VVM_CfP_2023}. 
While the geometric surface provides the structural foundation, the texture component is a key factor in visual realism and often constitutes a major portion of the overall bitrate. 
Texture compression is challenging not only because of its data volume, but also because texture must remain aligned with the underlying surface and preserve rendered visual quality.

\begin{figure}[t]
    \centering
    \includegraphics[width=0.48\textwidth]{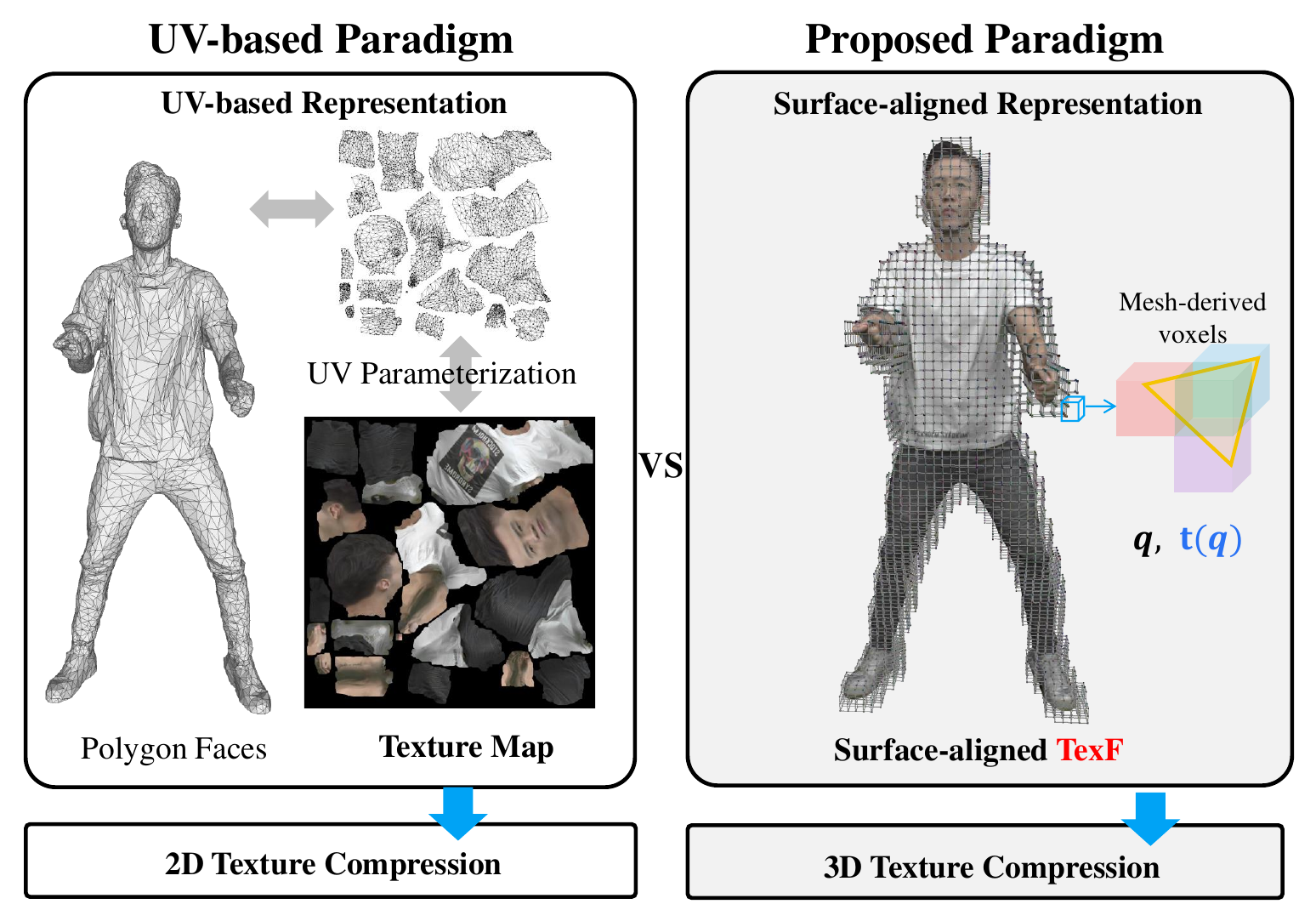} \vspace{-0.2cm}
    \caption{\textbf{UV-based and TexF mesh texture compression paradigms.}
    UV-based methods represent texture in 2D atlases for compression with established 2D codecs.
    In contrast, TexF represents texture directly in a surface-aligned 3D field for compression and rendering.
    }
\label{fig:overview}
\end{figure}

Existing mesh compression pipelines typically store geometry as polygonal meshes and appearance in separate 2D texture atlases through UV parameterization~\cite{maglo20153d,draco,vdmc25whitepaper}, as illustrated on the left of Fig.~\ref{fig:overview}.
Within this paradigm, mature image/video codecs such as HEVC~\cite{HEVC,zou25VDMC} and AV1~\cite{han2021av1,AOM_VVM_CfP_2023} produce compact bitstreams for storage and transmission.
Hardware-supported texture formats such as ASTC~\cite{nystad2012adaptive} and neural texture compression (NTC) methods~\cite{ntc2023} reduce GPU-resident memory while supporting random access during rendering.
Despite their different coding mechanisms, these methods share the same 2D UV atlas representation.
However, UV atlas construction can separate nearby surface regions into disjoint charts, weakening the spatial correlations exploitable by 2D codecs, while also introducing padding, packing, and UV-mapping overhead. Both effects can reduce coding efficiency, particularly for meshes with complex geometry.
These limitations motivate alternative ways to organize texture for compression.

Beyond compression, texture representation has long been explored outside the UV domain, primarily for appearance modeling, authoring, and rendering. UV-free methods associate texture with mesh elements or locations in 3D space.
Related representations include mesh-attached textures~\cite{burley2008ptex,yuksel2010meshcolors}, volumetric textures~\cite{peachey1985solid,benson2002octree,porumbescu2005shell}, and neural or implicit textures~\cite{oechsle2019texture,xiang2021neutex}.
Some of these efforts also address texture storage through sparse octrees~\cite{benson2002octree,lacoste2007appearance}, compressed 3D blocks~\cite{bajaj2000compression}, or voxel serialization~\cite{dolonius2020uvfree}.
Yet these storage reductions do not necessarily translate into better compression than established UV-based pipelines. Inefficient modeling of spatial correlations and additional indexing overhead can offset the savings from removing UV mappings, and even increase storage.
These considerations suggest that effective mesh texture compression requires jointly addressing texture organization and mapping overhead. We therefore investigate how mesh surface geometry can guide texture organization and coding in 3D, beyond simply removing UV mappings.

Our central idea is to use the mesh surface itself to determine where texture attributes are stored and how they are accessed.
We introduce \textbf{TexF}, a \emph{surface-aligned texture field} that stores texture attributes only in voxels intersected by the mesh surface (right of Fig.~\ref{fig:overview}).
This surface-aligned organization offers three advantages for compact representation and compression.
First, surface-localized sparsity supports high-resolution, high-fidelity textures without a dense volume, using sample counts comparable to those of 2D texture maps.
Second, texture locations can be derived from the mesh, reducing the storage overhead of separate surface-to-texture mappings and spatial indices.
Third, organizing texture attributes directly in 3D preserves their native spatial relationships for compression, avoiding the disruption of spatial coherence caused by UV charting or reordering.

We construct TexF through direct texture transfer from a UV-textured mesh.
We develop an efficient, differentiable renderer that queries texture attributes directly at mesh surface positions.
The renderer also enables inverse-rendering (IR) refinement~\cite{munkberg2022extracting,kerbl20233d,liu2024text,richardson2023texture}, using multi-view image-space supervision to reduce transfer and discretization errors.

We exploit TexF's surface-aligned structure for both bitstream and GPU-resident texture compression.
For bitstream compression, we encode voxel attributes directly using established 3D attribute codecs, including Unicorn~\cite{unicorn2025attr} and G-PCC~\cite{MPEG-PCC-TMC13}, exploiting spatial correlations in their native 3D neighborhoods.
Voxel locations are reconstructed from the decoded mesh, avoiding separate voxel-position coding and UV-mapping overhead.
For GPU-resident compression, we develop a neural codec that captures 3D spatial correlations in quantized multiresolution hash features.
A compact decoder reconstructs attributes on demand at surface positions, supporting random access during rendering~\cite{muller2022instant,ntc2023}.

Experiments on the standard MPEG~\cite{vdmc24ctc} and AOM~\cite{AOM_VVM_CfP_2023} benchmarks demonstrate that TexF with Unicorn significantly outperforms the evaluated UV-based pipelines in averaged rate--distortion performance, including those enhanced by UV reparameterization and alternative image codecs.
Competitive results with G-PCC further show that these benefits are not confined to a single coding backend.
For GPU-resident compression, TexF generally achieves better rate--distortion performance than conventional and neural UV texture compression, while supporting random-access decoding during rendering.

In summary, we present a mesh texture compression framework based on a surface-aligned texture field (TexF). Specifically:
\begin{itemize}[noitemsep,nolistsep,leftmargin=*]
\item TexF stores and accesses texture in a sparse 3D field aligned with the underlying mesh surface, enabling compact, high-resolution, and UV-free representation. An efficient differentiable renderer supports direct surface queries and inverse-rendering refinement.
\item We exploit TexF's native 3D spatial correlations for both bitstream and GPU-resident compression.
Bitstream compression reuses established 3D attribute codecs without separately transmitting voxel locations.
GPU-resident compression uses quantized hash features and a lightweight decoder for random-access surface queries without explicit per-voxel lookup entries.
\item We systematically compare TexF with UV-based and serialization-based alternatives, demonstrating improved average rate--distortion performance in both compression settings and analyzing how asset characteristics affect the relative performance of TexF and UV representations.
\end{itemize}

The rest of this paper is organized as follows. Sec. \ref{sec:RW} comprehensively reviews existing work on the representation of mesh texture and its compression. Sec. \ref{sec:our method} presents the proposed method, followed by extensive experiments and a series of promising future directions in Sec. \ref{sec:exp}. Finally, Sec. \ref{sec:con and dis} concludes this work

\begin{figure*}[t]
\centering
\includegraphics[width=0.98\linewidth]{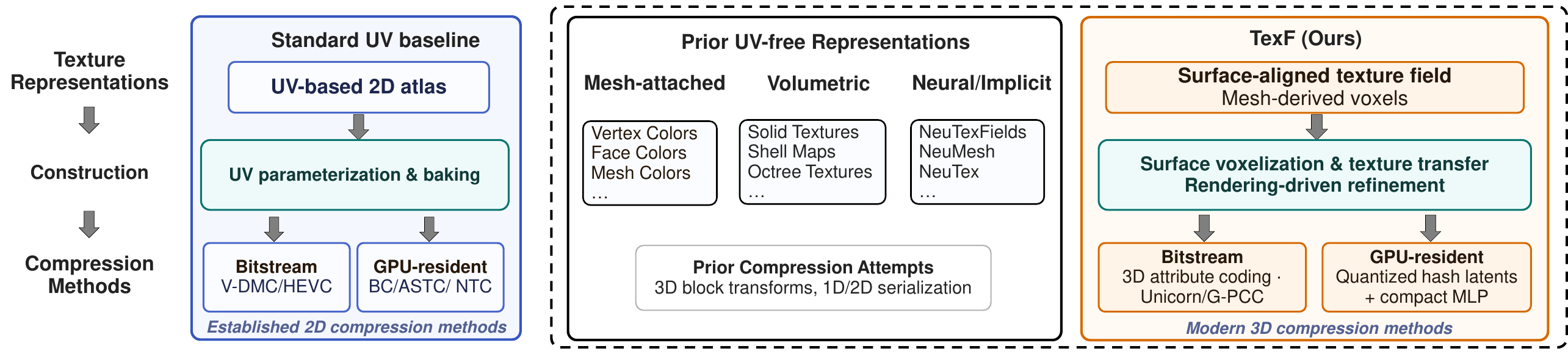}
\caption{\textbf{Mesh texture representations, construction, and compression.}
UV atlases are constructed through parameterization and baking and support established 2D codecs, whereas UV-free methods associate appearance with mesh elements or object-space locations.
Prior UV-free compression methods rely on regular 3D blocks or linearized 3D attributes.TexF uses a surface-aligned texture field as the common representation for bitstream and GPU-resident compression.}
\label{fig:related_taxonomy}
\end{figure*}

\section{Related Work}
\label{sec:RW}
Compression of a textured mesh involves both geometry and texture. Geometry coding represents vertex positions and connectivity using methods such as Draco~\cite{draco}; comprehensive reviews are provided in~\cite{maglo20153d,peng2005technologies}.
Our focus, instead, is mesh texture compression.

Since the texture representation determines the signal exposed to subsequent coding, we organize prior work according to how texture is associated with the mesh surface, broadly distinguishing UV-based and UV-free approaches.
UV atlases remain the prevailing representation in production pipelines, underlying both texture bitstreams for storage and transmission and GPU-resident payloads for rendering.
Fig.~\ref{fig:related_taxonomy} summarizes the representations, their construction, and the associated compression methods.
Here,  it is worth noting that \emph{UV-free} describes how texture is associated with the mesh surface, rather than the domain in which the resulting signal is compressed.

\subsection{UV-based Mesh Textures}
\noindent\textbf{UV parameterization and atlas construction.}
A UV-textured mesh associates its surface with coordinates in a 2D texture domain. Surface parameterization partitions the mesh into charts and unfolds them onto the plane. Parameterization methods seek to limit angular or area distortion~\cite{floater2005surface,zhou2004iso}, while atlas packing arranges the resulting charts in a UV atlas with gutters to improve image utilization and reduce filtering artifacts~\cite{UVAtlas}.
When constructing a new atlas, texture baking resamples the source appearance into the new UV layout.
The resulting regular 2D images are compatible with mature image/video coding tools and GPU texture formats.

\vspace{0.5em}
\noindent\textbf{Bitstream-oriented compression.}
For storage and transmission, UV-based texture can be encoded using established codecs such as H.265/HEVC~\cite{HEVC} and AV1~\cite{han2021av1}.
MPEG V-DMC~\cite{vdmc25whitepaper,zou25VDMC} organizes mesh textures into video-compatible representations and reuses established video coding tools. Reparameterization~\cite{UVAtlas} and atlas packing can improve texture coding efficiency, but may incur substantial preprocessing cost for complex meshes.

\vspace{0.5em}
\noindent\textbf{GPU-resident compression.}
For runtime GPU memory, block-based image formats derived from block truncation coding~\cite{delp2003image}, including S3TC~\cite{iourcha1999s3tc} and ASTC~\cite{nystad2012adaptive}, provide fixed-rate storage and hardware-supported random access.
Random-access neural texture compression (NTC)~\cite{ntc2023} instead represents material using quantized latent grids and a compact decoder, enabling on-demand reconstruction from a GPU-resident representation at the cost of per-material optimization.

\vspace{0.5em}
\noindent\textbf{Limitations of the UV domain.}
Despite the practical advantages of UV-based pipelines, chart-based unfolding changes the native surface neighborhoods exposed in the atlas: regions adjacent on the mesh may fall into different charts, weakening correlations across chart boundaries.
Chart packing also introduces padding, while the explicit surface-to-atlas mapping requires UV coordinates and associated indices.
Constructing a new atlas also incurs preprocessing cost, and texture resampling can introduce distortion before compression.
Careful parameterization and packing mitigate these costs but do not remove the dependence on chart layout, motivating complementary representations that preserve appearance directly in 3D space.

\subsection{UV-free Mesh Textures}
UV-free texture representations associate appearance directly with mesh elements or locations in the surrounding 3D space, avoiding dependence on a global UV atlas.
This long-standing idea has inspired diverse approaches to appearance modeling, authoring, and rendering, whereas its application to mesh texture compression remains comparatively limited.

\subsubsection{UV-free Texture Representations}
UV-free texture representations differ primarily in how they associate appearance with the mesh surface.

\vspace{0.5em}
\noindent\textbf{Mesh-attached textures.}
A separate family attaches appearance directly to mesh elements. Per-vertex colors provide the simplest example, although their spatial resolution is limited by the vertex density of the mesh.
Per-face methods such as Ptex~\cite{burley2008ptex} assign an independent 2D texture to each face and maintain a per-face adjacency map for cross-face filtering.
Mesh Colors~\cite{yuksel2010meshcolors} and Patch Textures~\cite{mallett2020patch} associate color samples with mesh elements without explicit UV coordinates.
However, their global organization follows mesh connectivity rather than a single regular lattice, complicating the direct reuse of mature image- or volume-based codecs. These systems primarily address parameterization-free authoring and rendering rather than texture compression.

\vspace{0.5em}
\noindent\textbf{Volumetric textures.}
Volumetric texture representations associate appearance directly with positions in 3D space.
Early solid textures~\cite{peachey1985solid,perlin1985image} define appearance procedurally as a function of 3D position and are particularly suited to materials with volumetric patterns, such as wood or stone.
Shell Maps~\cite{porumbescu2005shell} instead construct a surface-aligned volumetric parameterization by mapping a tetrahedral shell around the mesh into texture space.
Octree-based textures~\cite{benson2002octree,debry2002painting,lacoste2007appearance} instead organize sampled surface attributes in adaptive octree hierarchies for parameterization-free painting and rendering.
TexF instead defines a surface-aligned texture field directly in the mesh coordinate system, with voxel locations derived from the mesh rather than stored in a separate hierarchy.

\vspace{0.5em}
\noindent\textbf{Neural and implicit textures.}
In recent years, neural methods have represented appearance using continuous coordinate functions or learned features.
Oechsle et al.'s neural Texture Fields~\cite{oechsle2019texture} represent appearance as a continuous function $f_\theta(x,y,z)\rightarrow\mathrm{RGB}$ conditioned on shape features, avoiding explicit surface parameterization.
NeuMesh~\cite{yang2022neumesh} instead attaches learned geometry and texture features to mesh vertices for neural rendering and editing.
In contrast, NeuTex~\cite{xiang2021neutex} learns a continuous mapping from 3D surface points to a 2D neural texture domain.
These neural texture methods primarily target appearance modeling, reconstruction, or editing. TexF instead starts from explicit texture attributes on mesh-derived voxels, providing a common representation for both 3D attribute codecs and GPU-resident neural compression.

\begin{figure*}[t]
\centering
\includegraphics[width=0.98\linewidth]{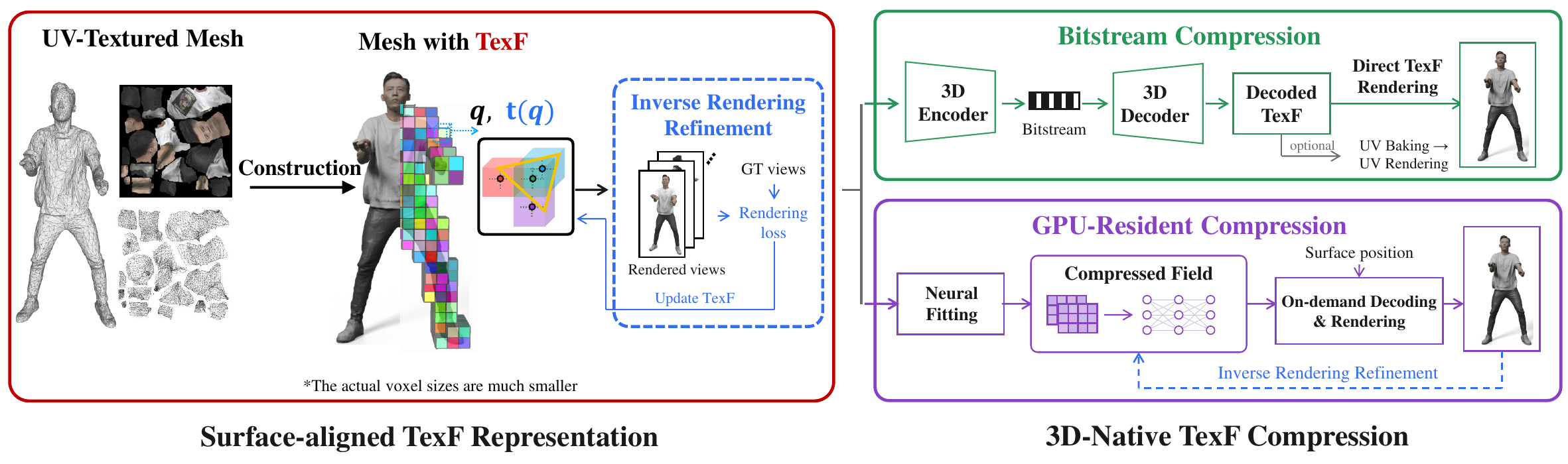}
\caption{\textbf{Illustration of \textit{Texture Field} (TexF) representation and compression.} In contrast to 2D UV mapping, TexF embeds texture attributes $\mathbf{t}(\mathbf{q})$ directly in a surface-aligned 3D field, and queries them at mesh surface positions during rendering.
It supports bitstream compression with established 3D attribute codecs for storage and transmission and GPU-resident neural compression for random-access rendering.
Inverse Rendering optimization improves the fidelity of both explicit TexF attributes and compressed neural fields.}
\label{fig:pipeline}
\end{figure*}

\subsubsection{UV-free Representations for Mesh Texture Compression}

UV-free representations have been widely studied for appearance modeling, authoring, and rendering, but comparatively few works address mesh texture compression. These works differ in how they organize surface attributes for coding.

As an early effort, CB3TM~\cite{bajaj2000compression} retains surface-relevant regions of a solid texture and applies 3D Haar-wavelet coding with coefficient truncation to local $4^3$ subvolumes for independent reconstruction. Although not mesh-specific, the 3D mode of ASTC~\cite{nystad2012adaptive} provides fixed-rate coding of regular 3D blocks. Both exploit local 3D correlations and support random access, but their dense coding units limit the use of fine-grained surface sparsity.

Dolonius et al.~\cite{dolonius2020uvfree,dolonius2019compressing} used octree-derived Sparse Voxel Directed Acyclic Graphs (SVDAGs) to compactly represent voxelized surfaces and address separately stored attributes. The attributes are linearized along Morton or Hilbert curves, then compressed with a custom random-access codec or mapped to 2D for existing image and GPU texture codecs. These methods exploit spatial coherence through the serialized ordering. 
While SVDAG-based methods use a separate voxel hierarchy for texture access, TexF uses the mesh surface as the common spatial basis for representation, compression, and rendering. This surface-aligned organization supports native 3D attribute coding and direct neural queries at surface positions.

\subsection{Related 3D Coding and Optimization Techniques}
Beyond mesh-texture representations, research on 3D coding and differentiable rendering provides tools that TexF draws on for construction and compression.

\vspace{0.5em}
\noindent\textbf{3D spatial coding.}
Broader volumetric-data research has developed compression methods for regular and sparse 3D signals~\cite{nystad2012adaptive,gobbetti2012covra,dado2016geometry,kim2024neuralvdb}. These techniques generally target volume data or voxel scenes rather than appearance associated with a separate triangle mesh.
Point cloud codecs provide related tools for attributes defined on known 3D support, but do not themselves define a mesh-texture representation or renderer. V-PCC projects point cloud patches into packed 2D images, whereas G-PCC operates on geometry and attributes in the 3D point set~\cite{schwarz2018emerging,MPEG-PCC-TMC13}. Learned point cloud attribute codecs, such as Unicorn~\cite{unicorn2025attr}, further explore data-driven spatial transforms and entropy models~\cite{DeepRAHT,huo2025rendering}. TexF makes these codecs applicable to mesh textures by organizing attributes on mesh-derived voxels, whose positions can be reconstructed from the decoded mesh without separate transmission.

\vspace{0.5em}
\noindent\textbf{Multiresolution hash encoding.}
Multiresolution hash encoding~\cite{muller2022instant} maps spatial coordinates to trainable features stored in compact hash tables.
We adapt this encoding to GPU-resident TexF compression, using quantized features and lightweight decoding inspired by NTC~\cite{ntc2023}.
By replacing explicit voxel attributes with a directly queryable compressed field, the codec avoids storing per-voxel lookup entries.

\vspace{0.5em}
\noindent\textbf{Inverse rendering.}
Inverse rendering optimizes geometry or appearance by propagating image-space supervision through a differentiable rendering process~\cite{munkberg2022extracting,kerbl20233d,liu2024text,richardson2023texture}. TexF uses this supervision both to reduce transfer and discretization errors during representation construction and to refine the compressed neural field.

\section{{Proposed Method}}
\label{sec:our method}

Conventional mesh texture compression operates on 2D UV atlases. We instead organize texture attributes in a \emph{surface-aligned texture field} (\textbf{TexF}), whose sparse voxel support is derived from the mesh surface. Thus surface positions can directly query the 3D field, providing a common spatial basis for texture representation, compression, and rendering.

As illustrated in Fig.~\ref{fig:pipeline}, we construct TexF by voxelizing the mesh surface and transferring the source texture, then refine its attributes through differentiable rendering to reduce transfer and discretization errors. Based on this representation, we develop two complementary schemes: bitstream compression with established 3D attribute codecs for disk storage and transmission, and GPU-resident neural compression for memory-efficient random access during rendering.

\subsection{Surface-aligned TexF Representation and Construction}

TexF associates texture attributes with voxels derived from the mesh surface, preserving spatial locality without UV parameterization. We efficiently construct this support through surface voxelization and initialize its attributes from the source texture.

\subsubsection{Surface-aligned TexF Construction}

Given a textured mesh $\mathbf{M}$, we first discretize its canonical axis-aligned bounding box into a $K\times K\times K$ grid. TexF and the mesh geometry therefore share a common coordinate system, allowing surface points to query the field directly in 3D space without an intermediate parameterization such as UV mapping.
The surface-aligned TexF retains only voxels intersecting the mesh surface, referred to as active voxels. Their indices form the active set $\mathbf{Q}$, with $\mathbf{q}\in\mathbb{Z}^3$ and $0\le\mathbf{q}<K$.
In practice, we construct this sparse support efficiently through surface sampling and voxelization without allocating the dense $K^3$ grid.
This surface-localized sparsity supports high-resolution texture representation with sample counts comparable to those of 2D texture maps, avoiding the high memory cost of dense volumetric textures.
Each voxel $\mathbf{q}$ in the active set $\mathbf{Q}$ stores texture attributes $\mathbf{t}(\mathbf{q})$, defining TexF as
\begin{equation}
\mathcal{T}=\{(\mathbf{q},\mathbf{t}(\mathbf{q})) \mid \mathbf{q}\in\mathbf{Q}\},
\label{eq:representation}
\end{equation}
where $\mathbf{t}(\mathbf{q})$ encodes base color (i.e., RGB) by default and can be extended channel-wise to include other physically-based rendering (PBR) material attributes, such as roughness and metallicity.

A textured mesh is thus represented by its geometry $\mathbf{M}$ and the associated TexF $\mathcal{T}$, forming a mesh with TexF.
Surface positions query texture directly in their shared 3D coordinate system, without UV coordinates or a separate surface parameterization. 
Because the voxel support is derived from the mesh and grid configuration, its coordinates need not be separately transmitted. This organization preserves local 3D spatial relationships among texture attributes, supporting direct surface queries and spatially coherent attribute coding, as described in Secs.~\ref{sec:rep_render} and~\ref{sec:compression}.

\vspace{0.5em}
\noindent\textbf{Remark.}
TexF's spatial structure also offers representational versatility.
It naturally shares a voxel grid with grid-based implicit geometry,
such as an SDF, for joint geometry--appearance modeling.
Its attributes can also be sampled at mesh vertices for vertex-color rendering.
Examples are provided in the \textit{Supplementary Material}.

\begin{figure}[t]
\centering
\includegraphics[width=0.98\linewidth]{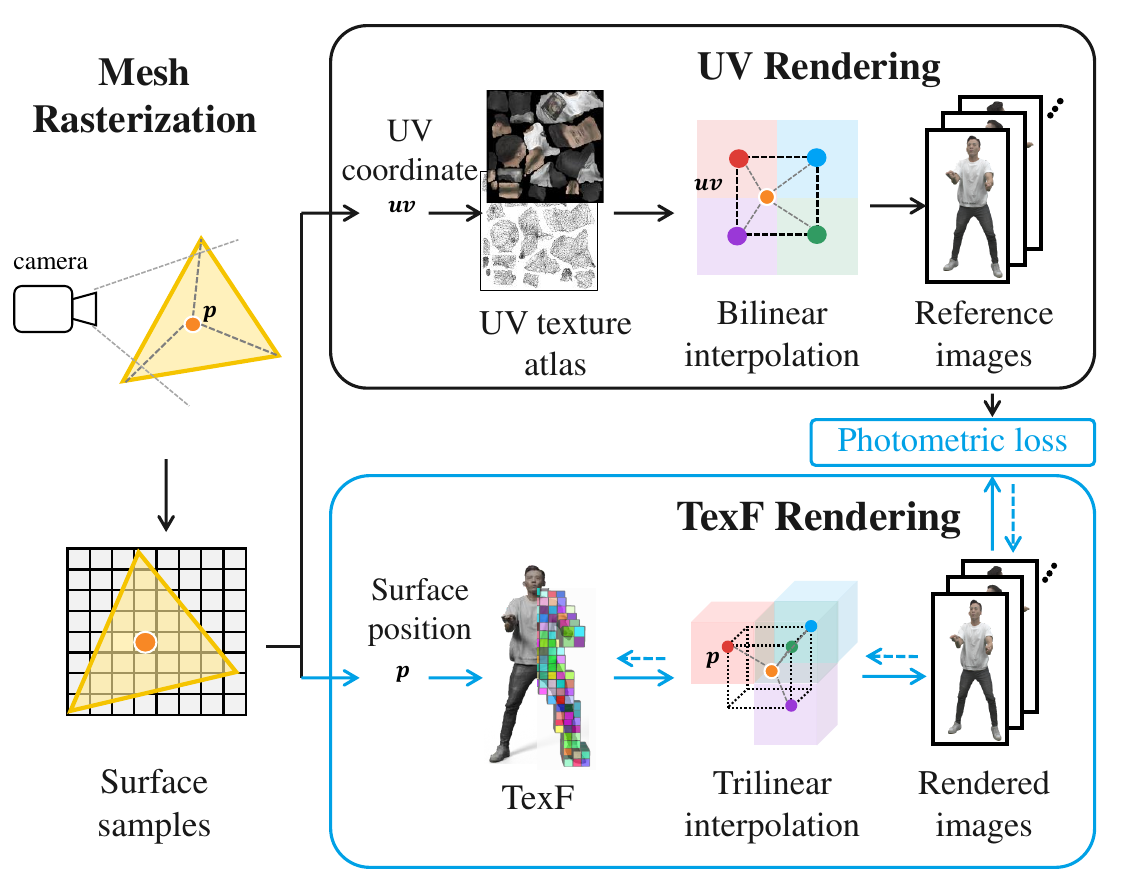}
\caption{\textbf{TexF rendering and refinement.}
UV rendering samples a 2D texture using interpolated UV coordinates, whereas TexF queries texture directly at rasterized surface positions by trilinearly interpolating neighboring voxel attributes.
A photometric loss between UV reference images and TexF renderings guides voxel attribute refinement.
Rendering from compressed features is shown in Fig.~\ref{fig:compression_memory}.}
\label{fig:rendering}
\vspace{-0.4cm}
\end{figure}

\subsubsection{UV Texture Transfer to TexF}
To support conventional UV-textured assets, we initialize TexF by sampling the mesh surface and transferring attributes from the source texture.
For each surface sample, its barycentric coordinates determine both its 3D position and UV coordinate. The UV coordinate is used to sample the source material maps, while the 3D position assigns the retrieved attributes to the corresponding voxel. Attributes assigned to the same voxel are averaged.
Conversely, TexF can be baked into a UV texture by rasterizing a target UV layout and querying the field at the corresponding surface positions, enabling compatibility with standard UV rendering pipelines.
Voxel discretization and interpolation introduce transfer errors, which we reduce through rendering-driven refinement in Sec.~\ref{sec:rep_render}.

\begin{figure}[t]
\centering
\includegraphics[width=\linewidth]{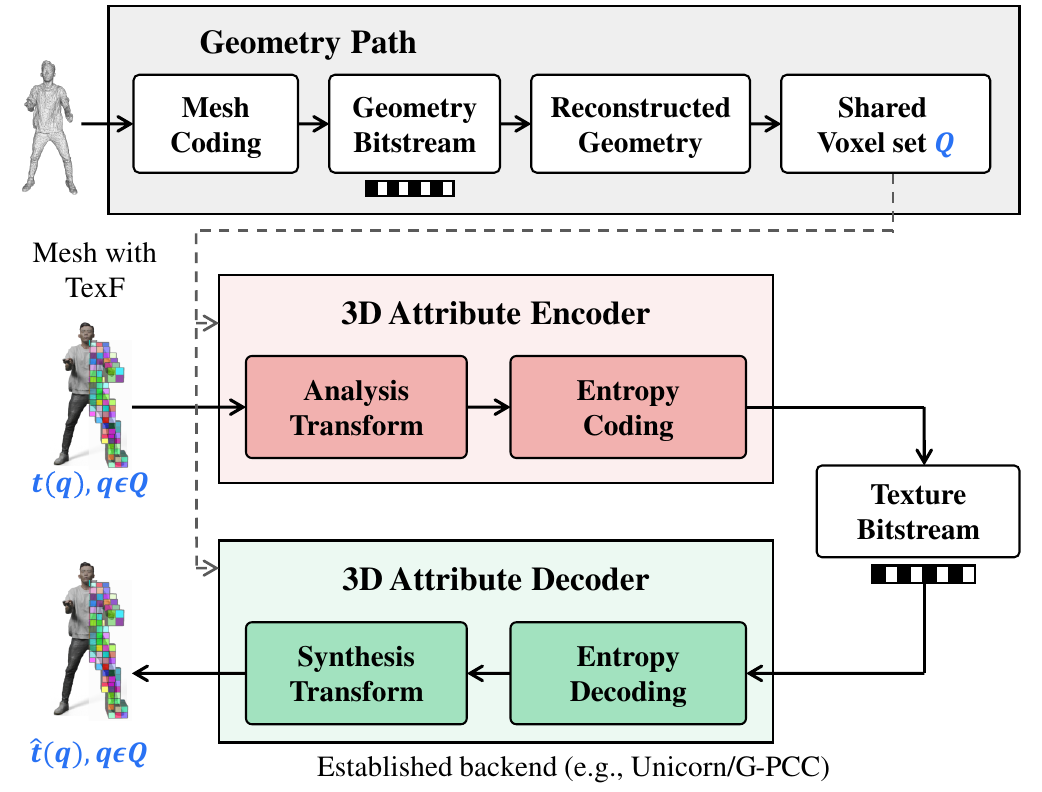}
\caption{\textbf{Bitstream-oriented TexF compression.} The encoder and decoder derive the same voxel set $\mathbf{Q}$ from the reconstructed mesh. An established 3D attribute codec compresses the corresponding TexF attributes without separately transmitting voxel coordinates.
Here, \(\mathbf{t}(\mathbf{q})\) and \(\hat{\mathbf{t}}(\mathbf{q})\) denote the input and reconstructed texture attributes at voxel \(\mathbf{q}\), respectively.
}
\label{fig:compression_bitstream}
\label{fig:compression}
\end{figure}

\begin{figure*}[t]
\centering
\includegraphics[width=\linewidth]{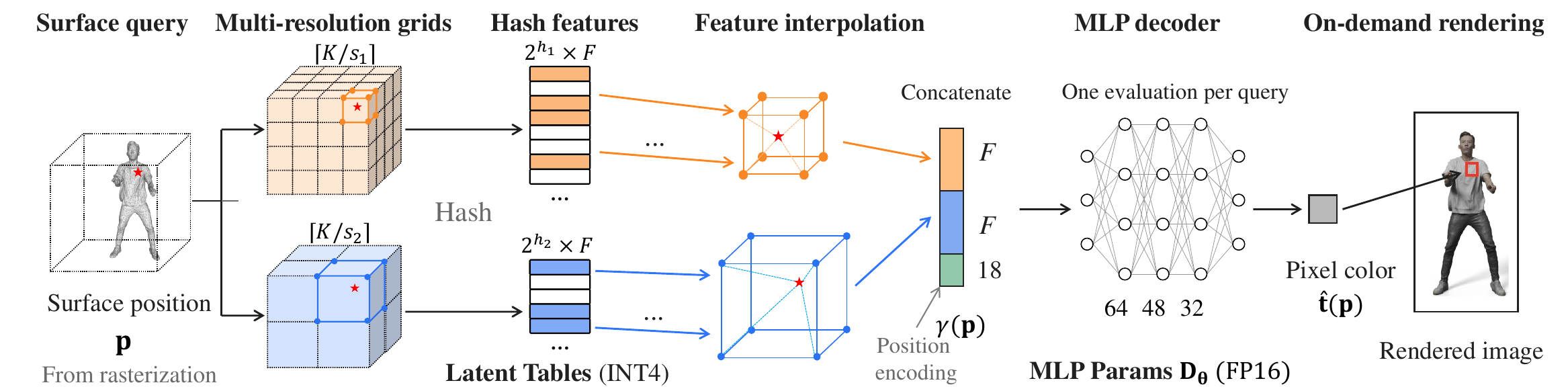}
\caption{\textbf{GPU-resident TexF decoding and rendering.}
At each rasterized surface position, quantized hash features are trilinearly interpolated and combined with positional encoding.
A single MLP evaluation then reconstructs the pixel color for on-demand rendering.}
\label{fig:compression_memory}
\end{figure*}

\subsection{TexF Rendering and Refinement}\label{sec:rep_render}

\subsubsection{Differentiable TexF Rendering}

TexF rendering first rasterizes the mesh to determine visible surface positions, then queries the field at these positions to obtain texture attributes.
As illustrated in Fig.~\ref{fig:rendering}, UV rendering samples a 2D texture through interpolated UV coordinates, whereas TexF queries the field directly at $\mathbf{p}$.
For the voxel representation defined in Eq.~\eqref{eq:representation}, the color $\mathbf{c}(\mathbf{p})$ is obtained by trilinearly interpolating neighboring texture attributes:
\vspace{-0.1cm}
\begin{equation}
\mathbf{c}(\mathbf{p}) = \sum_{i \in \mathcal{N}(\mathbf{p})} \omega_i \mathbf{t}(\mathbf{q}_i),\label{eq:grid_query} \vspace{-0.15cm}
\end{equation}
where $\mathcal{N}(\mathbf{p})$ denotes the set of active grid points within the $2\times2\times2$ neighborhood of $\mathbf{p}$, and the trilinear weights $\omega_i$ are determined by the query position within the grid cell. This interpolation is mathematically analogous to bilinear sampling in UV rendering, but operates in 3D with weights renormalized over the available voxels.
In the rare case that no active corner is available, the nearest active voxel is used as a fallback.
We implement this rendering path using differentiable rasterization~\cite{laine2020modular} and sparse-grid interpolation operators~\cite{xiang2024flexgemm}.
The interpolation in Eq.~\eqref{eq:grid_query} is differentiable and can be evaluated in parallel, supporting real-time rendering and efficient backpropagation to the voxel attributes.
GPU-resident TexF uses the same rasterization stage, but interpolates latent features and decodes them with an MLP to obtain texture attributes (Sec.~\ref{sec:gpu_compression}).

\subsubsection{Rendering-driven Optimization}
We use our differentiable renderer to refine texture appearance through inverse rendering. We render reference images $\{\mathbf{I}_i\}_{i=1}^N$ from the original lossless UV-textured mesh and corresponding images $\{\widehat{\mathbf{I}}_i\}_{i=1}^N$ from the target mesh with TexF under the same views. We minimize the photometric objective
\vspace{-0.1cm}
\begin{equation}
\mathcal{D}=\sum_{i=1}^{N}\left[\left\|\widehat{\mathbf{I}}_i-\mathbf{I}_i\right\|_1+1-\mathrm{SSIM}(\widehat{\mathbf{I}}_i,\mathbf{I}_i)\right],
\label{eq:render-loss}
\vspace{-0.1cm}
\end{equation}
where the $\ell_1$ term measures pixel-wise color errors and the SSIM term encourages structural similarity. 
Starting from the transferred attributes, we optimize $\mathbf{t}(\mathbf{q})$ while keeping the mesh geometry and voxel support fixed.

Image-space supervision accounts for the sampling and interpolation used during rendering, allowing refinement to compensate for errors left by direct texture transfer and voxel discretization. It therefore improves the texture representation without increasing its capacity.
This refinement is also efficient, requiring only a modest optimization cost in our experiments.
The same objective also optimizes the latent features and decoder parameters of the compressed neural field (Sec.~\ref{sec:gpu_compression}).

Beyond compression refinement, the differentiable renderer could support mesh texturing directly from multi-view images. Extending it to jointly optimize geometry and appearance is another direction for future work.

\subsection{TexF Bitstream Compression}\label{sec:compression}

For disk storage and transmission, we seek a compact entropy-coded texture bitstream~\cite{vdmc21cfp,AOM_VVM_CfP_2023}. TexF derives voxel locations from the mesh rather than encoding them separately.
As illustrated in Fig.~\ref{fig:compression}, mesh geometry is encoded using an existing mesh codec~\cite{maglo20153d,peng2005technologies,draco}.
The encoder and decoder apply the same voxelization procedure to the reconstructed mesh to obtain the voxel set $\mathbf{Q}$.
These shared locations establish the correspondence between decoded attributes and TexF voxels, allowing the texture stream to carry coded attributes without per-voxel coordinates.

To encode the attributes, we leverage established 3D coding tools.
Just like UV-based pipelines use image or video codecs such as HEVC~\cite{HEVC} and AV1~\cite{han2021av1}, TexF interfaces with point cloud attribute codecs through its compatible coordinate--attribute structure.
Their transform, prediction, and entropy-coding tools can therefore exploit spatial correlations directly in mesh-derived voxel neighborhoods, without requiring a texture-specific codec architecture.

The interface is \textit{codec-agnostic}: compatible 3D attribute codecs can be used without changing the TexF representation.
We use Unicorn~\cite{unicorn2025attr} as the primary backend, leveraging its multiscale neural transform and conditional entropy model.
We also evaluate conventional RAHT~\cite{de2016compression}-based attribute coding in G-PCC~\cite{MPEG-PCC-TMC13} to examine whether TexF's compression gains extend across different coding architectures.
Beyond these two backends, the same interface allows TexF to benefit from advances in compatible 3D attribute codecs without changing the representation.
At the decoder, reconstructed attributes are restored to their corresponding voxel locations, recovering $\widehat{\mathcal{T}}$.

\subsection{GPU-Resident TexF Compression}\label{sec:gpu_compression}

GPU-resident compression aims to reduce texture memory while supporting low-latency random access during rendering~\cite{ntc2023}.
Although TexF is spatially sparse, its explicit attributes and lookup structure can still occupy substantial GPU memory at high resolutions. For example, our uncompressed implementation stores three 8-bit RGB values and a 64-bit lookup key per voxel.
We therefore develop 3D Neural Texture Compression (\textbf{3DNTC}), which exploits TexF's 3D-native organization to compress surface appearance into quantized features and a lightweight decoder.
Unlike NTC~\cite{ntc2023}, which decodes texture attributes in the 2D UV domain, 3DNTC reconstructs them directly at continuous mesh surface positions, without UV mapping or explicit per-voxel lookup entries.

\subsubsection{Quantized Hash Representation}
3DNTC uses two spatially downsampled hash grids with INT4 features to capture texture variations at different spatial scales, together with a shared four-layer FP16 MLP (Fig.~\ref{fig:compression_memory}).
At level $i\in\{1,2\}$, the grid resolution is $\lceil K/s_i\rceil$, where $K$ is the reference spatial resolution set according to the source texture resolution, and $s_i$ is the downsampling factor (e.g., 4 or 8).
Following multiresolution hash encoding~\cite{muller2022instant}, each level uses a table of $2^{h_i}$ entries with $F$ feature channels, indexed by hashing integer grid coordinates.
These entries store shared latent features, without explicit per-voxel lookup keys.
Thus, the downsampling factor $s_i$ controls spatial granularity, while $h_i$ and $F$ determine latent storage; varying these parameters adjusts the rate--distortion trade-off.

\subsubsection{Surface-Query Decoding}
Rasterization provides visible surface positions as in Sec.~\ref{sec:rep_render}.
For each position $\mathbf{p}$, we hash the eight surrounding grid-corner coordinates at each level and trilinearly interpolate their features.
The interpolated features are concatenated across levels into $\Phi(\mathbf{p})\in\mathbb{R}^{2F}$ and combined with an 18-dimensional positional encoding $\gamma(\mathbf{p})$.
A single evaluation of the decoder $D_{\boldsymbol{\theta}}$ then reconstructs color:
\begin{equation}
\widehat{\mathbf{t}}(\mathbf{p})
= D_{\boldsymbol{\theta}}
\!\left(\Phi(\mathbf{p}),\gamma(\mathbf{p})\right),
\label{eq:ntc_decode}
\end{equation}
Interpolating features before decoding yields the attribute directly at $\mathbf{p}$, avoiding separate neural decoding of the eight voxel-corner colors.
The decoder has three hidden layers with 64, 48, and 32 channels and hardGELU activations, followed by a linear output layer; 
its parameters are shared across all surface queries within each asset.
Only the packed latent tables and decoder parameters need to remain resident as texture data.
Queries are independent and require no sequential entropy decoding, enabling on-demand reconstruction during rendering.

\subsubsection{Training and Refinement}
For each asset, we sample continuous surface positions and obtain target attributes from the refined explicit TexF.
We jointly optimize the latent tables and MLP to minimize attribute mean squared error, simulating INT4 quantization during training.
We then refine both against ground-truth multi-view renderings using Eq.~\eqref{eq:render-loss}, improving rendered fidelity without increasing the payload.
At deployment, only the packed latent tables and decoder parameters are needed as GPU-resident texture data.
Detailed settings are provided in the \textit{Supplementary Material}.

\section{Experiments}
\label{sec:exp}
In this section, we first present the common experimental settings and assess TexF's representation fidelity, compactness, and rendering efficiency. We then compare its bitstream and GPU-resident compression performance with UV-based and other alternatives, followed by ablation studies and further analysis.

\subsection{Experimental Settings}

\subsubsection{Datasets}
\begin{figure}[t]
    \centering
    \includegraphics[width=0.995\linewidth]{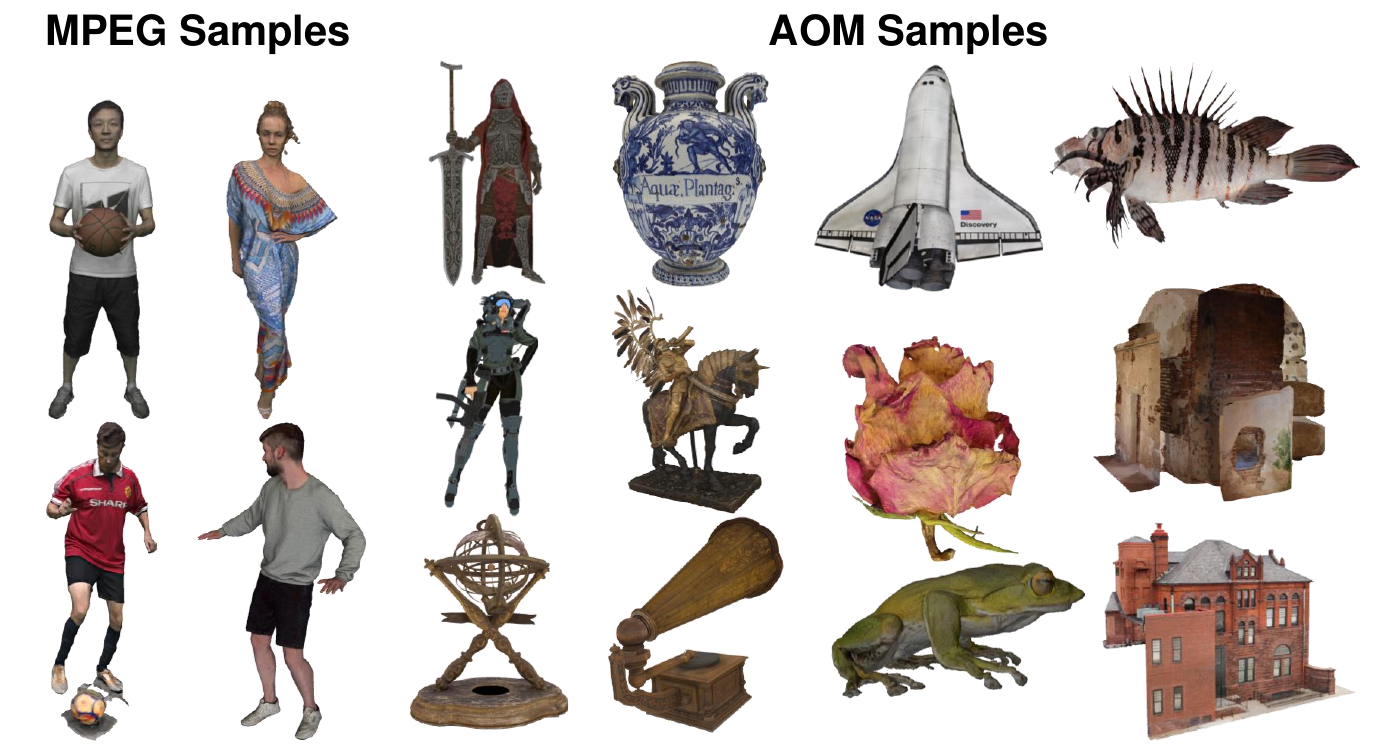}
    \caption{Representative samples from MPEG (8 assets) and AOM (32 assets). AOM covers a broader range of asset categories. It also has more complex geometry on average (410K vs. 37K faces).}
    \label{fig:dataset}
\end{figure}

To evaluate performance across diverse geometries and textures, we use eight human models from the MPEG mesh compression benchmark~\cite{vdmc24ctc} and 32 assets from the AOMedia mesh coding test set~\cite{AOM_VVM_CfP_2023}. The AOM assets include scanned objects, plants and animals, DCC-created assets, and architectural models.
Together, the two benchmarks comprise 40 assets with texture resolutions ranging from 2K to 4K.\footnote{We use 1K, 2K, and 4K to denote resolutions of 1024, 2048, and 4096, respectively.}
Detailed asset statistics and the complete test sets are provided in the \textit{Supplementary Material}, and representative assets are shown in Fig. \ref{fig:dataset}.

\subsubsection{Evaluation Protocol and Metrics}
To isolate texture-coding efficiency, all methods use the same geometry and evaluation protocol.
Following the common test conditions~\cite{vdmc24ctc} and MPEG official metrics\footnote{\url{https://github.com/MPEGGroup/mpeg-pcc-mmetric}}~\cite{mmmetric-paper}, we assess quality using renderings from 16 fixed orthographic views distributed on a Fibonacci sphere at $2048\times2048$ resolution, evaluated using PSNR, SSIM, and LPIPS~\cite{zhang2018unreasonable} metrics, complemented by point-sampled 3D-PSNR following the standard evaluation procedure.
The 16 fixed orthographic evaluation views are held out from rendering-driven optimization, which instead uses independently sampled perspective views.

We compare only the texture-related payload, since \textit{all methods use the same geometry}.
Rate is measured as bitstream size (KB) or GPU-resident memory (MB). 
We quantify relative bitrate savings using BD-BR~\cite{BDrate} and provide detailed R--D curves. 
Detailed payload accounting is provided in the corresponding sections.
Computational efficiency is evaluated by runtime. We use an NVIDIA RTX~A6000 GPU (48~GB) with Intel Xeon Silver 4309Y CPUs for runtime measurements.

\subsection{Representation and Rendering Performance}\label{sec:rep_analysis}
Because the representation is the basis for subsequent compression, we first assess the fidelity and compactness of TexF, together with its construction and rendering efficiency.
We compare TexF with UV-reparameterized textures (UV-Reparam) generated using UVAtlas~\cite{UVAtlas}, a tool commonly used in mesh compression~\cite{vdmc25whitepaper,AOM_VVM_CfP_2023} to improve spatial continuity and coding efficiency through chart reparameterization and packing.
Both are evaluated against renderings of the source UV-textured mesh, following the evaluation protocol described above. We also examine how rendering-driven refinement improves TexF fidelity.

\begin{figure}[t]
\centering
\begin{subfigure}[t]{\columnwidth}
    \centering
    \includegraphics[width=0.32\linewidth]{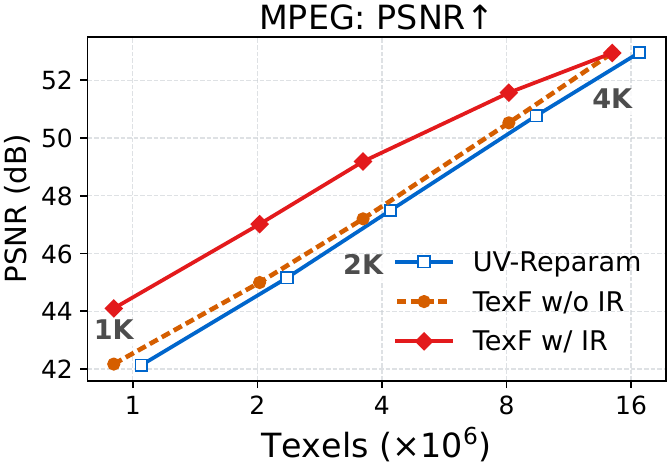}\hfill
    \includegraphics[width=0.32\linewidth]{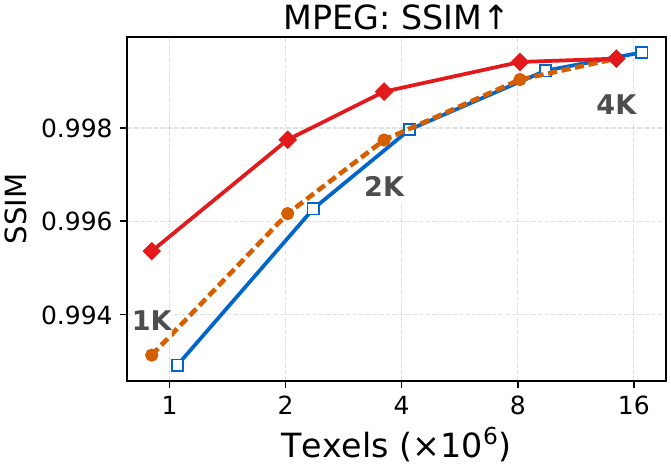}\hfill
    \includegraphics[width=0.32\linewidth]{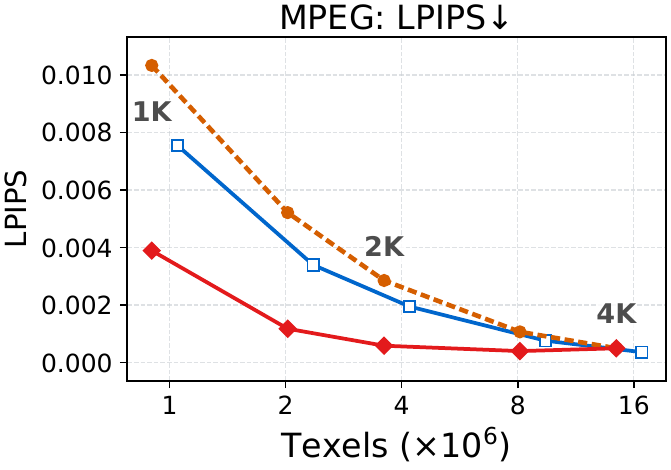}
\end{subfigure}

\vspace{0.15em}

\begin{subfigure}[t]{\columnwidth}
    \centering
    \includegraphics[width=0.32\linewidth]{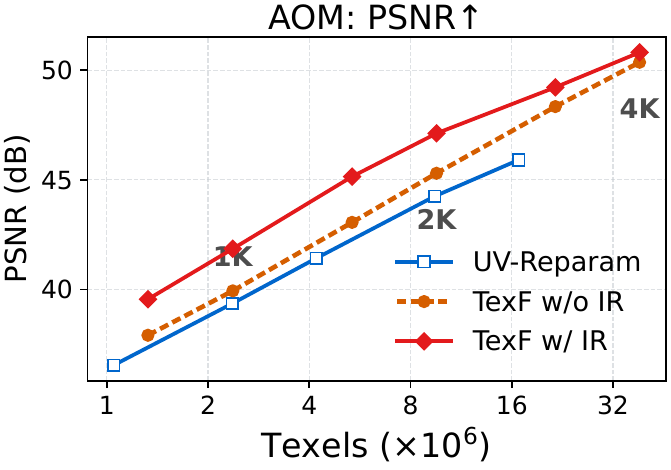}\hfill
    \includegraphics[width=0.32\linewidth]{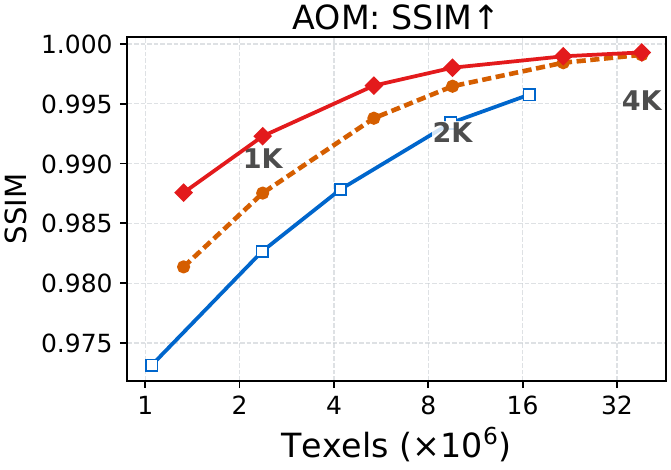}\hfill
    \includegraphics[width=0.32\linewidth]{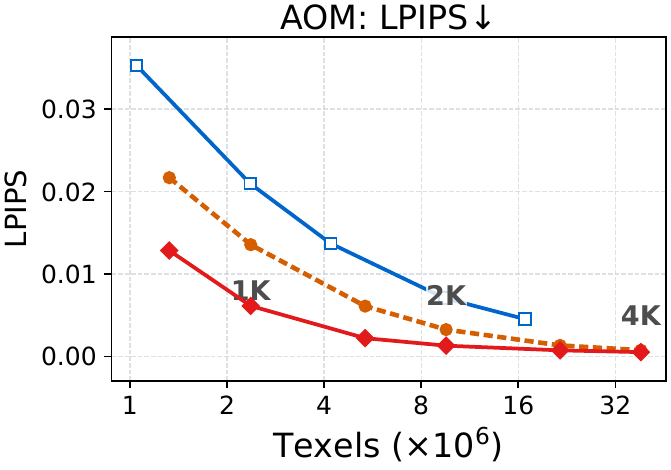}
\end{subfigure}

\vspace{0.15em}

\begin{subfigure}[t]{\columnwidth}
    \centering
    \includegraphics[height=0.22\linewidth]{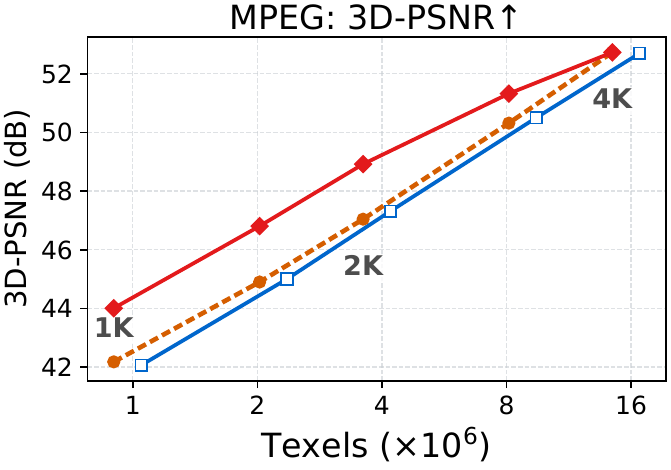}\hspace{0.02\linewidth}%
    \includegraphics[height=0.22\linewidth]{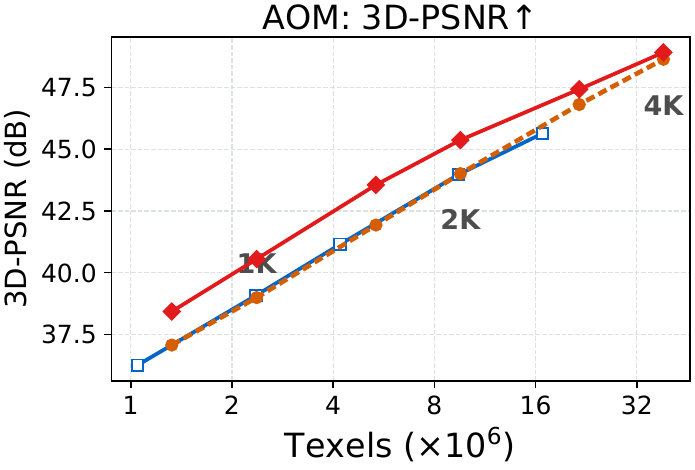}
\end{subfigure}
\caption{\textbf{Representation fidelity versus capacity.}
Across four quality metrics on the MPEG and AOM datasets, TexF offers a favorable capacity--quality trade-off, with further gains from IR.}
\label{fig:representation_capacity_quality}
\end{figure}

\begin{figure}[t]
\centering
\includegraphics[width=0.99\linewidth]{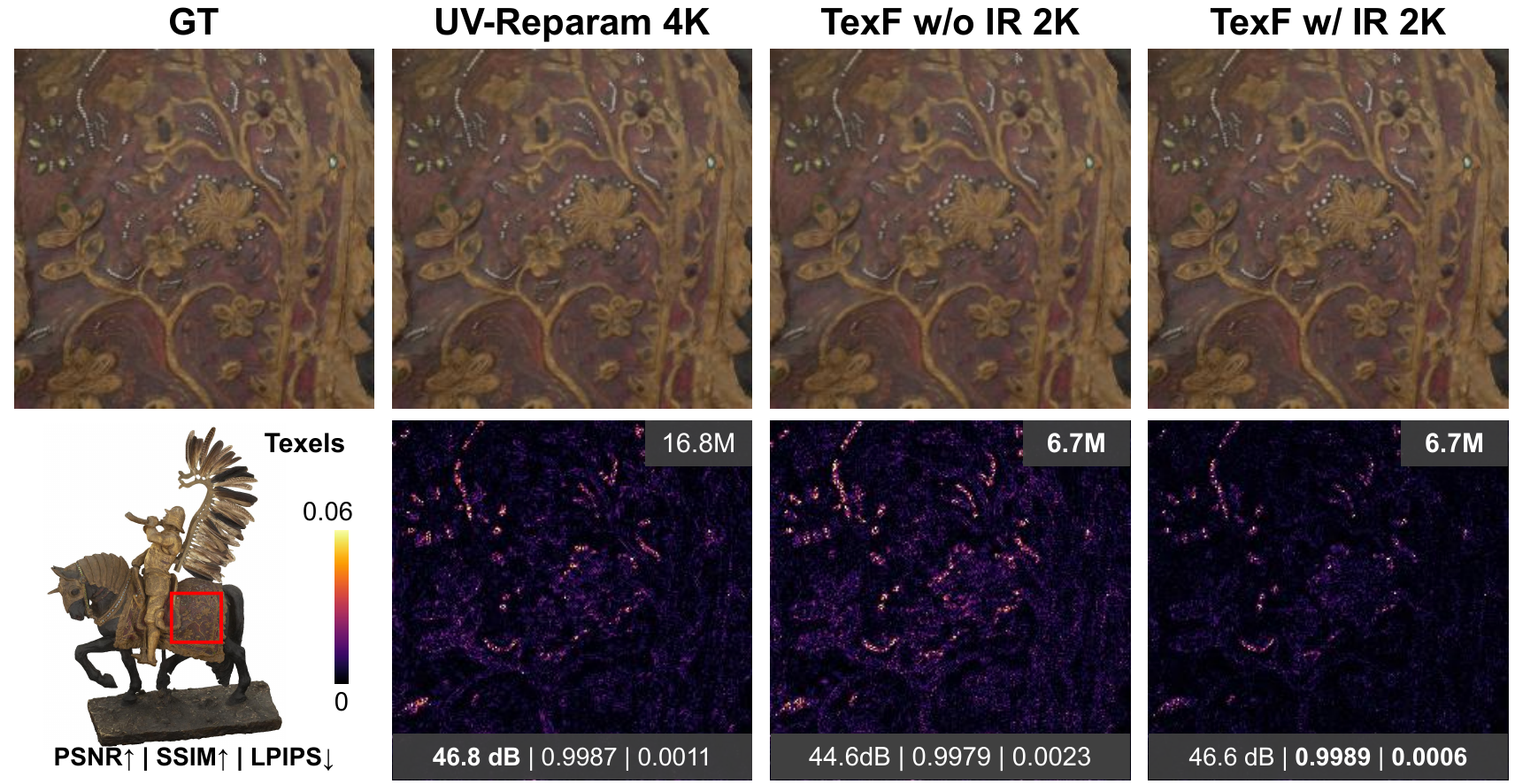}
\caption{\textbf{Visual comparison of representation fidelity.} TexF preserves comparable or better texture detail with fewer texels, while IR further improves fidelity.}
\label{fig:vis-rep}
\end{figure}

\begin{table}[t]
\centering
\small
\setlength{\tabcolsep}{3.0pt}
\caption{\textbf{Representation construction cost and rendering efficiency.}
TexF remains substantially faster to construct even with IR, while maintaining high rendering throughput.
}
\label{tab:scalability}
\resizebox{\columnwidth}{!}{%
\begin{tabular}{llcccccc}
\toprule
\textbf{Metric} & \textbf{Method} &
\multicolumn{3}{c}{\textbf{MPEG}} & \multicolumn{3}{c}{\textbf{AOM}} \\
\cmidrule(lr){3-5}\cmidrule(lr){6-8}
& & \textbf{1K} & \textbf{2K} & \textbf{4K} & \textbf{1K} & \textbf{2K} & \textbf{4K} \\
\midrule
\multirow{3}{*}{Construction [s]}
& UV-Reparam & 97 & 109 & 113 & 1674 & 1724 & 1716 \\
& TexF w/o IR & 0.04 & 0.14 & 0.43 & 0.05 & 0.16 & 0.45 \\
& TexF w/ IR & 4.90 & 5.45 & 7.29 & 5.37 & 6.75 & 9.83 \\
\midrule
\multirow{2}{*}{Rendering [FPS]}
& UV & 227 & 231 & 230 & 136 & 135 & 135 \\
& TexF & 222 & 194 & 190 & 127 & 107 & 107 \\
\bottomrule
\end{tabular}
}
\end{table}

\subsubsection{Representation Fidelity and Compactness}
We vary the resolutions of TexF and UV-Reparam from 1K to 4K to obtain different representation capacities.
Here, $K$ denotes spatial resolution for TexF and atlas resolution for UV-Reparam. 
Note that equal resolutions do not imply equal capacity. We therefore measure capacity by the actual number of \textit{texels}: voxels for TexF and pixels for UV-Reparam.

Although TexF is defined at a spatial resolution of $K^3$, its surface-aligned design stores attributes only in voxels intersecting the mesh surface.
Its texel count therefore grows approximately quadratically with $K$, similar to the scaling of 2D texture maps. This \textit{surface-localized sparsity} supports resolutions of 4K and beyond.
Fig.~\ref{fig:representation_capacity_quality} shows that fidelity improves with capacity for both representations, while TexF generally achieves comparable or higher fidelity than UV-Reparam with fewer texels.
Visual comparisons in Fig.~\ref{fig:vis-rep} further demonstrate its ability to preserve fine texture details.

\vspace{0.5em}
\noindent\textbf{Inverse Rendering Refinement.}
Direct UV-to-TexF texture transfer provides a strong initialization. Inverse rendering (IR) further compensates for transfer errors, improving fidelity without increasing representation capacity.
By default, we perform 30 optimization iterations, each using a batch of eight newly sampled perspective views, totaling 240 random views at a rendering resolution of $1024\times1024$.
Cameras look at the object center with a $45^\circ$ FOV. Azimuth, sine elevation, and distance are drawn from uniform distributions over $[0,2\pi)$, $[-1,1]$, and $[1.2,3.0]$, respectively.
These training views are separate from the 16 fixed orthographic Fibonacci evaluation views mentioned above and differ in both viewing direction and projection.

As shown in Fig.~\ref{fig:representation_capacity_quality}, IR improves PSNR and 3D-PSNR by 1.4 and 1.2~dB on average, alongside higher SSIM and lower LPIPS.
Gains in LPIPS and 3D-PSNR extend beyond the objective in Eq.~\eqref{eq:render-loss}. Evaluation with different viewpoints and projection further suggests genuine improvements in texture fidelity, rather than overfitting to specific training views or metrics.
The visual comparisons in Fig.~\ref{fig:vis-rep} also illustrate the texture details recovered by refinement.
The improvement is more pronounced at lower resolutions and diminishes at higher resolutions, particularly at 4K, where dense sampling already provides high-fidelity transfer, leaving little room for refinement.

\subsubsection{{Representation Construction and Rendering Efficiency}}

TexF is constructed through surface voxelization and texture transfer, whereas UVAtlas requires more complex UV reparameterization and atlas packing.
As shown in Table~\ref{tab:scalability}, TexF construction takes less than 1~s on average even at 4K resolution, and remains below 10~s with refinement.
By contrast, UVAtlas requires approximately {100}~s on MPEG and more than 1600~s on the more complex AOM dataset. Overall, TexF reduces representation construction time by orders of magnitude relative to UVAtlas reparameterization.
At high resolutions like 4K, IR can be shortened or omitted to balance fidelity and construction cost.

Following the evaluation protocol above, we measure rendering throughput at $2048\times2048$ resolution. TexF averages 190--222 FPS on MPEG and 107--127 FPS on AOM across the 1K--4K resolutions. TexF thus supports real-time rendering through sparse 3D queries, with throughput close to that of UV rendering.

\subsection{Bitstream Compression Performance}
We evaluate whether surface-aligned TexF improves texture compression over established UV-based pipelines and whether its gains extend across different 3D coding backends.

\subsubsection{Experimental Configurations}
We denote each configuration by its representation and codec.

\begin{itemize}
    \setlength{\itemsep}{0.15em}
    \setlength{\topsep}{0.25em}
    \setlength{\parsep}{0pt}

    \item \textbf{UV-Source.}
    We retain the source UV parameterization and encode its texture atlas.
    We use the V-DMC reference software~\cite{vdmc25whitepaper,VDMCSoftware}, with its default HEVC texture codec~\cite{HEVC} as the baseline, and additionally evaluate AV1~\cite{han2021av1} and ELIC~\cite{he2022elic} to assess potential gains from alternative texture codecs. UV mapping data and HEVC textures are encoded following the V-DMC CTC~\cite{vdmc24ctc}.

    \item \textbf{UV-Reparam.}
    We use V-DMC's default UVAtlas pipeline~\cite{UVAtlas} to reparameterize the mesh and pack its charts. UV mapping and texture coding follow the same settings as \textit{UV-Source}.

    \item \textbf{TexF.}
    TexF replaces the UV atlas with a surface-aligned representation.
    We encode its attributes directly in 3D using the released pretrained Unicorn model~\cite{unicorn2025attr} without retraining as the primary backend (\textit{TexF/Unicorn}).
    We additionally encode TexF with conventional G-PCC in its official configuration~\cite{MPEG-PCC-TMC13} (\textit{TexF/G-PCC}) to evaluate its compatibility with different 3D codecs.

    \item \textbf{Serialized.} 
    We evaluate serialization followed by 2D coding as an alternative to direct 3D attribute coding, inspired by prior voxel-texturing work~\cite{dolonius2019compressing}. We order and pack the attributes of refined TexF representations into 2D images using Morton or Hilbert curves, then encode them with HEVC. We denote the two variants as \textit{Serialized-Morton/HEVC} and \textit{Serialized-Hilbert/HEVC}.
\end{itemize}

All methods use the same original geometry, losslessly coded with V-DMC; we therefore compare only texture-related bitstreams, including UV mapping data for UV-based methods.
Detailed codec settings and additional results under lossy geometry are provided in the \textit{Supplementary Material}.

\begin{table*}[t]
\centering
\caption{\textbf{Bitstream compression.} BD-BR is computed against \textit{UV-Source/HEVC}; negative values indicate bitrate savings.}
\label{tab:compression}
\footnotesize
\setlength{\tabcolsep}{2.2pt}
\renewcommand{\arraystretch}{0.98}
\resizebox{\textwidth}{!}{%
\begin{tabular}{@{}l@{\hspace{4pt}}cccc@{\hspace{5pt}}ccc@{\hspace{8pt}}cccc@{\hspace{5pt}}ccc@{}}
\toprule
\multirow{3}{*}{\textbf{Method}}
& \multicolumn{7}{c}{\textbf{MPEG Dataset}}
& \multicolumn{7}{c}{\textbf{AOM Dataset}} \\
\cmidrule(lr){2-8}\cmidrule(l){9-15}
& \multicolumn{4}{c}{\textbf{BD-BR (\%) $\downarrow$}} & \multicolumn{3}{c}{\textbf{Time (s)}}
& \multicolumn{4}{c}{\textbf{BD-BR (\%) $\downarrow$}} & \multicolumn{3}{c}{\textbf{Time (s)}} \\
\cmidrule(lr){2-5}\cmidrule(lr){6-8}\cmidrule(lr){9-12}\cmidrule(l){13-15}
& \textbf{PSNR} & \textbf{SSIM} & \textbf{LPIPS} & \textbf{3D-PSNR} & \textbf{Prep.} & \textbf{Enc.} & \textbf{Dec.}
& \textbf{PSNR} & \textbf{SSIM} & \textbf{LPIPS} & \textbf{3D-PSNR} & \textbf{Prep.} & \textbf{Enc.} & \textbf{Dec.} \\
\midrule
\textbf{UV-Source}/HEVC
& \multicolumn{4}{c}{\textit{Anchor}} & \multirow{3}{*}{1.8} & 41.1 & 6.7
& \multicolumn{4}{c}{\textit{Anchor}} & \multirow{3}{*}{3.1} & 64.6 & 10.5 \\
\textbf{UV-Source}/AV1
& -7.4 & -8.4 & -3.8 & -7.4 & & 71.4 & 4.6
& -5.9 & -5.7 & -5.0 & -5.8 & & 139.3 & 7.5 \\
\textbf{UV-Source}/ELIC
& -10.0 & -14.1 & -2.8 & -10.0 & & 2.5 & 1.6
& -13.0 & -11.0 & -10.1 & -13.3 & & 3.6 & 2.3 \\
\cmidrule(lr){1-15}
\textbf{UV-Reparam}/HEVC
& -28.2 & -29.7 & -27.3 & -27.7 & \multirow{3}{*}{111} & 38.1 & 8.5
& -45.4 & -42.7 & -42.1 & -42.5 & \multirow{3}{*}{1716} & 66.7 & 14.3 \\
\textbf{UV-Reparam}/AV1
& \ranksecond{-33.8} & -35.9 & \rankthird{-30.0} & \ranksecond{-33.3} & & 81.1 & 5.5
& -48.4 & -45.4 & -45.0 & -45.6 & & 112.9 & 8.8 \\
\textbf{UV-Reparam}/ELIC
& \rankthird{-32.8} & \rankthird{-36.8} & -25.2 & -32.5 & & 2.5 & 1.6
& \rankthird{-52.4} & \rankthird{-48.1} & \rankthird{-47.1} & \ranksecond{-50.0} & & 3.9 & 2.6 \\
\cmidrule(lr){1-15}
\textbf{Serialized-Morton}/HEVC
& 111.9 & 133.5 & 54.6 & 113.0 & 7.5 & 25.0 & 3.1
& 20.8 & 35.0 & -2.2 & 40.9 & 16.9 & 70.1 & 7.2 \\
\textbf{Serialized-Hilbert}/HEVC
& 77.6 & 96.4 & 28.8 & 78.5 & 10.3 & 24.6 & 3.3
& 4.9 & 12.1 & -17.8 & 19.6 & 50.1 & 62.9 & 7.1 \\
\cmidrule(lr){1-15}
\textbf{TexF}/G-PCC
& \rankthird{-32.8} & \ranksecond{-41.1} & \rankfirst{-52.0} & \rankthird{-32.6} & 5.1 & 8.9 & 8.1
& \ranksecond{-55.1} & \ranksecond{-59.5} & \ranksecond{-68.0} & \rankthird{-47.1} & 5.8 & 21.4 & 18.7 \\
\textbf{TexF}/Unicorn
& \rankfirst{-46.7} & \rankfirst{-48.9} & \ranksecond{-45.5} & \rankfirst{-46.4} & 5.8 & 15.1 & 8.1
& \rankfirst{-63.3} & \rankfirst{-64.2} & \rankfirst{-70.1} & \rankfirst{-53.2} & 7.4 & 38.8 & 20.0 \\
\cmidrule(lr){1-15}
\textbf{TexF}/Unicorn (w/o IR)
& -51.7 & -49.2 & -41.8 & -51.5 & 0.2 & 15.1 & 8.1
& -62.9 & -62.4 & -67.7 & -54.2 & 0.2 & 38.8 & 20.0 \\
\textbf{TexF}/Unicorn (UV-baked)
& -45.9 & -48.4 & -43.7 & -45.2 & 5.8 & 15.1 & 8.1
& -59.8 & -63.5 & -68.2 & -52.7 & 7.4 & 38.8 & 20.0 \\
\bottomrule
\end{tabular}%
}
\vspace{0.35em}
\parbox{\textwidth}{\footnotesize Preprocessing time includes representation construction and any required fitting or serialization; representation-level costs are reported in Table~\ref{tab:scalability}. \colorbox{rankgold}{Gold}, \colorbox{ranksilver}{silver}, and \colorbox{rankbronze}{bronze} backgrounds indicate the best, second-best, and third-best BD-BR values, respectively. Ablations are excluded from ranking.}
\end{table*}

\begin{figure*}[t]
\centering
\includegraphics[width=0.78\linewidth]{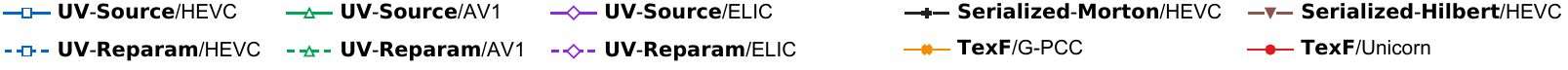}\par
\vspace{0.5em}
\includegraphics[width=0.249\linewidth]{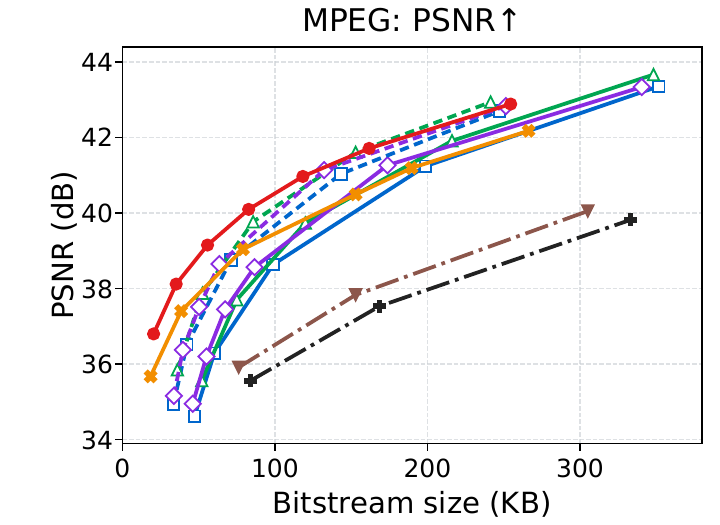}\hfill
\includegraphics[width=0.249\linewidth]{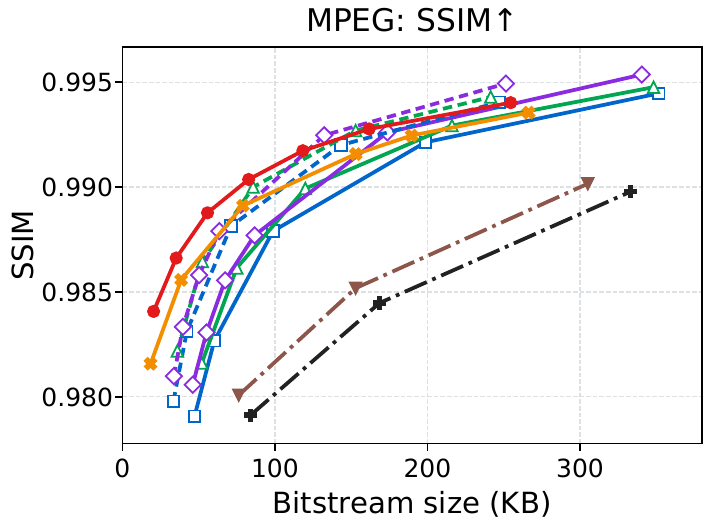}\hfill
\includegraphics[width=0.249\linewidth]{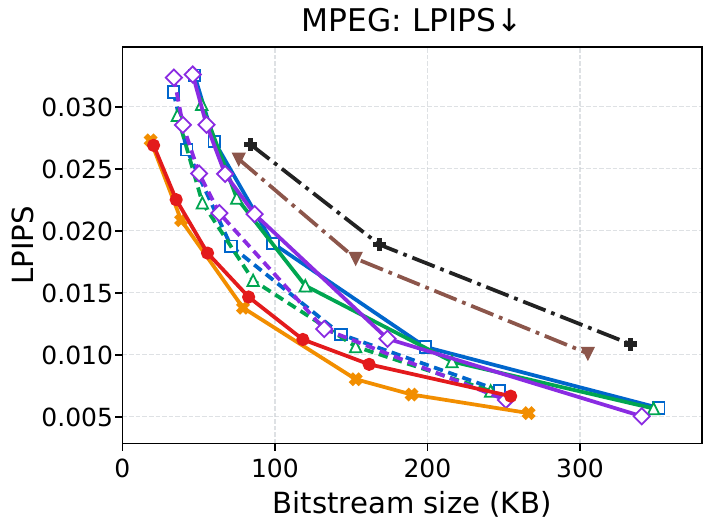}\hfill
\includegraphics[width=0.249\linewidth]{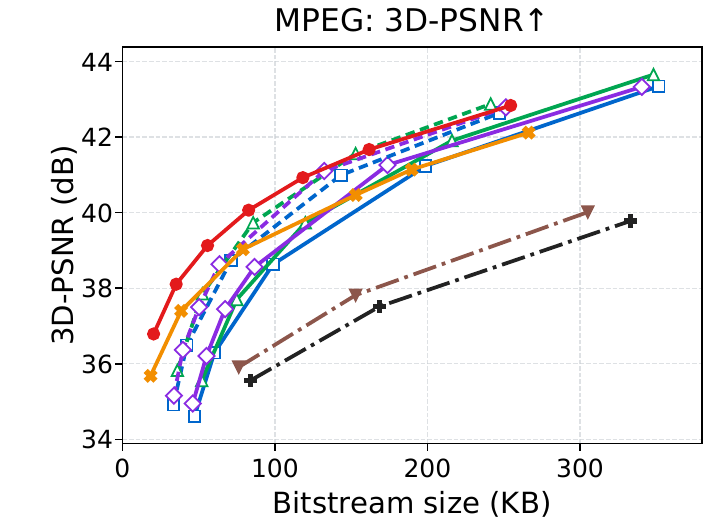}

\vspace{1.0em}

\includegraphics[width=0.249\linewidth]{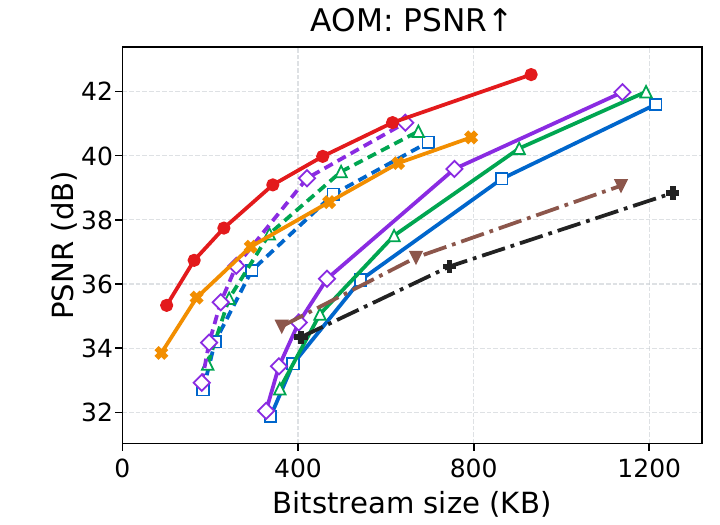}\hfill
\includegraphics[width=0.249\linewidth]{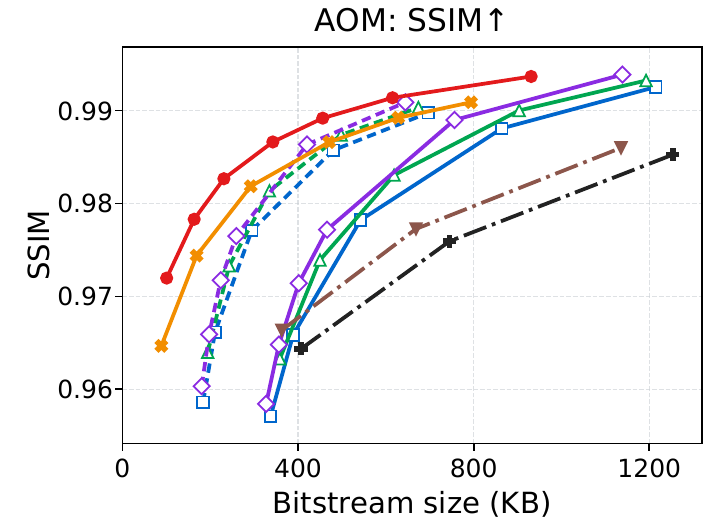}\hfill
\includegraphics[width=0.249\linewidth]{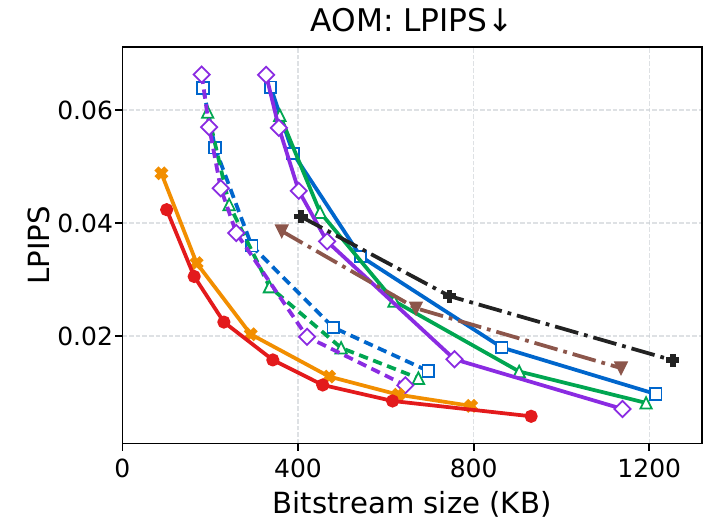}\hfill
\includegraphics[width=0.249\linewidth]{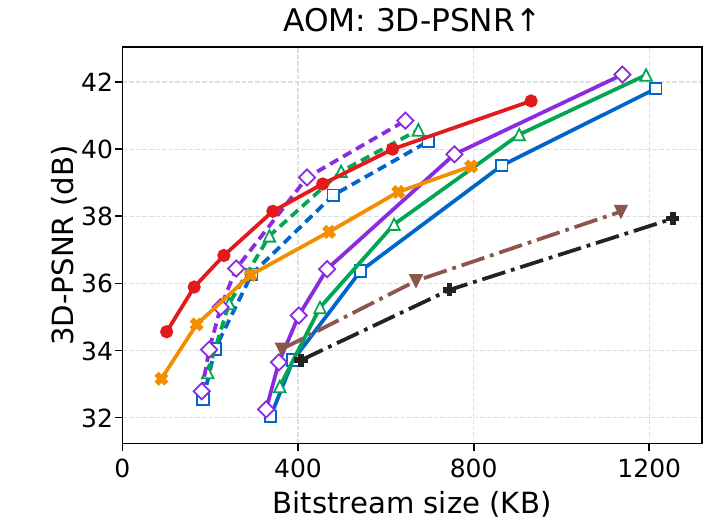}
\caption{\textbf{Rate--distortion performance on the MPEG and AOM datasets.}}
\label{fig:bitstream_rd_kb}
\label{fig:bitstream_serialized_psnr}
\label{fig:bitstream_rd_joint_comparison}
\end{figure*}

\begin{figure*}[thbp]
\centering
\includegraphics[width=\linewidth]{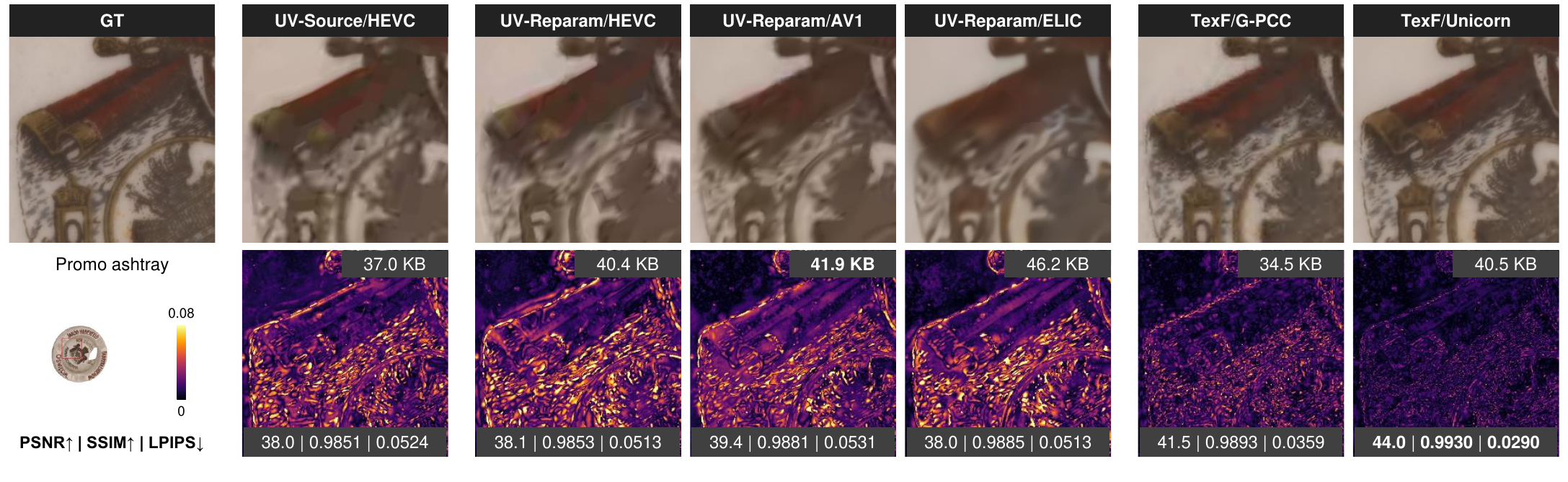}

\includegraphics[width=\linewidth]{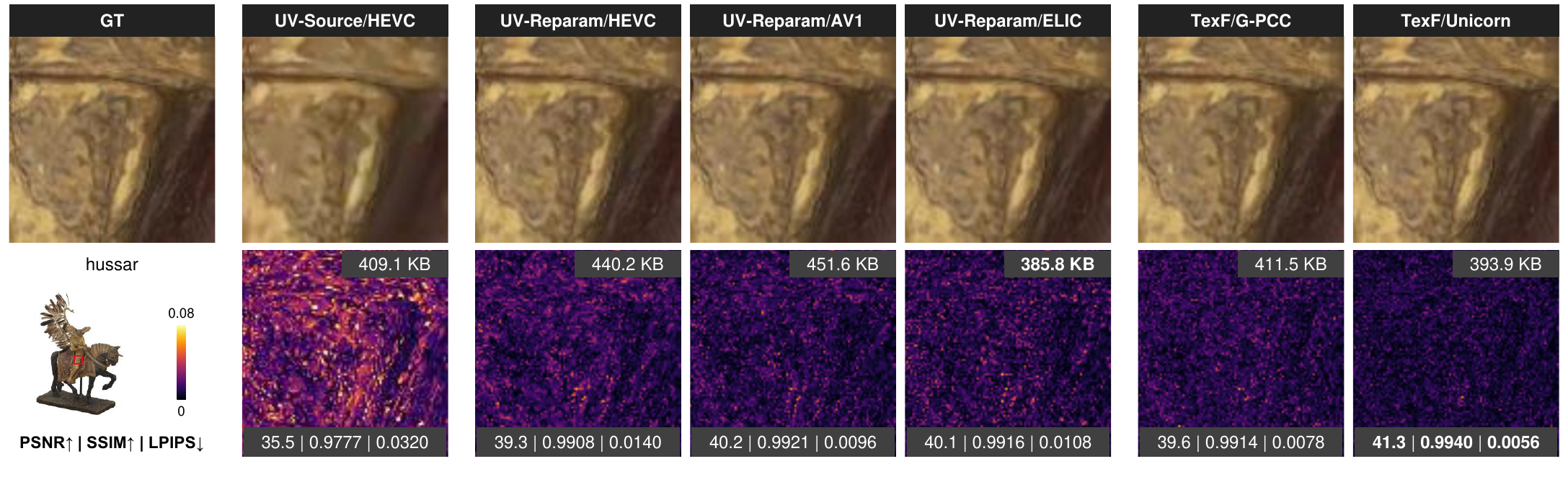}
\caption{\textbf{Visual comparison of texture reconstruction under bitstream compression.}}
\label{fig:bitstream_visual_comparison}
\end{figure*}

\subsubsection{Rate--Distortion Performance}
Fig.~\ref{fig:bitstream_rd_kb} compares the R--D performance of the different methods, while Table~\ref{tab:compression} summarizes their BD-BR results using \textit{UV-Source/HEVC} as the anchor.
With the default HEVC codec, UV reparameterization yields BD-BR savings of approximately 27\%--30\% on MPEG and 42\%--45\% on AOM across the four quality metrics, demonstrating the benefit of a more compression-friendly atlas. Replacing HEVC with AV1 or ELIC generally improves coding efficiency further. 

TexF/Unicorn outperforms all evaluated UV configurations across the four metrics on both datasets, including UV-Reparam with AV1 or ELIC. Relative to \textit{UV-Source/HEVC}, it achieves approximately 46\%--49\% BD-BR savings on MPEG and 53\%--70\% on AOM.
\textit{TexF/G-PCC} also delivers substantial savings, particularly under LPIPS, and remains competitive with or better than the strongest UV baselines in most cases. The gains with both Unicorn and G-PCC show that TexF's compression benefits are not confined to a single coding backend.

\begin{figure*}[t]
\centering
\begin{subfigure}[t]{0.36\textwidth}
\centering
\includegraphics[width=\linewidth]{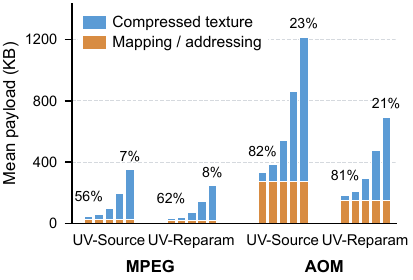}
\caption{UV/HEVC}
\label{fig:uv_payload_hevc}
\end{subfigure}\hfill
\begin{subfigure}[t]{0.36\textwidth}
\centering
\includegraphics[width=\linewidth]{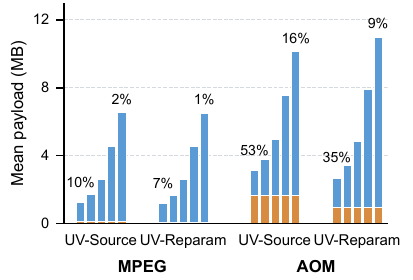}
\caption{UV/ASTC}
\label{fig:uv_payload_astc}
\end{subfigure}\hfill
\begin{subfigure}[t]{0.23\textwidth}
\centering
\includegraphics[width=\linewidth]{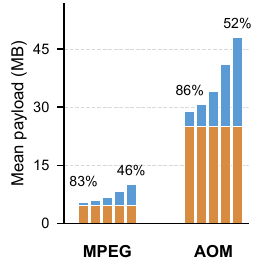}
\caption{Serialized-Morton/ASTC}
\label{fig:morton_payload_astc}
\end{subfigure}
\caption{\textbf{Texture-related payload breakdown.}
(a) Bitstream compression; (b,c) GPU-resident compression.
Texture bitrate increases from left to right within each group.
Percentages denote the UV or DAG share of the total payload.}\label{fig:uv_payload_breakdown}
\end{figure*}

UV-based pipelines must transmit UV mapping data, whereas TexF derives texture locations directly from the shared mesh through its surface-aligned design. For each UV representation, the mapping payload remains fixed across texture bitrate points.
As Fig.~\ref{fig:uv_payload_breakdown}(\subref{fig:uv_payload_hevc}) shows, this payload accounts for a larger share at lower texture rates and on the more complex AOM assets.
For UV-Source/HEVC on AOM, its share decreases from 82\% to 23\% as texture bitrate increases.
This helps explain TexF's greater compression gains in these settings and highlights the benefits of avoiding UV mapping.

Among the serialization baselines in Fig.~\ref{fig:bitstream_serialized_psnr}, Serialized-Hilbert outperforms Serialized-Morton, yet both of them remain substantially less efficient than TexF/Unicorn and TexF/G-PCC. These results show that removing UV mapping and applying a mature 2D codec to serialized voxel attributes do not by themselves ensure competitive compression. They highlight the importance of matching texture organization with coding tools that exploit its spatial correlations.

The visual comparisons in Fig.~\ref{fig:bitstream_visual_comparison} further illustrate these differences. At comparable rates, TexF preserves coherent surface details more effectively, whereas the UV-based reconstructions exhibit localized blurring and discontinuities around some chart boundaries.
Additional visual comparisons are provided in the \textit{Supplementary Material}.
Per-asset performance and compatibility with conventional UV rendering are examined separately in Sec.~\ref{sec:ablation_analysis}.

\subsubsection{Computational Efficiency}

Table~\ref{tab:compression} reveals a clear preprocessing trade-off. \textit{UV-Source} avoids reparameterization, whereas \textit{UV-Reparam} incurs an atlas construction cost of approximately 100--1800~s. TexF requires approximately 5--8~s of preprocessing, dominated by rendering-driven refinement. Even without refinement, TexF retains its compression advantage with less than one second of preprocessing, as shown in Sec.~\ref{sec:ir_ablation}. It thus combines efficient construction with strong compression performance.

Encoding and decoding costs depend on the backend. In our measurements, Unicorn and G-PCC encode faster than the evaluated HEVC and AV1 implementations, whereas ELIC offers the shortest encoding and decoding times.

\subsection{GPU-Resident Compression Performance}

GPU-resident compression keeps appearance data in GPU memory for random-access decoding during rendering. We evaluate its rate--distortion performance and decoding cost.

\subsubsection{Experimental Configurations}
We evaluate the same four representation pipelines as in the bitstream experiments, using codecs for GPU-resident random-access decoding. For both UV variants, the resident payload includes the compressed texture image and the associated UV mapping data.
\begin{itemize}
    \setlength{\itemsep}{0.15em}
    \setlength{\topsep}{0.25em}
    \setlength{\parsep}{0pt}
    \item \textbf{UV-Source.}
    We compress the source texture using ASTC~\cite{nystad2012adaptive}, a fixed-rate GPU texture format, or NTC~\cite{ntc2023}, a learned texture codec supporting random access. For NTC, we use NVIDIA's RTXNTC SDK~\cite{RTXNTCSoftware}.

    \item \textbf{UV-Reparam.}
    We apply the same two codecs to the reparameterized atlas.

    \item \textbf{TexF.}
    TexF is compressed directly in 3D using 3DNTC (\textit{TexF/3DNTC}). Its resident payload comprises the quantized latent tables and decoder parameters.

    \item \textbf{Serialized.} We reuse the refined TexF inputs of 3DNTC, pack their attributes into 2D images using Morton or Hilbert curves, and compress them with ASTC. 
    Following the addressing strategy of prior voxel-texturing work~\cite{dolonius2020uvfree,dolonius2019compressing}, we use a compact DAG to locate attributes at queried positions. The resident payload includes both the ASTC blocks and the DAG.
\end{itemize}

Detailed codec settings and operating points are listed in the \textit{Supplementary Material}.

\begin{table*}[t]
\centering
\caption{\textbf{GPU-resident compression.} BD-BR is computed against \textit{UV-Source/ASTC}; negative values indicate memory savings. Frame time includes decoding and rendering.}
\label{tab:memory_compression}
\footnotesize
\setlength{\tabcolsep}{2.5pt}
\renewcommand{\arraystretch}{0.98}
\resizebox{\textwidth}{!}{%
\begin{tabular}{@{}l@{\hspace{4pt}}cccc@{\hspace{5pt}}c@{\hspace{3pt}}c@{\hspace{8pt}}cccc@{\hspace{5pt}}c@{\hspace{3pt}}c@{}}
\toprule
\multirow{3}{*}{\textbf{Method}}
& \multicolumn{6}{c}{\textbf{MPEG Dataset}}
& \multicolumn{6}{c}{\textbf{AOM Dataset}} \\
\cmidrule(lr){2-7}\cmidrule(l){8-13}
& \multicolumn{4}{c}{\textbf{BD-BR (\%) $\downarrow$}} & \multicolumn{2}{c}{\textbf{Time}}
& \multicolumn{4}{c}{\textbf{BD-BR (\%) $\downarrow$}} & \multicolumn{2}{c}{\textbf{Time}} \\
\cmidrule(lr){2-5}\cmidrule(lr){6-7}\cmidrule(lr){8-11}\cmidrule(l){12-13}
& \textbf{PSNR} & \textbf{SSIM} & \textbf{LPIPS} & \textbf{3D-PSNR} & \textbf{Enc. (s)} & \textbf{Frame (ms)}
& \textbf{PSNR} & \textbf{SSIM} & \textbf{LPIPS} & \textbf{3D-PSNR} & \textbf{Enc. (s)} & \textbf{Frame (ms)} \\
\midrule
\textbf{UV-Source}/ASTC
& \multicolumn{4}{c}{\textit{Anchor}} & 3.5 & --
& \multicolumn{4}{c}{\textit{Anchor}} & 4.5 & -- \\
\textbf{UV-Source}/NTC
& \ranksecond{-6.8} & \ranksecond{-3.9} & \ranksecond{2.5} & \ranksecond{-6.6} & 34.4 & 8.9
& \ranksecond{-4.1} & \ranksecond{-4.4} & \ranksecond{-2.2} & \ranksecond{-3.9} & 36.3 & 21.3 \\
\cmidrule(lr){1-13}
\textbf{UV-Reparam}/ASTC
& 15.2 & 8.7 & 65.5 & 18.0 & 3.7 & --
& 12.2 & 29.7 & \rankthird{50.5} & 18.8 & 3.4 & -- \\
\textbf{UV-Reparam}/NTC
& \rankthird{2.9} & \rankthird{-1.5} & \rankthird{56.7} & \rankthird{5.7} & 34.0 & 8.9
& \rankthird{9.2} & \rankthird{28.8} & 51.8 & \rankthird{15.4} & 36.4 & 20.4 \\
\cmidrule(lr){1-13}
\textbf{Serialized-Morton}/ASTC
& 223.0 & 254.1 & 290.2 & 231.1 & 6.0$^\ddagger$ & --
& 589.9 & 651.4 & 650.3 & 699.7 & 14.0$^\ddagger$ & -- \\
\textbf{Serialized-Hilbert}/ASTC
& 215.7 & 244.9 & 272.0 & 223.4 & 5.9$^\ddagger$ & --
& 577.9 & 639.5 & 637.1 & 685.1 & 13.5$^\ddagger$ & -- \\
\cmidrule(lr){1-13}
\textbf{TexF}/3DNTC
& \rankfirst{-33.2} & \rankfirst{-26.8} & \rankfirst{-7.4} & \rankfirst{-30.3} & 1006 & 10.8
& \rankfirst{-31.0} & \rankfirst{-23.3} & \rankfirst{-12.0} & \rankfirst{-14.5} & 1707 & 23.1 \\
\bottomrule
\end{tabular}%
}
\vspace{0.35em}
\parbox{\textwidth}{\footnotesize
NTC frame times are measured with the RTXNTC SDK renderer~\cite{RTXNTCSoftware}.
ASTC rendering is not measured because the RTX~A6000 lacks native ASTC support.
$^\ddagger$Encoding times exclude CPU preprocessing for DAG construction and serialization.
}
\end{table*}

\begin{figure*}[t]
\centering
\includegraphics[width=0.62\linewidth]{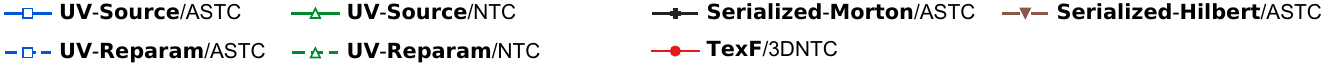}\par
\vspace{0.5em}
\includegraphics[width=0.249\linewidth]{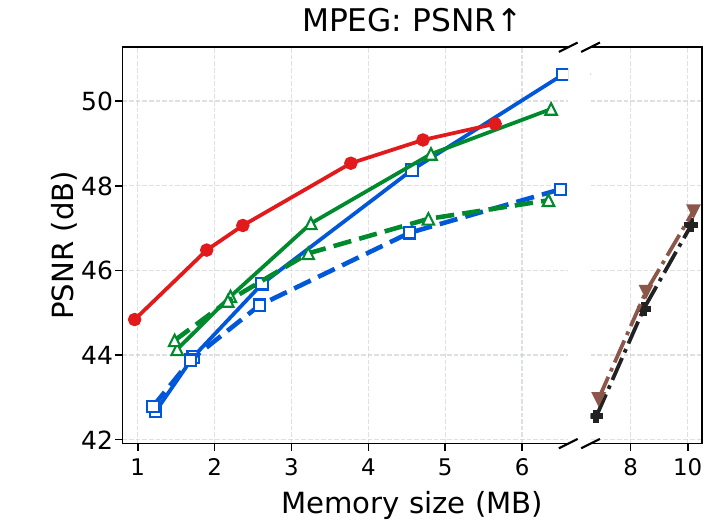}\hfill
\includegraphics[width=0.249\linewidth]{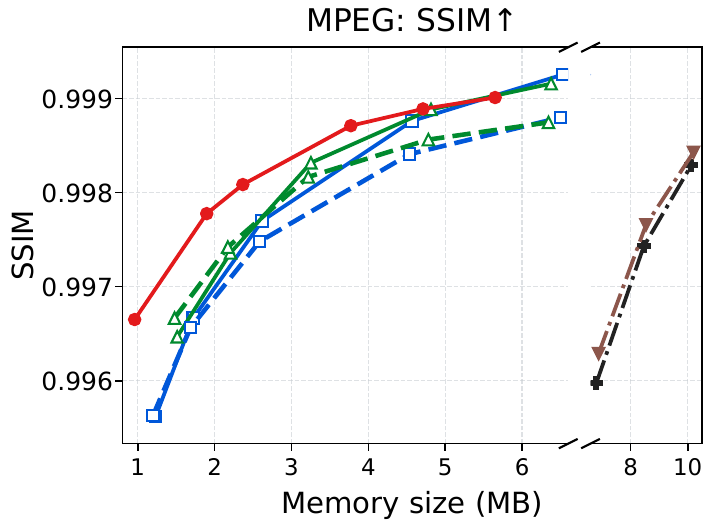}\hfill
\includegraphics[width=0.249\linewidth]{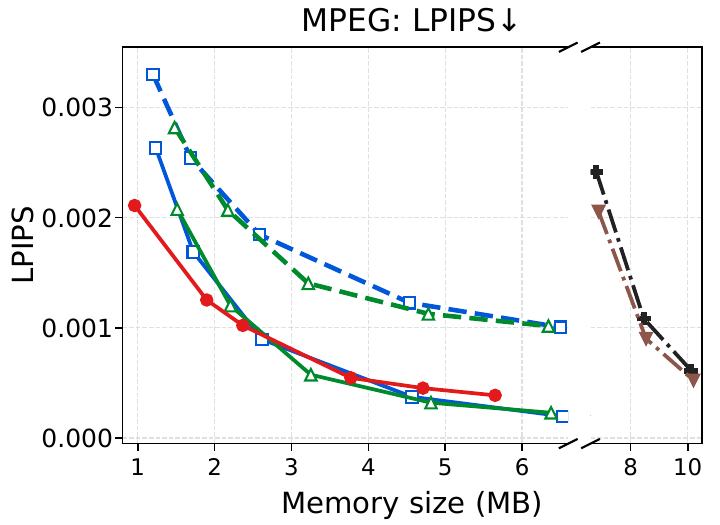}\hfill
\includegraphics[width=0.249\linewidth]{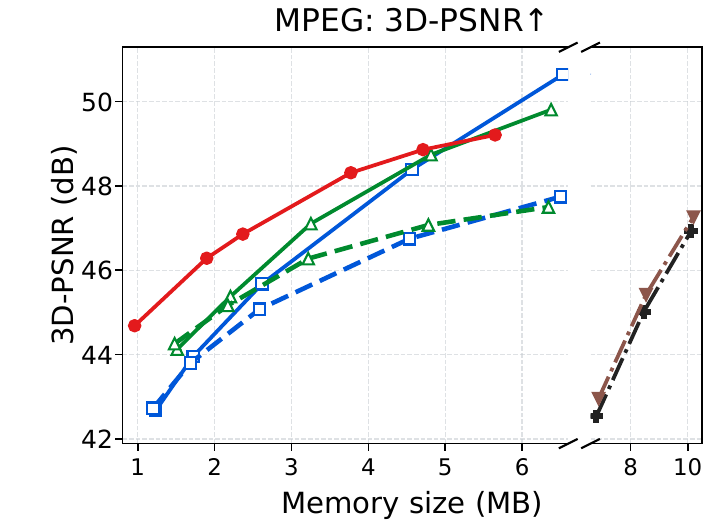}

\vspace{1.0em}

\includegraphics[width=0.249\linewidth]{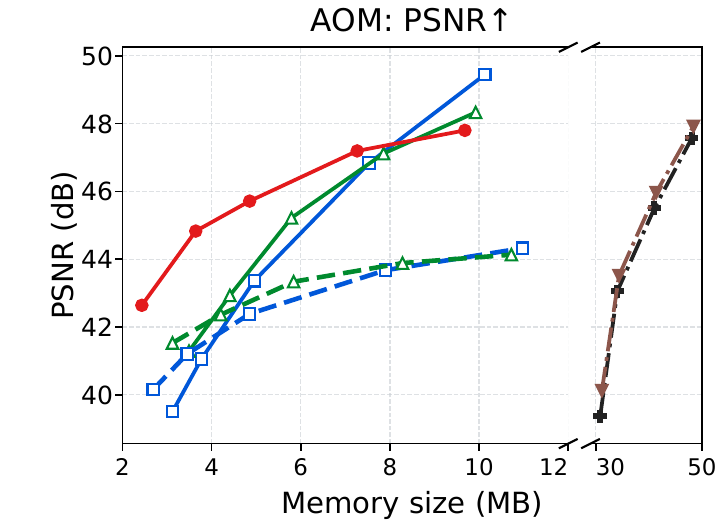}\hfill
\includegraphics[width=0.249\linewidth]{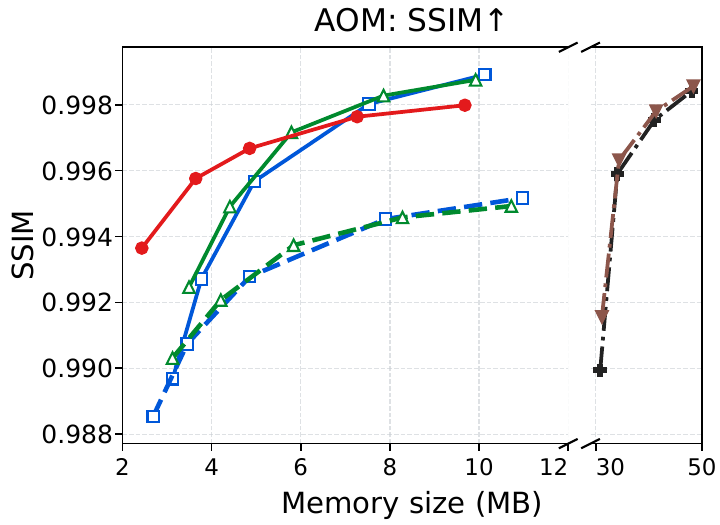}\hfill
\includegraphics[width=0.249\linewidth]{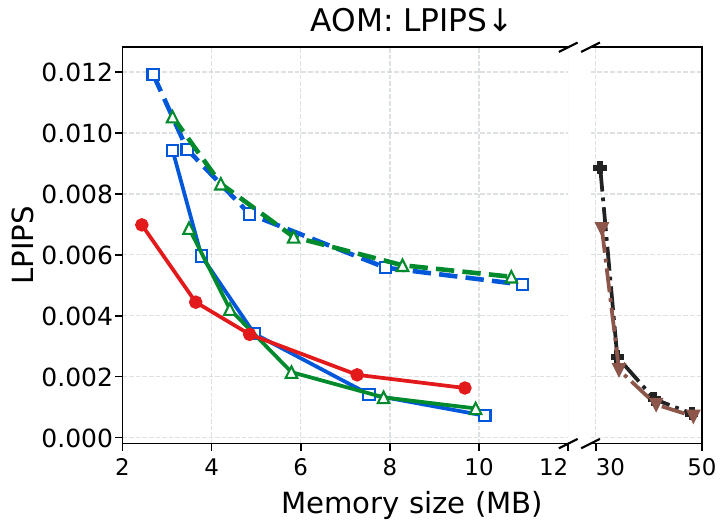}\hfill
\includegraphics[width=0.249\linewidth]{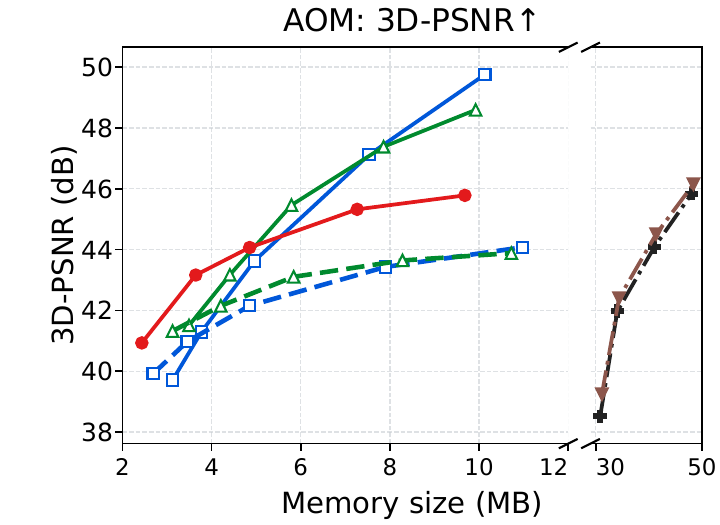}
\caption{\textbf{GPU-resident rate--distortion performance on the MPEG and AOM datasets.} Axis breaks accommodate the substantially larger memory footprints of the serialization-based methods; horizontal scales differ across the breaks.}
\label{fig:memory_rd_mb}
\label{fig:memory_serialized_psnr}
\label{fig:memory_rd_joint_comparison}
\vspace{-0.1cm}
\end{figure*}

\begin{figure*}[thbp]
\centering
\includegraphics[width=\linewidth]{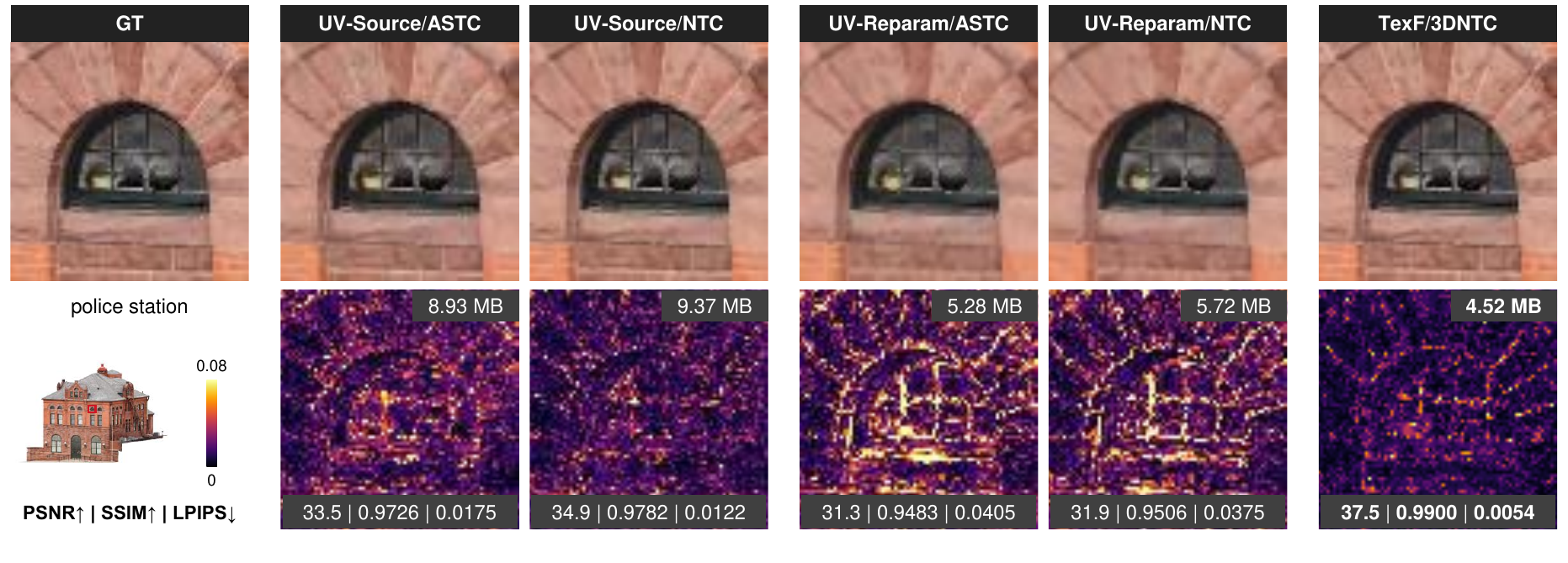}
\includegraphics[width=\linewidth]{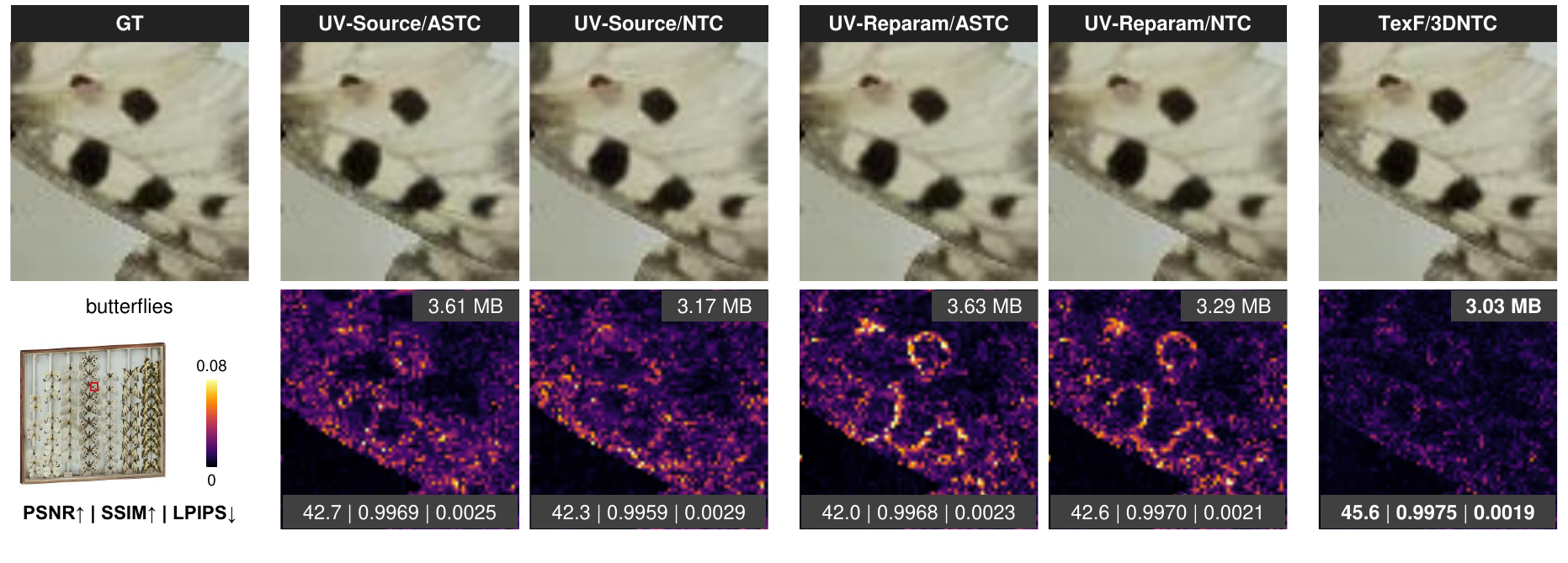}
\caption{\textbf{Visual comparison of texture reconstruction under GPU-resident compression.}}
\label{fig:memory_visual_comparison}
\end{figure*}

\subsubsection{Rate--Distortion Performance}
We compare the rate--distortion curves in Fig.~\ref{fig:memory_rd_mb} and report BD-BR relative to \textit{UV-Source/ASTC} in Table~\ref{tab:memory_compression}.
Among the UV baselines, reparameterization worsens average R--D performance across all four metrics. For example, \textit{UV-Reparam/ASTC} requires 12\%--15\% more memory than \textit{UV-Source/ASTC} at matched PSNR, although it is slightly better at the lowest-memory point. Texture transfer to a new atlas introduces resampling error, limiting fidelity at high memory budgets. Unlike in bitstream compression, improved atlas continuity does not offset this error under the evaluated random-access codecs. Replacing ASTC with NTC on the source atlas provides moderate gains, saving 6.8\% and 4.1\% memory at matched PSNR on MPEG and AOM, respectively, primarily at low memory budgets.

TexF/3DNTC instead provides clear gains in the low- and medium-memory regimes. Relative to \textit{UV-Source/ASTC}, it saves 33\% memory on MPEG and 31\% on AOM at matched PSNR. 
It also outperforms UV-Source/NTC and both UV-Reparam variants in average BD-BR across all four metrics.
Avoiding UV mapping also contributes to these savings. On AOM, the fixed UV payload accounts for 53\% of UV-Source/ASTC's resident memory at the lowest texture rate and 16\% at the highest (Fig.~\ref{fig:uv_payload_breakdown}(\subref{fig:uv_payload_astc})).
Visual comparisons in Fig.~\ref{fig:memory_visual_comparison} further illustrate these quality differences.

Among the serialized variants, Serialized-Hilbert offers a modest improvement over Serialized-Morton, but both require substantially more resident memory than 3DNTC. Serialization enables reuse of ASTC, but rearranging voxel attributes into 2D layouts can weaken the spatial correlations available to the codec.
Even without addressing overhead, both serialized ASTC variants require more memory than 3DNTC at matched quality.
Random access also requires a mapping from voxel locations to serialized attributes; even a compact DAG introduces an additional memory cost.
For Serialized-Morton/ASTC, addressing accounts for roughly 50\% of the resident memory at the highest texture rate and over 80\% at the lowest on both datasets (Fig.~\ref{fig:uv_payload_breakdown}(\subref{fig:morton_payload_astc})).
In contrast, 3DNTC uses direct hash addressing without an explicit per-voxel index. These results highlight the importance of both exploiting 3D spatial correlations and limiting addressing overhead, beyond removing UV mapping alone.

\subsubsection{Computational Efficiency}
GPU-resident compression requires low-latency, random-access decoding during rendering.
As shown in Table~\ref{tab:memory_compression}, using the same hardware and rendering resolution as above, 3DNTC averages 10.8~ms per frame on MPEG (93~FPS) and 23.1~ms on AOM (43~FPS), of which attribute decoding accounts for 6.5 and 15.8~ms, respectively. These results demonstrate real-time rendering with random-access decoding.

We measure NTC using its official renderer; differences between rendering pipelines limit direct comparisons. ASTC rendering is not measured because the RTX~A6000 lacks native ASTC support.
We also compare the per-query arithmetic cost of MLP evaluation and interpolation accumulation. 3DNTC requires an average of 8.96~KMACs on MPEG and 9.96~KMACs on AOM, close to NTC's 8.28~KMACs.
The modest increase arises from the wider decoder input and trilinear interpolation.
The comparable arithmetic cost suggests potential for further GPU acceleration.

Both methods use 100K fitting steps, following NTC's default setting~\cite{ntc2023}. Total encoding takes approximately 17~min on MPEG and 28~min on AOM for 3DNTC, versus 35~s for NTC.
NTC already benefits from custom CUDA training, which achieved an approximately $10\times$ speedup over its PyTorch reference in the original study~\cite{ntc2023}. This suggests scope for further implementation-level acceleration of 3DNTC.

\subsection{Ablation Studies and Further Analysis}\label{sec:ablation_analysis}

We conduct controlled experiments to separate the contribution of rendering-driven optimization from that of the representation itself, analyze the asset-dependent behavior of TexF and UV atlases, and evaluate TexF's compatibility with UV-based rendering pipelines.

\begin{figure}[t]
\centering
\includegraphics[width=0.325\linewidth]{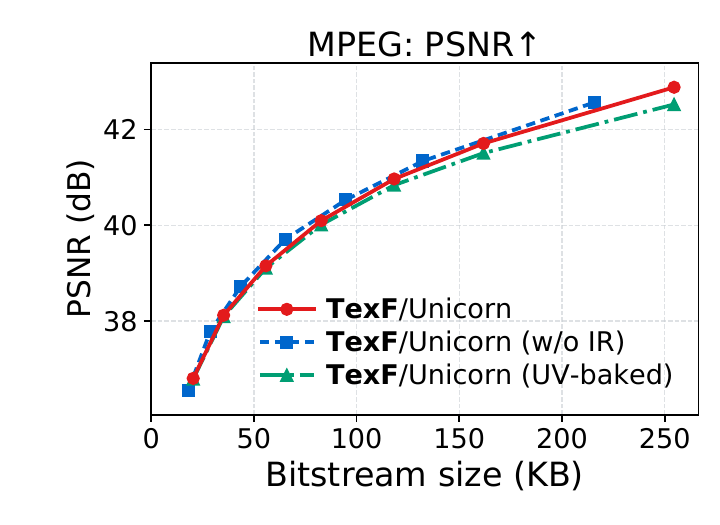}\hfill
\includegraphics[width=0.325\linewidth]{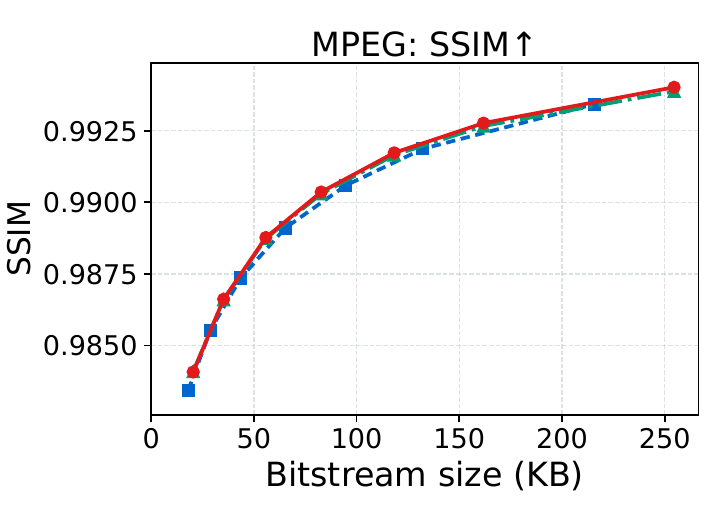}\hfill
\includegraphics[width=0.325\linewidth]{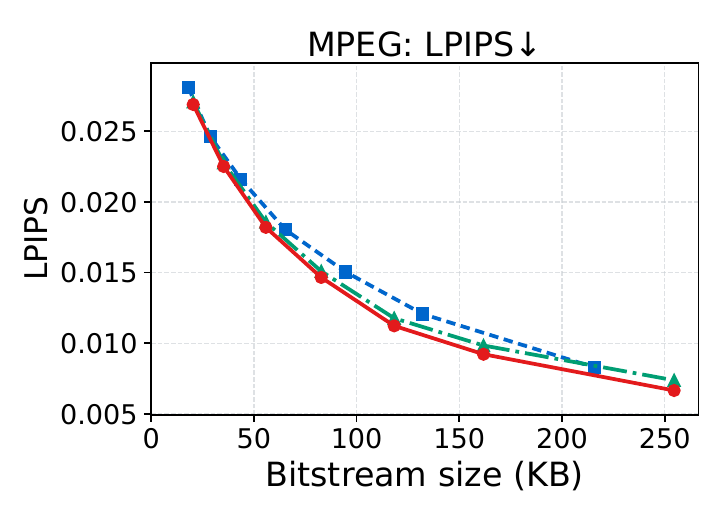}
\par\vspace{0.6em}
\includegraphics[width=0.325\linewidth]{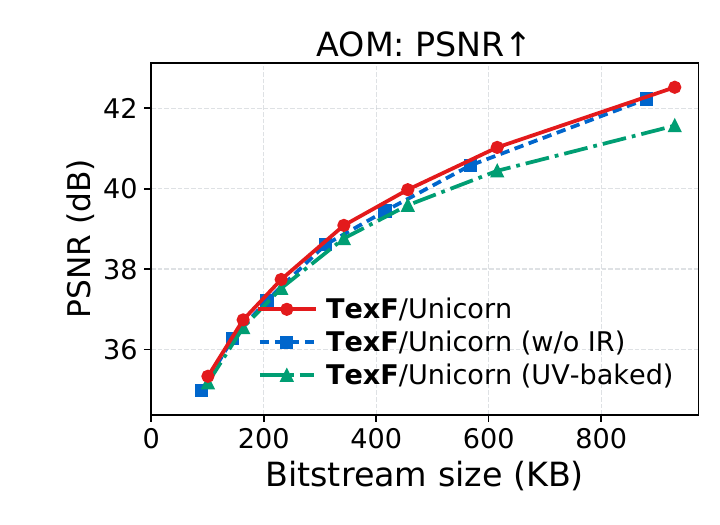}\hfill
\includegraphics[width=0.325\linewidth]{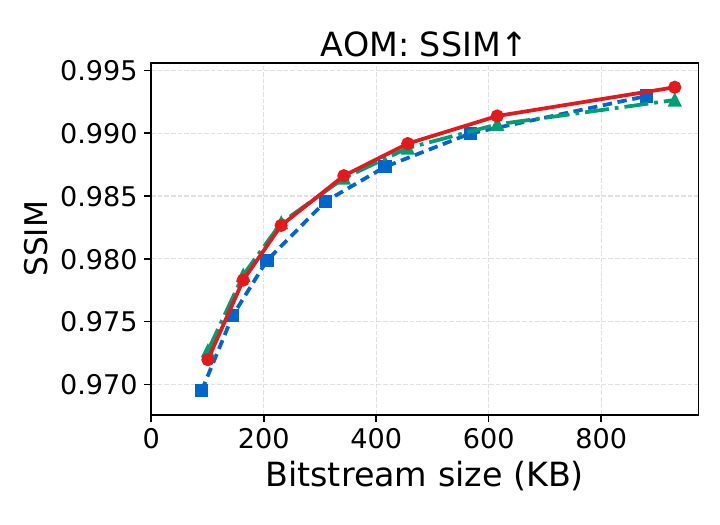}\hfill
\includegraphics[width=0.325\linewidth]{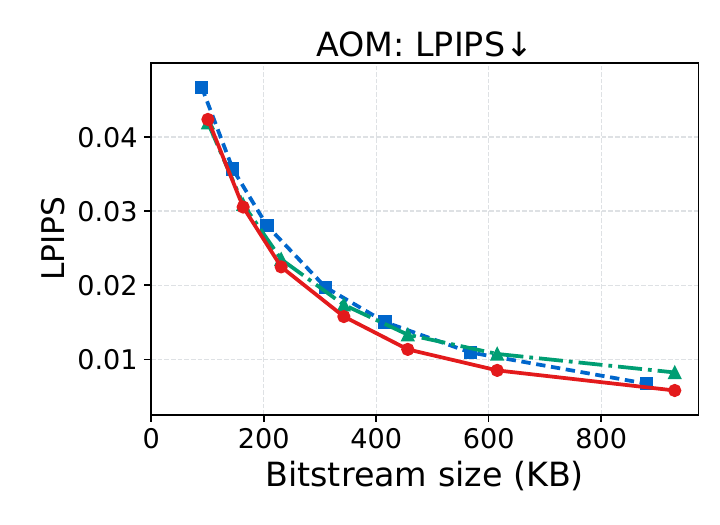}
\caption{\textbf{Bitstream compression ablations.} Comparison of \textit{TexF/Unicorn}, its variant without fitting, and its UV-baked output, using the same operating points as the main experiments.}
\label{fig:bitstream_ir_uv_bake}
\label{fig:supp_ir_ablation_rd}
\label{fig:supp_uv_baked_rd}
\end{figure}

\subsubsection{Effect of Rendering-Driven Optimization}\label{sec:ir_ablation}

Beyond the representation-level benefits shown in Sec.~\ref{sec:rep_analysis}, we examine how IR optimization affects compression performance.
As shown in Fig.~\ref{fig:bitstream_ir_uv_bake} and Table~\ref{tab:compression}, TexF/Unicorn retains its compression advantage without pre-compression IR.
IR changes the average BD-BR by less than 2 percentage points on either dataset.
IR improves the input fidelity, but preserving recovered detail may require additional bits, so its effect on bitstream R--D performance is mixed. 
By improving fidelity at a given resolution, IR can reduce the need for larger TexF representations and their associated storage and processing costs.
We use IR by default to improve input fidelity, but it can be omitted when preprocessing time is a priority.

Our UV baselines follow V-DMC's standard texture preprocessing without additional IR. An auxiliary experiment with \textit{UV-Reparam/HEVC} shows a similar trend: IR improves uncompressed texture fidelity but yields only modest coding gains.

For GPU-resident compression, IR directly refines the compressed 3DNTC field at a fixed resident size, rather than its input representation. The resulting quality improvements correspond to PSNR BD-BR savings of 3.1\% on MPEG and 10.5\% on AOM relative to the unrefined models. AOM also shows substantial improvements in SSIM and LPIPS. Detailed results are provided in the \textit{Supplementary Material}.

\begin{figure}[t]
\centering
\includegraphics[width=\linewidth]{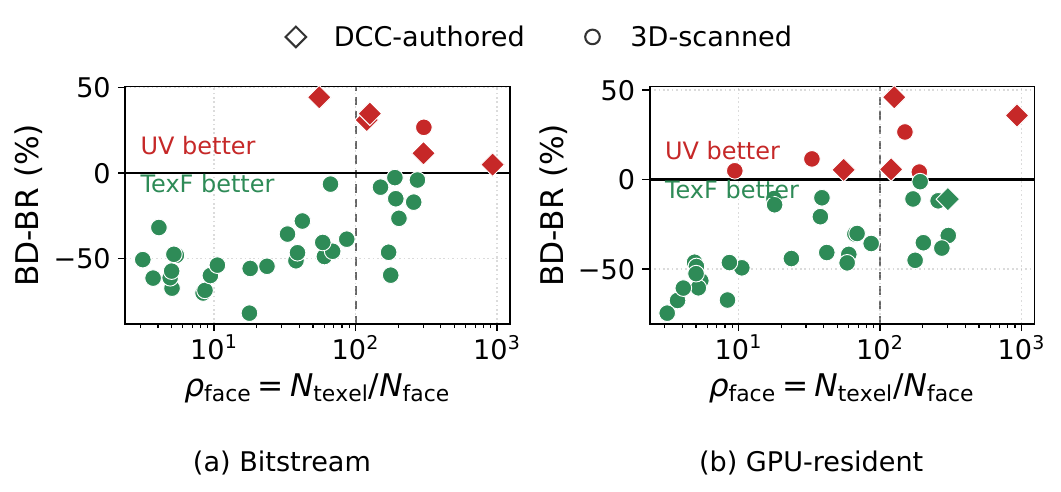}
\caption{\textbf{Per-asset compression versus texels per face.} PSNR BD-BR versus $\rho_{\mathrm{face}}=N_{\mathrm{texel}}/N_{\mathrm{face}}$: (a) \textit{TexF/Unicorn} versus \textit{UV-Reparam/HEVC}; (b) \textit{TexF/3DNTC} versus \textit{UV-Source/ASTC}. Diamonds and circles denote DCC-authored and 3D-scanned assets, respectively. Dashed vertical lines mark the empirical reference of 100 texels per face. }
\label{fig:asset_wise_analysis}
\end{figure}

\begin{figure}[thbp]
\centering
\includegraphics[width=\linewidth]{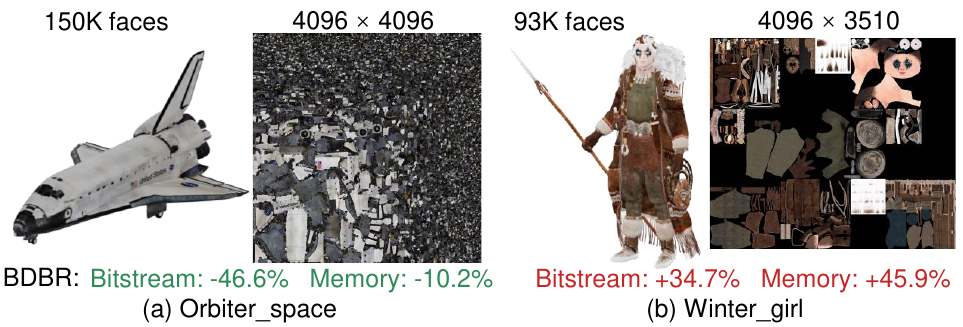}
\caption{\textbf{Representative assets with different compression behavior.} A 3D-scanned asset (left) and a DCC-authored asset (right), each shown with its textured mesh and source texture. BD-BR uses the comparisons in Fig.~\ref{fig:asset_wise_analysis}.}
\label{fig:asset_examples}
\end{figure}

\subsubsection{Per-Asset Performance}\label{sec:asset_analysis}

To understand how asset characteristics affect the relative performance of TexF and UV, we further examine per-asset compression performance across all 40 assets. We measure the average number of source texels per face as $\rho_{\mathrm{face}} = \frac{N_{\mathrm{texel}}}{N_{\mathrm{face}}}$, where $N_{\mathrm{texel}}$ is the number of source texels covered by the UV mapping and $N_{\mathrm{face}}$ is the number of faces. Due to space limitations, complete results are provided in the \textit{Supplementary Material}.
Fig.~\ref{fig:asset_wise_analysis} reports per-asset PSNR BD-BR for \textit{TexF/Unicorn} against \textit{UV-Reparam/HEVC} in bitstream compression, and \textit{TexF/3DNTC} against \textit{UV-Source/ASTC} in GPU-resident compression.
TexF achieves negative PSNR BD-BR on 32 assets for bitstream compression and 30 for GPU-resident compression.

As shown in Fig.~\ref{fig:asset_wise_analysis}, TexF generally achieves larger savings on assets with fewer texels per face, particularly below approximately 100, while UV becomes more competitive at higher ratios.
UV is particularly competitive on the five DCC-authored animation assets, whose relatively coarse meshes incur little UV mapping overhead. Conversely, TexF offers greater gains on densely tessellated scanned assets, where it avoids the larger UV mapping overhead.
Fig.~\ref{fig:asset_examples} contrasts Orbiter\_space with Winter\_girl, a failure case in which TexF requires more storage. These observations suggest that TexF and UV offer complementary strengths, motivating future compression schemes that select between them based on asset characteristics.

\subsubsection{Compatibility with UV Rendering}\label{sec:uv_compatibility}

TexF natively supports direct compression and rendering, but the decoded representation can also be baked into a conventional texture image using a UV atlas generated after decoding. This optional path transmits neither the generated UV data nor the baked texture, so the appearance bitstream remains unchanged. 

UV baking reduces the average BD-BR savings by only 1.4 percentage points across the two datasets (Table~\ref{tab:compression}), largely retaining TexF's compression advantage. The R--D curves in Fig.~\ref{fig:bitstream_ir_uv_bake} further show that the quality loss is most noticeable in PSNR at high bitrates. TexF can therefore support conventional UV rendering with limited impact on compression performance, at a one-time conversion cost of approximately 1.6~s on MPEG and 19.8~s on AOM.

\subsubsection{Extension to PBR Materials}

TexF naturally extends from RGB textures to PBR materials by expanding its attribute channels to include properties such as metallicity and roughness. 
The same framework supports PBR rendering and inverse-rendering refinement, with 3DNTC jointly compressing the material channels.

\begin{figure}[t]
\centering
\includegraphics[width=\linewidth]{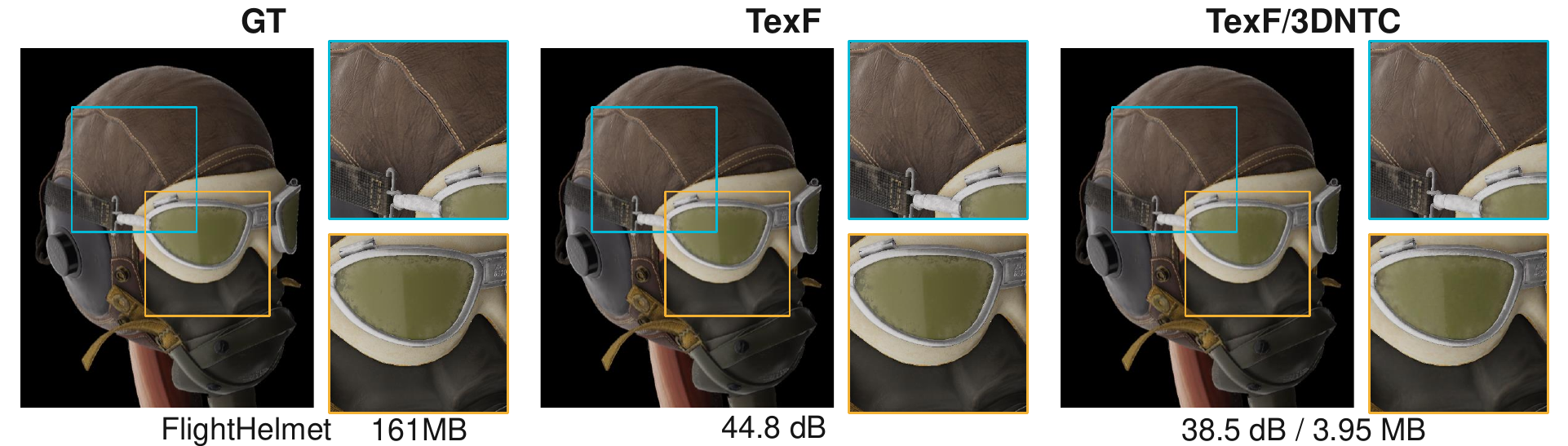}\par
\vspace{0.3em}
\includegraphics[width=\linewidth]{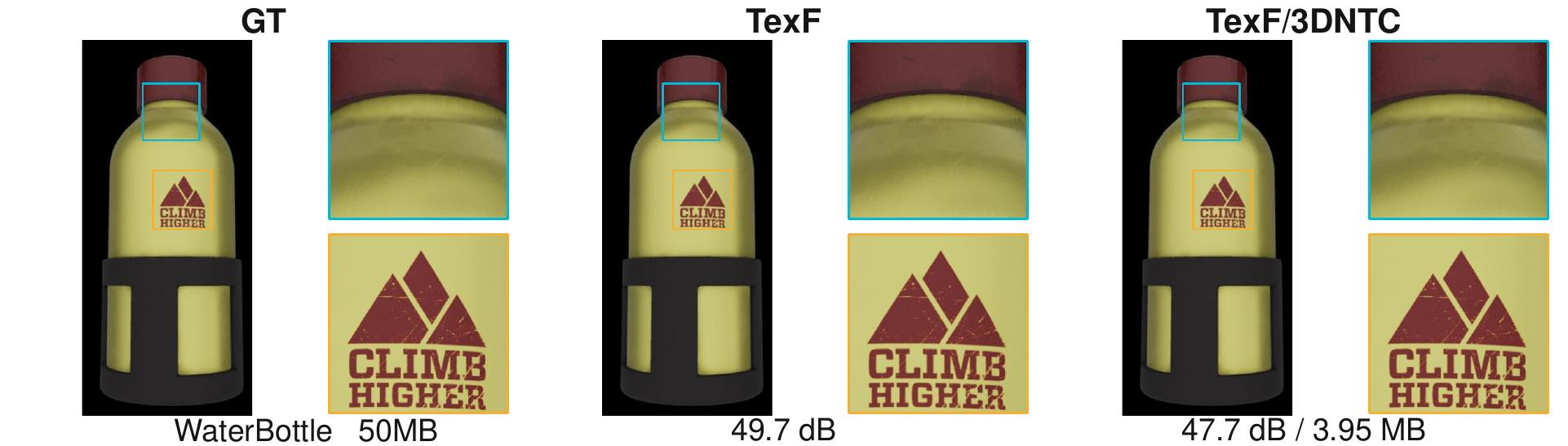}
\caption{\textbf{PBR representation and compression.}}
\label{fig:vis-PBR}
\end{figure}

Fig.~\ref{fig:vis-PBR} compares GT with uncompressed and 3DNTC-compressed TexF renderings of FlightHelmet and WaterBottle from the Khronos glTF Sample Assets. The visual and objective results show that both representations preserve material appearance, demonstrating the feasibility of UV-free PBR representation and compression. A comprehensive study of PBR compression is left for future work.

\subsection{Limitations and Discussion}
\label{sec:limitations}

\vspace{0.5em}
\noindent\textbf{Representation trade-offs.}
TexF's compression advantage depends in part on asset characteristics.
Some DCC-authored assets with coarse geometry and efficiently mapped textures favor UV coding, as shown in Sec.~\ref{sec:asset_analysis}, motivating asset-adaptive representation selection.
Additionally, at coarse voxel resolutions, closely spaced surface regions may map to the same voxel or fall within the same interpolation neighborhood, causing unintended attribute blending.
Developing texture representations and interpolation schemes that better account for local surface structure is a promising direction for mitigating such artifacts.

\vspace{0.5em}
\noindent\textbf{Computational cost and rendering support.}
Although 3DNTC supports low-latency decoding, its per-asset fitting remains costly, making the current encoder better suited to offline asset preparation.
The substantial training acceleration achieved by NTC through a custom GPU implementation~\cite{ntc2023} suggests potential for reducing this cost through similar optimizations.
Native TexF rendering currently lacks the full texture-filtering functionality of established UV pipelines.
However, its demonstrated real-time performance provides a foundation for extending native rendering capabilities.
For the bitstream pipeline, decoded TexF can also be baked into a newly generated UV atlas (Sec.~\ref{sec:uv_compatibility}), providing compatibility with existing UV rendering pipelines at a one-time conversion cost while largely retaining the compression gains.

\section{Conclusion}
\label{sec:con and dis}
We introduced TexF, a surface-aligned 3D texture representation for mesh texture compression. TexF avoids explicit UV parameterization, supports efficient construction and direct rendering, and provides a common representation for both bitstream and GPU-resident random-access compression. Experiments on the MPEG and AOM datasets demonstrate improved average rate--distortion performance in both compression settings, a favorable capacity--quality trade-off, and lower construction cost than UV reparameterization. TexF also supports real-time rendering.

Further analyses show that the compression gains persist without rendering-driven optimization and identify asset-dependent trade-offs with UV coding. Optional conversion to UV atlases provides compatibility with existing rendering pipelines. These results position TexF as a complementary 3D-native alternative within existing texture-compression workflows. Future work will explore joint geometry--texture rate allocation and optimization under a shared rendering objective.

\bibliographystyle{IEEEtran}
\bibliography{example_paper}

\clearpage

\setcounter{section}{0}
\setcounter{figure}{0}
\setcounter{table}{0}
\setcounter{equation}{0}
\setcounter{page}{1} 

\renewcommand{\thesection}{S\arabic{section}}
\renewcommand{\thefigure}{S\arabic{figure}}
\renewcommand{\thetable}{S\arabic{table}}
\renewcommand{\theequation}{S\arabic{equation}}

\twocolumn[
  \begin{center}
    \vspace*{1em}
    {\LARGE \bfseries Supplementary Material \par}
    \vspace{1em}
    \vspace{2em}
  \end{center}
]

\makeatletter
\setlength{\@fptop}{0pt}
\setlength{\@fpsep}{12pt}
\setlength{\@fpbot}{0pt plus 1fil}
\setlength{\@dblfptop}{0pt}
\setlength{\@dblfpsep}{12pt}
\setlength{\@dblfpbot}{0pt plus 1fil}
\makeatother

\section{Experimental Details}

\subsection{Bitstream Compression Settings}

All methods use the same original mesh, losslessly coded with V-DMC. We therefore compare texture-related bitstreams, including the mapping information required by each representation.

\subsubsection{UV-based Methods}
We implement the UV-based pipelines using the V-DMC reference software\footnote{\url{https://git.mpeg.expert/MPEG/3dgh/v-dmc/software/mpeg-vmesh-tm}}.
Unless otherwise specified, the coding configurations follow the V-DMC CTC~\cite{vdmc24ctc}.
We evaluate HEVC, AV1, and ELIC, using identical settings for UV-Source and UV-Reparam with each codec.
Table~\ref{tab:supp_uv_codec_settings} summarizes the UV texture codec settings for both compression modes.

    \textit{UV-Source} preserves the source UV layout and atlas size, applying V-DMC's padding to reduce boundary discontinuities before coding.
    \textit{UV-Reparam} uses V-DMC's integrated UVAtlas\footnote{\url{https://github.com/microsoft/UVAtlas}} pipeline to reparameterize the mesh and pack the charts into the default square atlas. We use the default \textsc{Quality} mode, falling back to \textsc{Fast} only for several complex assets whose parameterization exceeds one hour.
    UV payloads are compressed using V-DMC's built-in tools, with the UV-coordinate quantization parameter (QT) set to 12 for 2K assets and 13 for 4K assets, as specified by the respective CTCs~\cite{vdmc24ctc,AOM_VVM_CfP_2023}.

\begin{table}[b]
\centering
\caption{\textbf{UV texture codec settings.} UV-Source and UV-Reparam share the same settings.}
\label{tab:supp_uv_codec_settings}
\small
\setlength{\tabcolsep}{5pt}
\renewcommand{\arraystretch}{1.15}
\begin{tabular}{@{}lp{0.70\linewidth}@{}}
\toprule
Codec & Rate settings \\
\midrule
\multicolumn{2}{c}{\textit{Bitstream compression}} \\
\midrule
HEVC~\cite{HEVC}\footnotemark & QP: 45, 41, 35, 29, 25 \\
AV1~\cite{han2021av1}\footnotemark & CQ: 55, 47, 39, 31, 25 \\
ELIC~\cite{he2022elic}\footnotemark & Quality levels: 1--6 \\
\midrule
\multicolumn{2}{c}{\textit{GPU-resident compression}} \\
\midrule
ASTC~\cite{nystad2012adaptive}\footnotemark & Blocks: $12\times12$, $10\times10$, $8\times8$, $6\times6$, $5\times5$ \\
NTC~\cite{ntc2023}\footnotemark & Target bits/pixel: 1.11, 1.67, 2.50, 3.75, 5.00 \\
\bottomrule
\end{tabular}
\par\vspace{0.3em}
\parbox{\linewidth}{\footnotesize ASTC uses the \texttt{thorough} preset. ELIC uses released pretrained models without retraining; NTC uses default training settings.}
\end{table}
\footnotetext[\numexpr\value{footnote}-4\relax]{HM: \url{https://vcgit.hhi.fraunhofer.de/jvet/HM}}
\footnotetext[\numexpr\value{footnote}-3\relax]{libaom: \url{https://aomedia.googlesource.com/aom/}}
\footnotetext[\numexpr\value{footnote}-2\relax]{\url{https://github.com/VincentChandelier/ELiC-ReImplemetation}}
\footnotetext[\numexpr\value{footnote}-1\relax]{\texttt{astcenc}: \url{https://github.com/ARM-software/astc-encoder}}
\footnotetext{RTXNTC: \url{https://github.com/NVIDIA-RTX/RTXNTC}}

\subsubsection{TexF-based Methods}
We encode TexF attributes using Unicorn and G-PCC.
Table~\ref{tab:supp_texf_bitstream_rates} lists the TexF resolutions and codec rate parameters shared across assets.
Each representation is refined for 30 iterations using eight randomly sampled perspective views per iteration, totaling 240 views, at $1024\times1024$ resolution.

\begin{table}[!htbp]
\centering
\caption{\textbf{TexF bitstream compression settings.} Configurations are shared by MPEG and AOM; $1\mathrm{K}=1024$ for $\lambda$.}
\label{tab:supp_texf_bitstream_rates}
\small
\setlength{\tabcolsep}{5pt}
\begin{tabular}{lcc}
\toprule
Codec & $K$ & Rate parameter \\
\midrule
Unicorn~\cite{unicorn2025attr}\footnotemark & 1536 & $\lambda=1\mathrm{K},2\mathrm{K},4\mathrm{K}$ \\
 & 2048 & $\lambda=4\mathrm{K},8\mathrm{K}$ \\
 & 2560, 3584 & $\lambda=8\mathrm{K}$ \\
\midrule
G-PCC~\cite{MPEG-PCC-TMC13}\footnotemark & 1024 & $\mathrm{QP}=36,32,28,24$ \\
 & 1536, 2048 & $\mathrm{QP}=26$ \\
\bottomrule
\end{tabular}
\par\vspace{0.3em}
\parbox{\linewidth}{\footnotesize Unicorn uses the released RWTT-pretrained checkpoints without fine-tuning. G-PCC uses the reference attribute codec.}
\end{table}
\footnotetext[\numexpr\value{footnote}-1\relax]{\url{https://github.com/NJUVISION/Unicorn}}
\footnotetext{\url{https://github.com/MPEGGroup/mpeg-pcc-tmc13}}

\subsubsection{Serialization-based Methods}\label{sec:supp_serialized_hevc}
The \textit{Serialized-Morton/HEVC} and \textit{Serialized-Hilbert/HEVC} baselines use refined TexF representations at $K=2048$, following the same evaluation protocol as \textit{TexF/Unicorn} and \textit{TexF/G-PCC}. Attributes are ordered in 3D and packed into 2D images using the same ordering (Morton or Hilbert). We use the HM encoder at QPs 39, 35, 31, 27, and 23, with YUV 4:4:4 to avoid chroma subsampling.

\subsection{GPU-Resident Compression Settings}\label{sec:supp_memory_settings}

\subsubsection{UV-based Methods}
UV-Source and UV-Reparam use the same codec settings, with UV coordinates stored in UINT16 format.
ASTC and NTC settings are listed in Table~\ref{tab:supp_uv_codec_settings}.

\subsubsection{TexF/3DNTC}
We set $K$ to 2048 or 4096 according to the longest side of the source texture. Table~\ref{tab:supp_3dntc_rates} lists the downsampling factors, feature channels, and hash-table sizes for each operating point.

\begin{table}[!htbp]
\centering
\caption{\textbf{TexF/3DNTC compression settings.}}
\label{tab:supp_3dntc_rates}
\small
\setlength{\tabcolsep}{6pt}
\begin{tabular}{ccccc}
\toprule
Dataset & Point & $(s_1,s_2)$ & $F$ & {Table size} \\
\midrule
MPEG & 1 & $(6,12)$ & 16 & $(2^{15},2^{14})$ \\
 & 2 & $(6,12)$ & 16 & $(2^{16},2^{15})$ \\
 & 3 & $(6,12)$ & 20 & $(2^{16},2^{15})$ \\
 & 4 & $(4,8)$ & 16 & $(2^{17},2^{16})$ \\
 & 5 & $(4,8)$ & 20 & $(2^{17},2^{16})$ \\
 & 6 & $(4,8)$ & 24 & $(2^{17},2^{16})$ \\
\midrule
AOM & 1 & $(6,12)$ & 16 & $(2^{16},2^{15})$ \\
 & 2 & $(6,12)$ & 24 & $(2^{16},2^{15})$ \\
 & 3 & $(6,12)$ & 32 & $(2^{16},2^{15})$ \\
 & 4 & $(6,12)$ & 24 & $(2^{17},2^{16})$ \\
 & 5 & $(6,12)$ & 32 & $(2^{17},2^{16})$ \\
\bottomrule
\end{tabular}
\par\smallskip
\parbox{\linewidth}{\footnotesize Table sizes are specified for $2048\times2048$ textures and increase with texture area, up to fourfold.}
\end{table}

We train 3DNTC for 100k steps with an MSE loss against TexF attributes, using batch sizes of 65,536 for $K=2048$ and 262,144 for $K=4096$. Adam learning rates start at 0.1 for latents and 0.005 for the decoder and follow cosine decay to zero. 
We then refine the compressed field using the same IR settings as in the main paper, with the number of iterations increased to 100.

The per-query MAC counts reported in the main paper are computed as follows.
With MLP widths $(2F+18,64,48,32,16)$, decoding costs $C=144F+6272$ MACs per query, including latent interpolation. This averages 8.96 and 9.96~KMACs on MPEG and AOM, respectively.
Under the same accounting, NTC costs $C_{\mathrm{NTC}}=8192+8F_{\mathrm{NTC}}$ MACs per query, averaging 8.28~KMACs for $F_{\mathrm{NTC}}=(8,12,8,12,16)$.

\subsubsection{Serialization-based Methods}
The \textit{Serialized-Morton/ASTC} and \textit{Serialized-Hilbert/ASTC} baselines reuse 3DNTC's refined TexF inputs and resolutions, with the Morton/Hilbert packing described in Sec.~\ref{sec:supp_serialized_hevc}. ASTC uses the same settings as the UV baseline. 
Following the addressing strategy of Dolonius et al.~\cite{dolonius2019compressing,dolonius2020uvfree}, we use a compact counted DAG to locate voxel attributes in the serialized texture. This structure enables random access and is included in the reported memory alongside the ASTC-compressed attributes.

\section{Additional Results and Analysis}

\subsection{Effect of Post-Compression Refinement}

\begin{figure}[!htbp]
\centering
\begin{subfigure}[t]{\linewidth}
\centering
\includegraphics[width=0.325\linewidth]{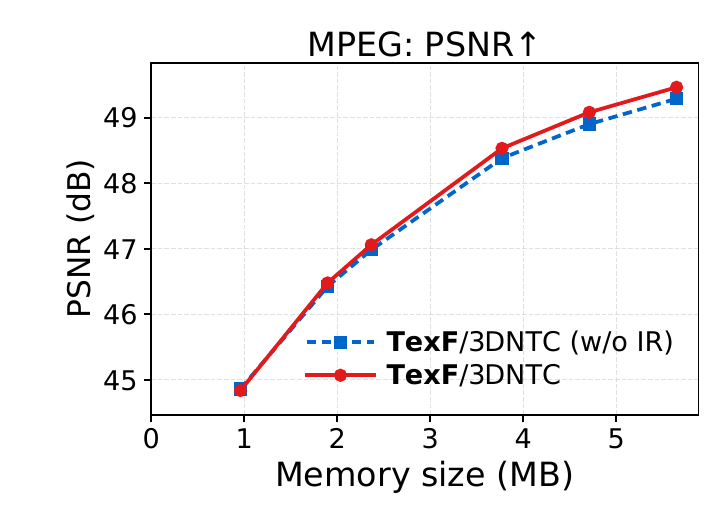}\hfill
\includegraphics[width=0.325\linewidth]{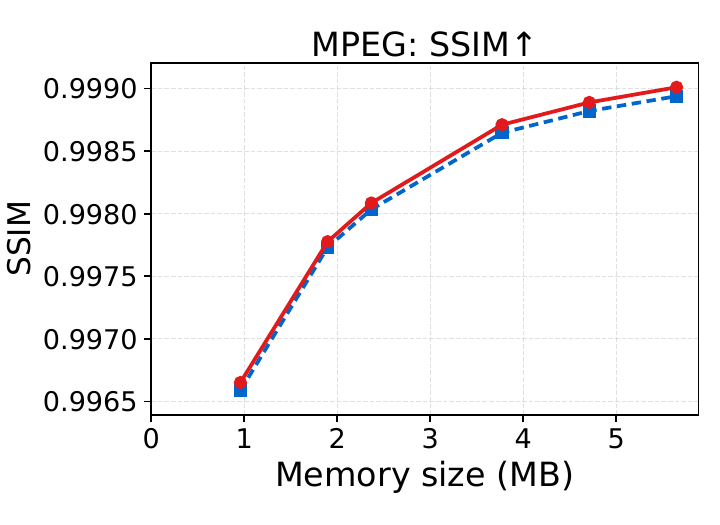}\hfill
\includegraphics[width=0.325\linewidth]{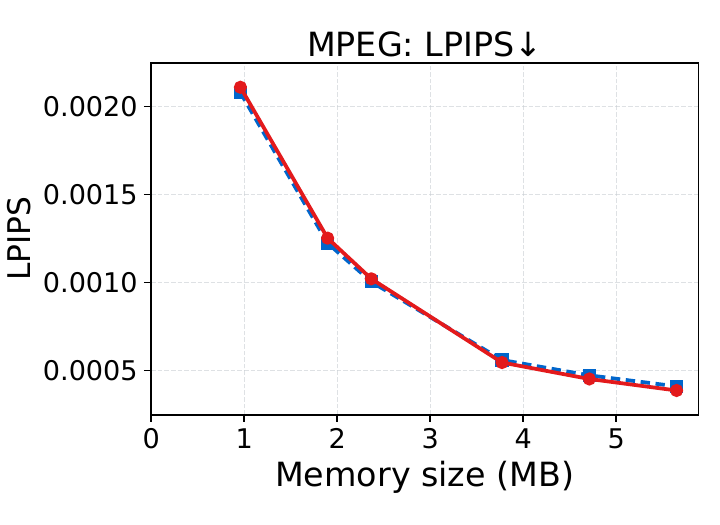}
\caption{MPEG}
\end{subfigure}
\par\vspace{0.6em}
\begin{subfigure}[t]{\linewidth}
\centering
\includegraphics[width=0.325\linewidth]{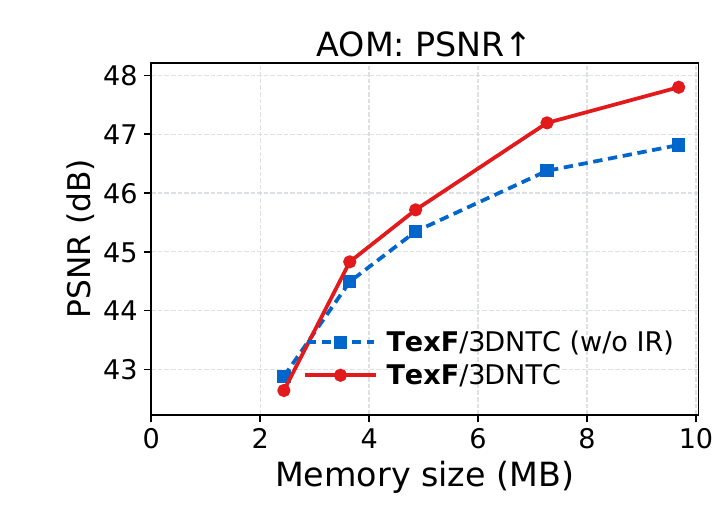}\hfill
\includegraphics[width=0.325\linewidth]{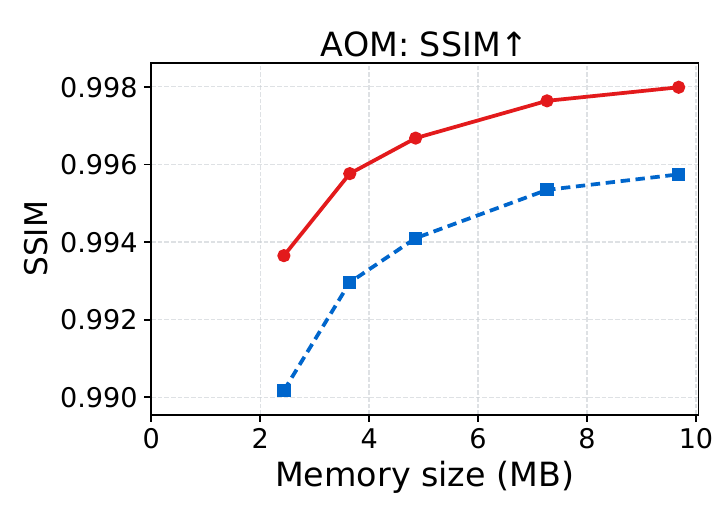}\hfill
\includegraphics[width=0.325\linewidth]{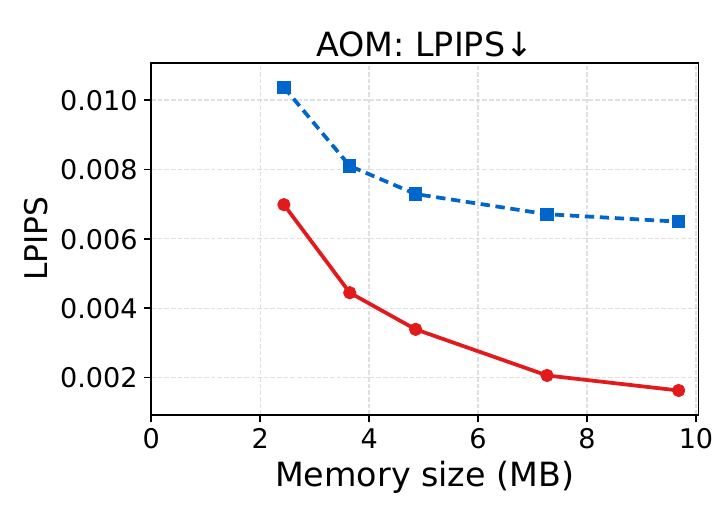}
\caption{AOM}
\end{subfigure}
\caption{\textbf{Effect of post-compression IR optimization on GPU-resident compression.} The paired TexF models use the same resident size at each operating point.}
\label{fig:supp_memory_ir_ablation_rd}
\end{figure}

Fig.~\ref{fig:supp_memory_ir_ablation_rd} compares TexF/3DNTC before and after post-compression refinement at identical resident sizes. Refinement improves rate--distortion performance, with larger gains on AOM.

\subsection{Per-Asset Compression Analysis}

\begin{figure*}[p]
    \centering
    \includegraphics[width=0.995\linewidth]{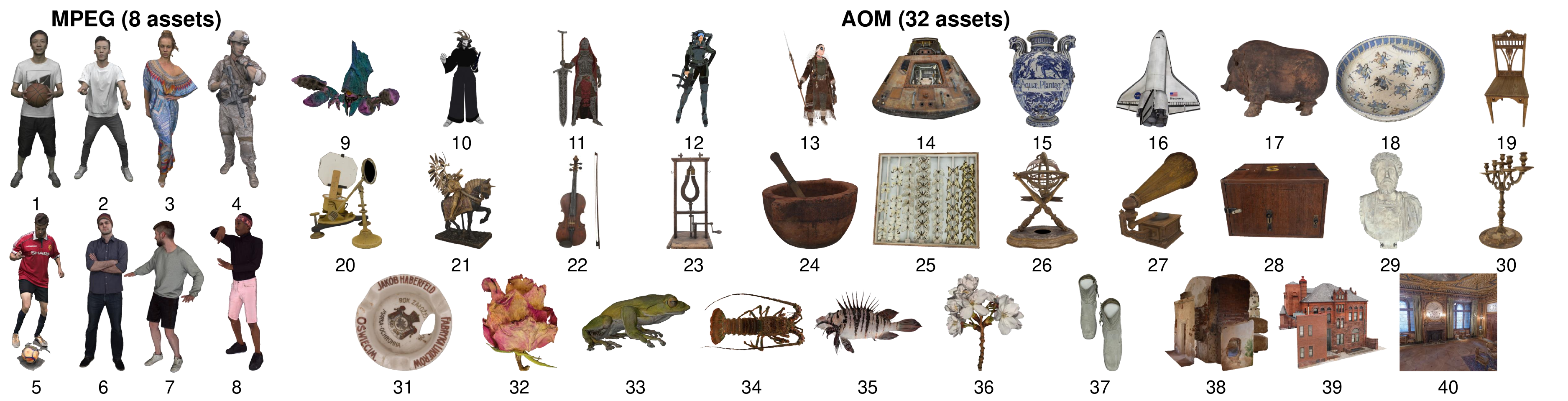}
    \caption{Complete MPEG and AOM test sets. Asset indices correspond to Table~\ref{tab:supp_per_asset_bdbr}.}
    \label{fig:dataset_full}
\end{figure*}

Figure~\ref{fig:dataset_full} shows the complete MPEG and AOM test sets.
Table~\ref{tab:supp_per_asset_bdbr} reports mesh and texture statistics, average source texels per face, and per-asset BD-BR. Bitstream results compare \textit{TexF/Unicorn} with \textit{UV-Reparam/HEVC}, while GPU-resident results compare \textit{TexF/3DNTC} with \textit{UV-Source/ASTC}.
As noted in the main paper, TexF offers compression advantages on most evaluated assets, while several DCC-authored assets favor UV coding.

\subsection{Additional Visual Comparisons}

\begin{figure*}[t]
\centering
\includegraphics[width=\linewidth]{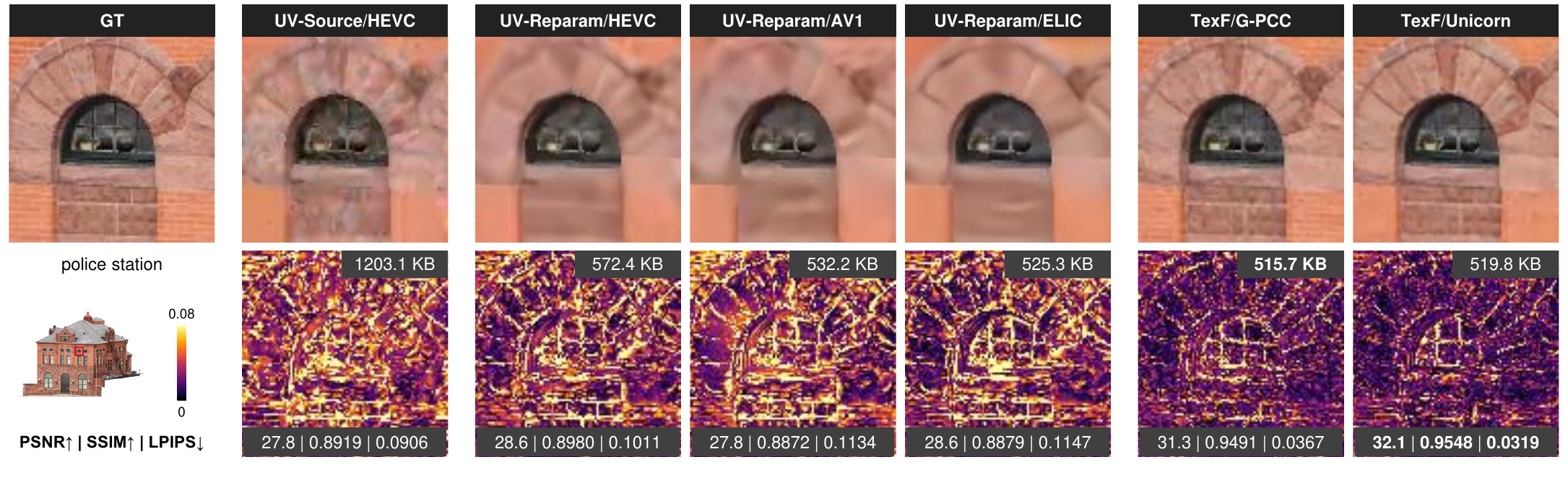}
\vspace{0.3em}
\includegraphics[width=\linewidth]
{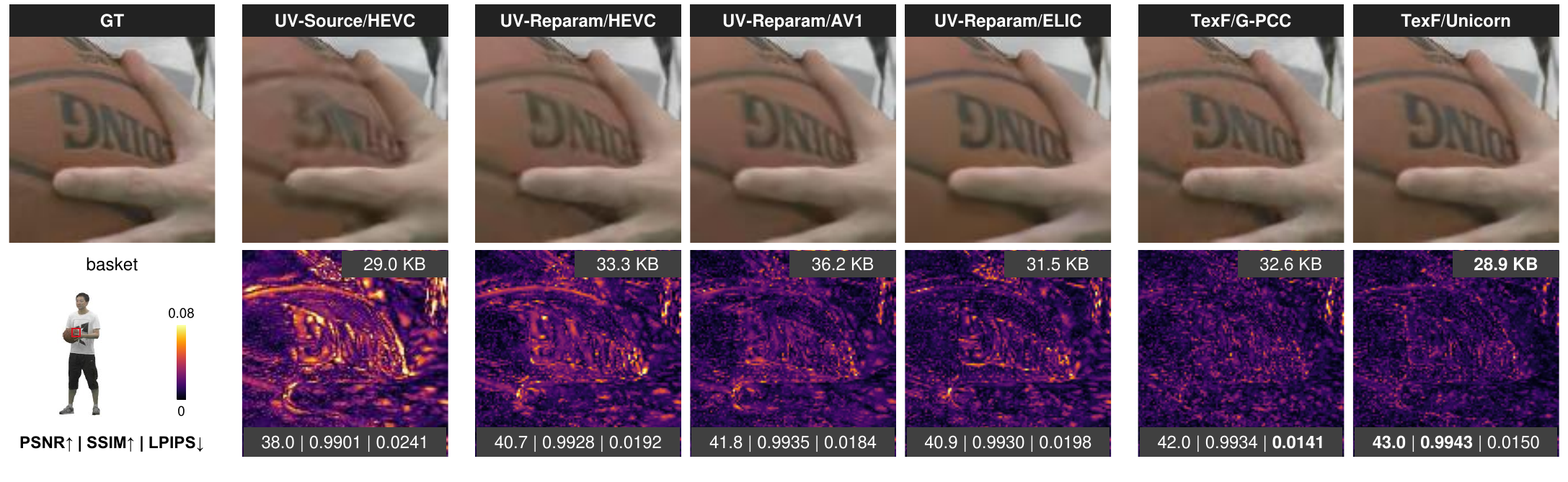}
\caption{\textbf{Additional visual comparisons under bitstream compression.}}
\label{fig:supp_bitstream_building_comparison}
\end{figure*}

\begin{figure*}[b]
\centering
\includegraphics[width=0.8\linewidth]{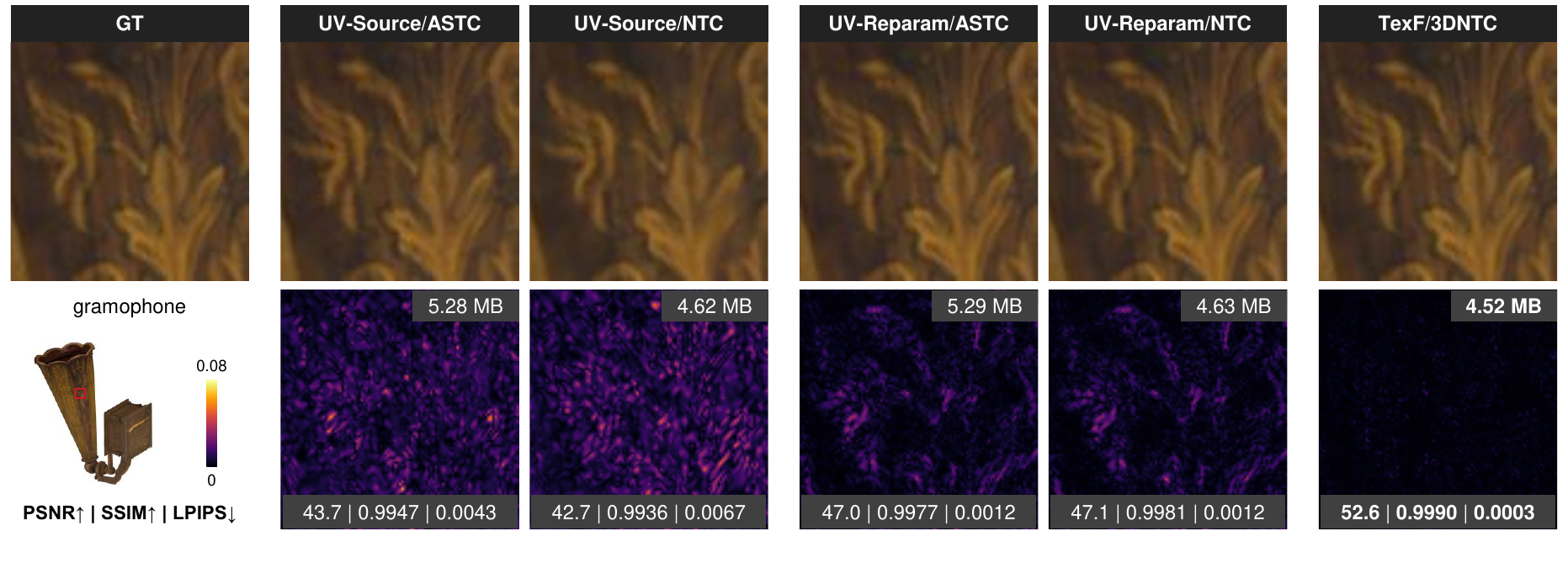}
\caption{\textbf{Additional visual comparisons under GPU-resident compression.}}
\label{fig:supp_memory_visual_comparison_1}
\end{figure*}

Figures~\ref{fig:supp_bitstream_building_comparison} and~\ref{fig:supp_memory_visual_comparison_1} provide additional reconstructions under bitstream and GPU-resident compression, respectively.

\section{Extensions}

\subsection{Texture Compression under Lossy Geometry}

The preceding experiments use lossless geometry to isolate texture compression. We further examine whether TexF retains its advantage under lossy geometry compression. Following the V-DMC CTC~\cite{vdmc24ctc}, we evaluate all eight MPEG assets and 32 AOM assets at a representative intermediate-rate operating point for geometry compression. \textit{TexF/Unicorn} and \textit{UV-Reparam/HEVC} use the same decoded mesh. Quality is evaluated against the original assets, and only texture-related bitstreams are compared, since the geometry payload is shared.

\begin{figure*}[t]
\centering
\includegraphics[width=0.249\linewidth]{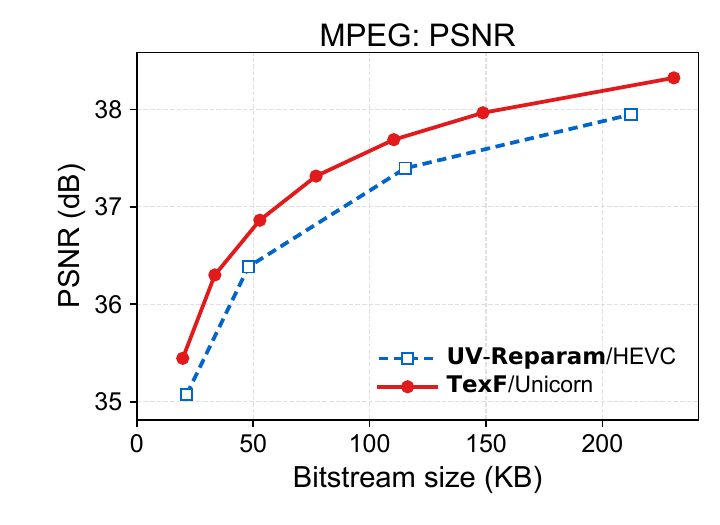}\hfill
\includegraphics[width=0.249\linewidth]{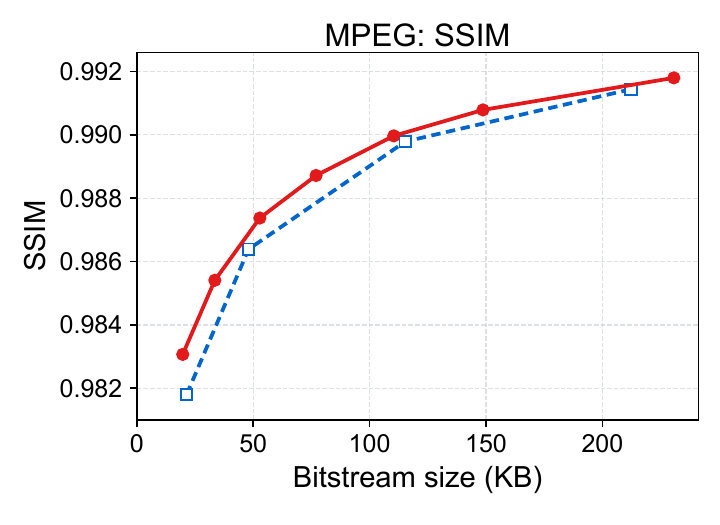}\hfill
\includegraphics[width=0.249\linewidth]{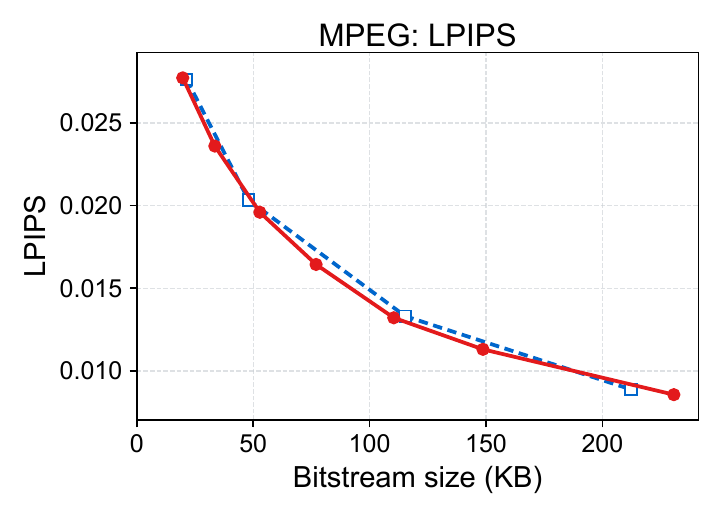}\hfill
\includegraphics[width=0.249\linewidth]{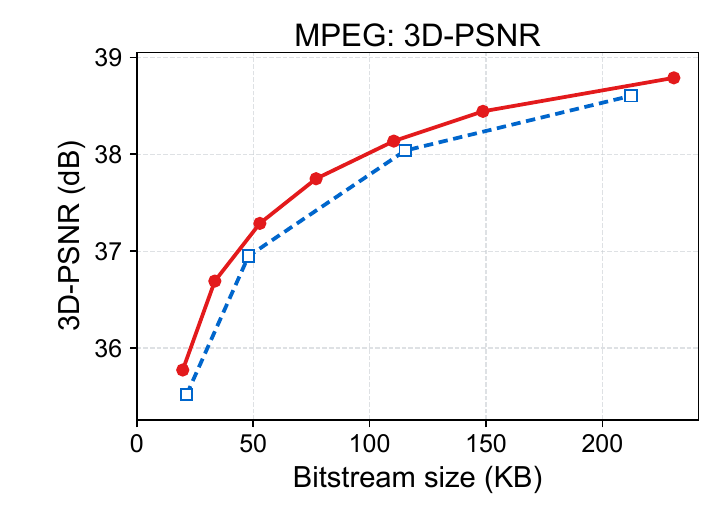}

\par\vspace{1.0em}

\includegraphics[width=0.249\linewidth]{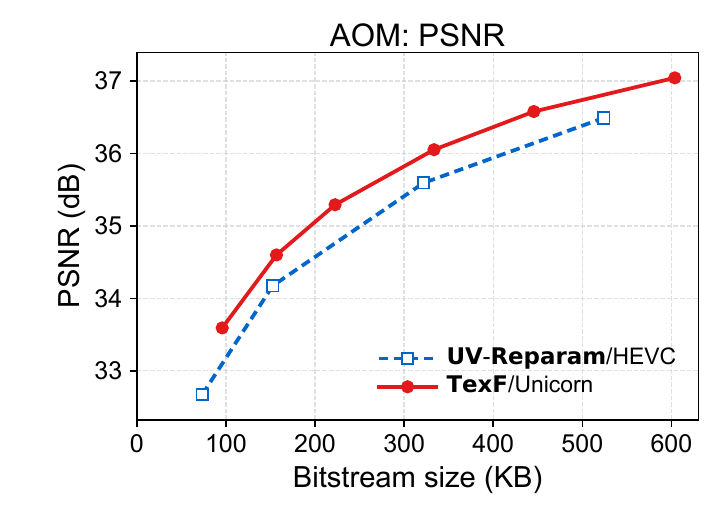}\hfill
\includegraphics[width=0.249\linewidth]{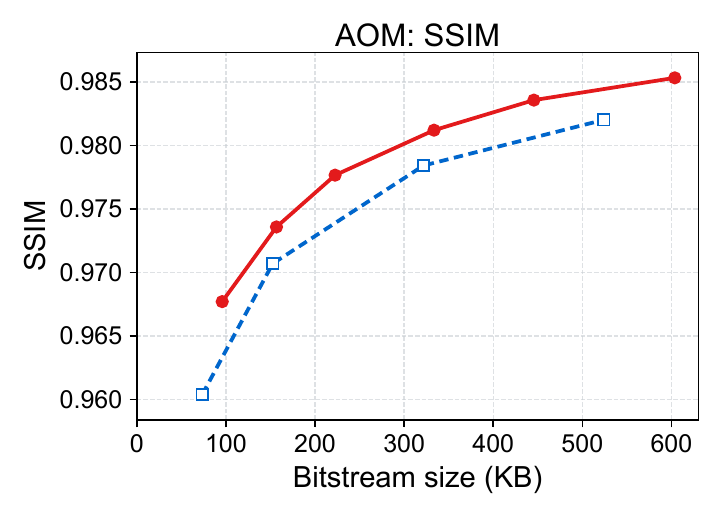}\hfill
\includegraphics[width=0.249\linewidth]{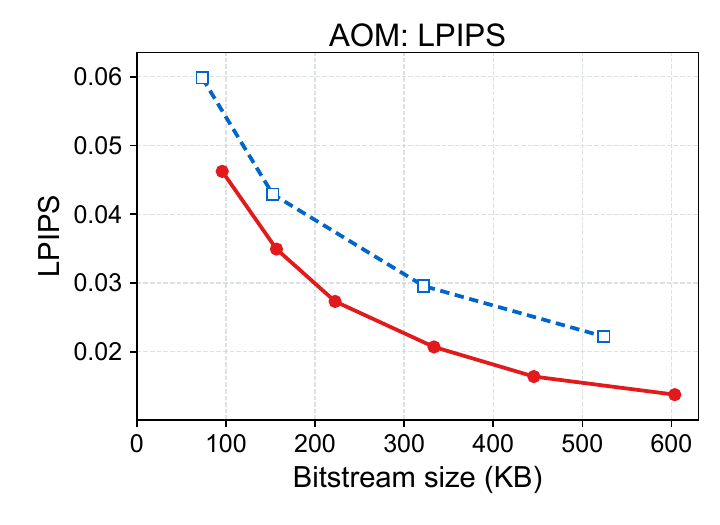}\hfill
\includegraphics[width=0.249\linewidth]{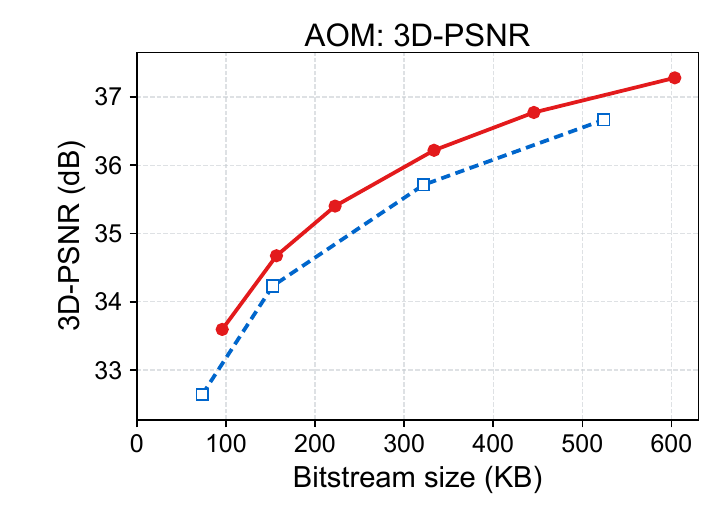}
\caption{\textbf{Texture compression under lossy geometry on MPEG and AOM}.}
\label{fig:supp_lossy_r3_rd}
\label{fig:supp_lossy_r3_rd_aom}
\end{figure*}

Fig.~\ref{fig:supp_lossy_r3_rd} and Table~\ref{tab:supp_lossy_r3_bdbr} show that TexF retains substantial bitrate savings under PSNR, SSIM, and 3D-PSNR on both datasets, while LPIPS performance is comparable to the UV baseline on MPEG and improves substantially on AOM. These results indicate that TexF's compression benefits extend to lossy geometry condition.

\begin{table}[!htbp]
\centering
\small
\setlength{\tabcolsep}{4pt}
\caption{\textbf{BD-BR (\%) under lossy geometry.}
\textit{TexF/Unicorn} is compared with \textit{UV-Reparam/HEVC} using averaged R--D curves.
Negative values indicate bitrate savings.}
\label{tab:supp_lossy_r3_bdbr}
\begin{tabular}{@{}lrrrr@{}}
\toprule
Dataset & PSNR & SSIM & LPIPS & 3D-PSNR \\
\midrule
MPEG & $-26.5$ & $-12.9$ & $-0.8$ & $-15.2$ \\
AOM & $-18.0$ & $-22.6$ & $-34.3$ & $-18.1$ \\
\bottomrule
\end{tabular}
\end{table}

\subsection{Compatibility with Implicit Geometry}
\label{sec:supp_vertex}

TexF can also represent appearance alongside implicit geometry, extending its use beyond the original mesh. We demonstrate this compatibility by coupling TexF with an SDF and rendering the extracted surface through a standard vertex-color pipeline.

\begin{figure}[!htbp]
\centering
\includegraphics[width=0.98\linewidth]{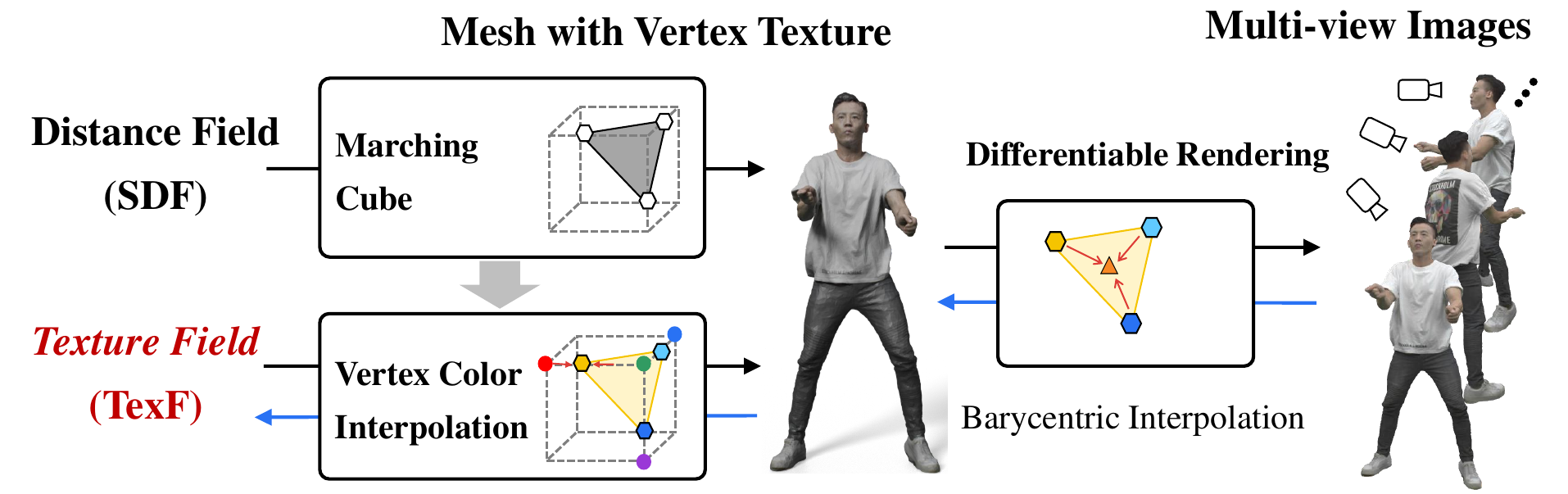}
\caption{\textbf{Compatibility with Implicit Geometry and Vertex-Based Rendering.} TexF can share a canonical grid with an implicit geometry field. After surface extraction, its attributes are interpolated onto mesh vertices and subsequently rendered through a standard vertex-color pipeline.}
\label{fig:vertex_rendering}
\end{figure}

The SDF and TexF share the same grid. After extracting a mesh with Marching Cubes~\cite{we1987marching}, we transfer TexF attributes to its vertices using $k$-nearest-neighbor interpolation. The rasterizer then interpolates these vertex attributes across triangles, avoiding per-pixel TexF queries (Fig.~\ref{fig:vertex_rendering}).

\begin{figure}[!htbp]
\centering
\includegraphics[width=0.49\linewidth]{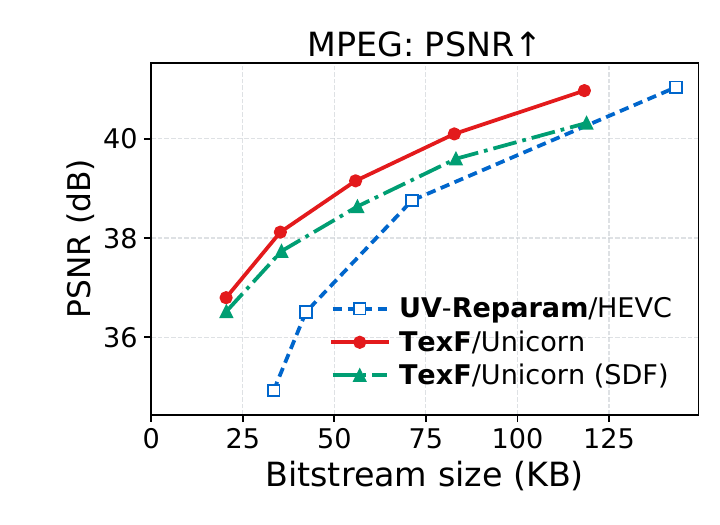}\hfill
\includegraphics[width=0.49\linewidth]{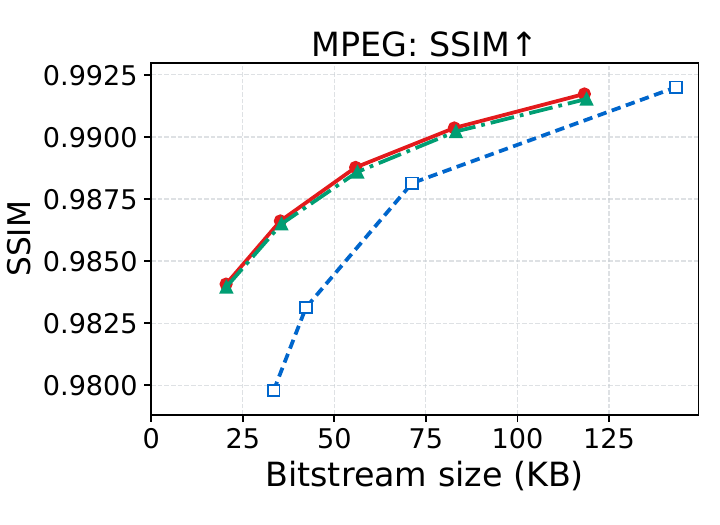}
\par\vspace{0.3em}
\includegraphics[width=0.49\linewidth]{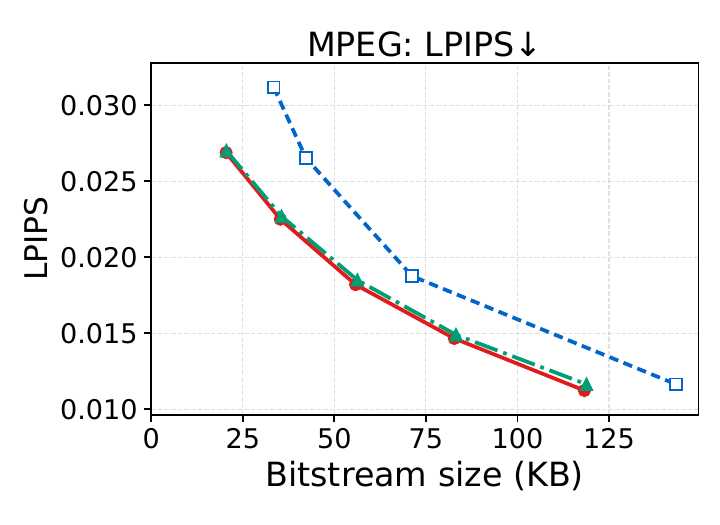}\hfill
\includegraphics[width=0.49\linewidth]{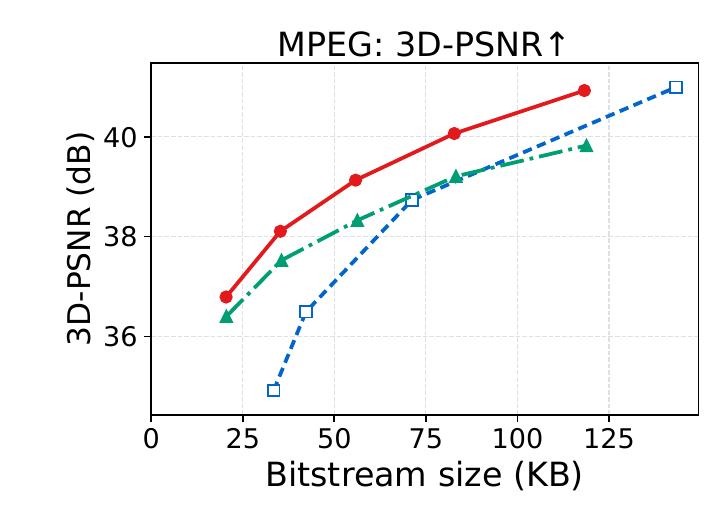}
\caption{\textbf{Texture compression with implicit geometry on MPEG.} TexF/Unicorn (SDF) uses vertex-based rendering on the SDF-derived mesh. UV-Reparam/HEVC is included for reference. Curves average eight assets; bitstream sizes exclude geometry.}
\label{fig:supp_sdf_rd}
\end{figure}

Fig.~\ref{fig:supp_sdf_rd} shows that TexF/Unicorn (SDF) retains an overall texture compression advantage over UV-Reparam/HEVC on MPEG, although its 3D-PSNR advantage diminishes at higher rates. Relative to the default TexF/Unicorn pipeline, SSIM and LPIPS remain close, whereas PSNR and 3D-PSNR decrease. Geometry and interpolation changes may explain these larger pointwise errors despite similar structural and perceptual quality. Only texture bitstreams are compared; geometry coding is outside this experiment.

This extension supports implicit geometry and vertex-based rendering at some cost in fidelity. The shared grid also suggests a direction for joint geometry and texture compression.

\clearpage

\begin{table*}[p]
\centering
\scriptsize
\setlength{\tabcolsep}{2.5pt}
\caption{\textbf{Source-asset characteristics and per-asset BD-BR.} Results cover 8 MPEG and 32 AOM assets. Green and red favor TexF and UV, respectively. $N_{\mathrm{texel}}/N_{\mathrm{face}}$ denotes the number of used texels per mesh face.
-- indicates that BD-BR is unavailable due to non-overlapping quality ranges.
}
\label{tab:supp_per_asset_bdbr}
\resizebox{\textwidth}{!}{%
\begin{tabular}{rlrrcrrrrrrrrr}
\toprule
 &  & \multicolumn{4}{c}{\textbf{Asset statistics}} & \multicolumn{4}{c}{\textbf{Bitstream BD-BR (\%)}} & \multicolumn{4}{c}{\textbf{Memory BD-BR (\%)}} \\
\cmidrule(lr){3-6}\cmidrule(lr){7-10}\cmidrule(lr){11-14}
Idx & Asset & Vertices & Faces & Texture & $\frac{\mathrm{Texels}}{\mathrm{Faces}}$ & PSNR & SSIM & LPIPS & 3D-PSNR & PSNR & SSIM & LPIPS & 3D-PSNR \\
\midrule
\multicolumn{14}{c}{\textit{MPEG}} \\
\midrule
1 & basket & 21K & 39K & 2048$\times$2048 & 60 & \textcolor{green!50!black}{-48.8} & \textcolor{green!50!black}{-49.1} & \textcolor{green!50!black}{-45.8} & \textcolor{green!50!black}{-49.7} & \textcolor{green!50!black}{-41.9} & \textcolor{green!50!black}{-16.9} & \textcolor{green!50!black}{-20.1} & \textcolor{green!50!black}{-40.5} \\
2 & dancer & 21K & 39K & 2048$\times$2048 & 59 & \textcolor{green!50!black}{-40.6} & \textcolor{green!50!black}{-42.2} & \textcolor{green!50!black}{-40.3} & \textcolor{green!50!black}{-40.6} & \textcolor{green!50!black}{-46.6} & \textcolor{green!50!black}{-19.7} & \textcolor{green!50!black}{-23.2} & \textcolor{green!50!black}{-46.2} \\
3 & longdress & 21K & 40K & 2048$\times$2048 & 66 & \textcolor{green!50!black}{-6.5} & \textcolor{green!50!black}{-16.2} & \textcolor{green!50!black}{-33.4} & \textcolor{green!50!black}{-6.4} & \textcolor{green!50!black}{-30.5} & \textcolor{green!50!black}{-35.5} & \textcolor{red}{6.4} & \textcolor{green!50!black}{-30.1} \\
4 & soldier & 23K & 40K & 2048$\times$2048 & 69 & \textcolor{green!50!black}{-45.7} & \textcolor{green!50!black}{-42.9} & \textcolor{green!50!black}{-57.6} & \textcolor{green!50!black}{-45.5} & \textcolor{green!50!black}{-30.2} & \textcolor{green!50!black}{-20.5} & \textcolor{green!50!black}{-1.4} & \textcolor{green!50!black}{-29.1} \\
5 & football & 24K & 40K & 4096$\times$4096 & 257 & \textcolor{green!50!black}{-17.0} & \textcolor{green!50!black}{-26.2} & \textcolor{green!50!black}{-24.5} & \textcolor{green!50!black}{-18.6} & \textcolor{green!50!black}{-11.9} & \textcolor{green!50!black}{-0.2} & \textcolor{green!50!black}{-5.4} & \textcolor{green!50!black}{-11.2} \\
6 & levi & 47K & 40K & 4096$\times$4096 & 177 & \textcolor{green!50!black}{-59.7} & \textcolor{green!50!black}{-63.8} & \textcolor{green!50!black}{-50.8} & \textcolor{green!50!black}{-58.5} & \textcolor{green!50!black}{-45.2} & \textcolor{green!50!black}{-38.6} & \textcolor{green!50!black}{-35.9} & \textcolor{green!50!black}{-30.5} \\
7 & mitch & 16K & 30K & 4096$\times$4096 & 304 & \textcolor{red}{26.7} & \textcolor{green!50!black}{-8.1} & \textcolor{green!50!black}{-20.6} & \textcolor{red}{25.1} & \textcolor{green!50!black}{-31.3} & \textcolor{green!50!black}{-38.6} & \textcolor{green!50!black}{-33.2} & \textcolor{green!50!black}{-30.6} \\
8 & thomas & 16K & 30K & 4096$\times$4096 & 273 & \textcolor{green!50!black}{-4.1} & \textcolor{green!50!black}{-16.9} & \textcolor{green!50!black}{-5.3} & \textcolor{green!50!black}{-5.3} & \textcolor{green!50!black}{-38.4} & \textcolor{green!50!black}{-42.7} & \textcolor{green!50!black}{-34.5} & \textcolor{green!50!black}{-36.9} \\
\midrule
\multicolumn{14}{c}{\textit{AOM}} \\
\midrule
9 & box\_squid & 15K & 20K & 4096$\times$4096 & 839 & \textcolor{red}{4.9} & \textcolor{green!50!black}{-3.9} & \textcolor{green!50!black}{-19.2} & \textcolor{red}{31.1} & \textcolor{red}{35.8} & \textcolor{red}{32.3} & \textcolor{red}{22.2} & \textcolor{red}{77.0} \\
10 & cyber\_samurai & 163K & 209K & 4096$\times$2730 & 55 & \textcolor{red}{44.2} & \textcolor{red}{103.6} & \textcolor{red}{131.2} & \textcolor{red}{--} & \textcolor{red}{5.3} & \textcolor{green!50!black}{-15.5} & \textcolor{green!50!black}{-36.5} & \textcolor{red}{--} \\
11 & grey\_knight & 23K & 36K & 4096$\times$4096 & 302 & \textcolor{red}{11.5} & \textcolor{red}{29.3} & \textcolor{red}{23.7} & \textcolor{red}{55.6} & \textcolor{green!50!black}{-11.0} & \textcolor{green!50!black}{-16.1} & \textcolor{green!50!black}{-25.6} & \textcolor{red}{32.3} \\
12 & mira\_w\_gun & 52K & 77K & 2730$\times$4096 & 120 & \textcolor{red}{31.0} & \textcolor{red}{7.6} & \textcolor{green!50!black}{-3.3} & \textcolor{red}{136.3} & \textcolor{red}{5.6} & \textcolor{red}{16.9} & \textcolor{red}{7.5} & \textcolor{red}{--} \\
13 & winter\_girl & 67K & 93K & 4096$\times$3510 & 126 & \textcolor{red}{34.7} & \textcolor{green!50!black}{-10.5} & \textcolor{green!50!black}{-23.7} & \textcolor{red}{--} & \textcolor{red}{45.9} & \textcolor{red}{18.9} & \textcolor{red}{14.6} & \textcolor{red}{--} \\
14 & apollo\_11 & 483K & 721K & 4096$\times$2730 & 8 & \textcolor{green!50!black}{--} & \textcolor{green!50!black}{--} & \textcolor{green!50!black}{--} & \textcolor{green!50!black}{--} & \textcolor{red}{--} & \textcolor{red}{43.8} & \textcolor{red}{--} & \textcolor{red}{--} \\
15 & apothecary & 31K & 60K & 4096$\times$4096 & 189 & \textcolor{green!50!black}{-2.7} & \textcolor{green!50!black}{-0.2} & \textcolor{green!50!black}{-34.5} & \textcolor{green!50!black}{-0.8} & \textcolor{red}{4.1} & \textcolor{red}{25.9} & \textcolor{red}{46.7} & \textcolor{red}{37.1} \\
16 & orbiter\_space & 129K & 150K & 4096$\times$4096 & 39 & \textcolor{green!50!black}{-46.6} & \textcolor{green!50!black}{-58.8} & \textcolor{green!50!black}{-60.4} & \textcolor{green!50!black}{-26.8} & \textcolor{green!50!black}{-10.2} & \textcolor{green!50!black}{-5.8} & \textcolor{green!50!black}{-30.7} & \textcolor{red}{20.0} \\
17 & piggy\_bank & 67K & 127K & 4096$\times$4096 & 87 & \textcolor{green!50!black}{-38.6} & \textcolor{green!50!black}{-32.9} & \textcolor{green!50!black}{-56.2} & \textcolor{green!50!black}{-39.0} & \textcolor{green!50!black}{-35.8} & \textcolor{green!50!black}{-33.5} & \textcolor{green!50!black}{-43.2} & \textcolor{green!50!black}{-35.8} \\
18 & ware\_bowl & 33K & 64K & 4096$\times$4096 & 192 & \textcolor{green!50!black}{-15.0} & \textcolor{green!50!black}{-12.6} & \textcolor{green!50!black}{-30.2} & \textcolor{green!50!black}{-14.7} & \textcolor{green!50!black}{-1.2} & \textcolor{red}{44.8} & \textcolor{red}{27.8} & \textcolor{red}{0.4} \\
19 & zakopane\_chair & 88K & 142K & 4096$\times$4096 & 42 & \textcolor{green!50!black}{-28.0} & \textcolor{green!50!black}{-27.1} & \textcolor{green!50!black}{-34.3} & \textcolor{green!50!black}{-27.8} & \textcolor{green!50!black}{-40.8} & \textcolor{green!50!black}{-47.6} & \textcolor{green!50!black}{-40.8} & \textcolor{green!50!black}{-39.6} \\
20 & heliostat & 189K & 325K & 4096$\times$4096 & 24 & \textcolor{green!50!black}{-54.5} & \textcolor{green!50!black}{-60.8} & \textcolor{green!50!black}{-54.2} & \textcolor{green!50!black}{-39.5} & \textcolor{green!50!black}{-44.2} & \textcolor{green!50!black}{-35.2} & \textcolor{green!50!black}{-45.9} & \textcolor{red}{22.9} \\
21 & hussar & 260K & 413K & 2730$\times$4096 & 11 & \textcolor{green!50!black}{-53.8} & \textcolor{green!50!black}{-51.9} & \textcolor{green!50!black}{-59.0} & \textcolor{green!50!black}{-53.3} & \textcolor{green!50!black}{-49.4} & \textcolor{green!50!black}{-51.9} & \textcolor{green!50!black}{-59.6} & \textcolor{green!50!black}{-45.5} \\
22 & violin & 261K & 460K & 4096$\times$4096 & 18 & \textcolor{green!50!black}{-81.8} & \textcolor{green!50!black}{-83.8} & \textcolor{green!50!black}{-76.9} & \textcolor{green!50!black}{-77.9} & \textcolor{green!50!black}{-10.7} & \textcolor{green!50!black}{-11.1} & \textcolor{green!50!black}{-15.5} & \textcolor{red}{81.8} \\
23 & electrodynamic & 160K & 256K & 2048$\times$4096 & 9 & \textcolor{green!50!black}{-68.5} & \textcolor{green!50!black}{-70.9} & \textcolor{green!50!black}{-68.3} & \textcolor{green!50!black}{-68.6} & \textcolor{green!50!black}{-46.5} & \textcolor{green!50!black}{-41.3} & \textcolor{green!50!black}{-51.6} & \textcolor{green!50!black}{-43.3} \\
24 & marble\_mortar & 15K & 30K & 2048$\times$4096 & 202 & \textcolor{green!50!black}{-26.5} & \textcolor{green!50!black}{-28.8} & \textcolor{green!50!black}{-45.6} & \textcolor{green!50!black}{-25.6} & \textcolor{green!50!black}{-35.4} & \textcolor{green!50!black}{-40.3} & \textcolor{green!50!black}{-45.0} & \textcolor{green!50!black}{-27.5} \\
25 & butterflies & 239K & 446K & 4096$\times$2730 & 5 & \textcolor{green!50!black}{-61.2} & \textcolor{green!50!black}{-34.8} & \textcolor{green!50!black}{-16.1} & \textcolor{green!50!black}{-39.5} & \textcolor{green!50!black}{-46.3} & \textcolor{green!50!black}{-33.0} & \textcolor{green!50!black}{-30.6} & \textcolor{green!50!black}{-33.3} \\
26 & armillary & 197K & 309K & 4096$\times$2048 & 5 & \textcolor{green!50!black}{-48.1} & \textcolor{green!50!black}{-50.3} & \textcolor{green!50!black}{-58.8} & \textcolor{green!50!black}{-38.8} & \textcolor{green!50!black}{-56.6} & \textcolor{green!50!black}{-55.0} & \textcolor{green!50!black}{-51.9} & \textcolor{green!50!black}{-39.7} \\
27 & gramophone & 328K & 643K & 4096$\times$4096 & 8 & \textcolor{green!50!black}{-70.1} & \textcolor{green!50!black}{-73.9} & \textcolor{green!50!black}{-73.2} & \textcolor{green!50!black}{-66.7} & \textcolor{green!50!black}{-67.4} & \textcolor{green!50!black}{-69.6} & \textcolor{green!50!black}{-70.2} & \textcolor{green!50!black}{-47.5} \\
28 & stereo\_cam & 178K & 318K & 4096$\times$4096 & 33 & \textcolor{green!50!black}{-35.7} & \textcolor{green!50!black}{-26.0} & \textcolor{green!50!black}{-39.7} & \textcolor{green!50!black}{-36.7} & \textcolor{red}{11.5} & \textcolor{green!50!black}{-2.5} & \textcolor{green!50!black}{-8.4} & \textcolor{red}{13.4} \\
29 & marc\_aurele & 246K & 490K & 2048$\times$2048 & 5 & \textcolor{green!50!black}{-67.3} & \textcolor{green!50!black}{-66.2} & \textcolor{green!50!black}{-67.8} & \textcolor{green!50!black}{-67.0} & \textcolor{green!50!black}{-48.6} & \textcolor{green!50!black}{-44.4} & \textcolor{green!50!black}{-50.7} & \textcolor{green!50!black}{-47.1} \\
30 & candle\_stick & 139K & 272K & 4096$\times$4096 & 38 & \textcolor{green!50!black}{-51.2} & \textcolor{green!50!black}{-52.6} & \textcolor{green!50!black}{-58.6} & \textcolor{green!50!black}{-51.0} & \textcolor{green!50!black}{-20.8} & \textcolor{green!50!black}{-9.9} & \textcolor{green!50!black}{-18.6} & \textcolor{green!50!black}{-19.1} \\
31 & promo\_ashtray & 28K & 54K & 4096$\times$4096 & 171 & \textcolor{green!50!black}{-46.3} & \textcolor{green!50!black}{-48.6} & \textcolor{green!50!black}{-47.1} & \textcolor{green!50!black}{-46.2} & \textcolor{green!50!black}{-10.9} & \textcolor{red}{41.1} & \textcolor{green!50!black}{-4.7} & \textcolor{green!50!black}{-9.8} \\
32 & dead\_rose & 43K & 50K & 4096$\times$4096 & 150 & \textcolor{green!50!black}{-8.3} & \textcolor{green!50!black}{-4.0} & \textcolor{green!50!black}{-14.8} & \textcolor{green!50!black}{-9.4} & \textcolor{red}{26.6} & \textcolor{red}{42.6} & \textcolor{red}{40.7} & \textcolor{red}{33.9} \\
33 & tree\_frog & 1122K & 1333K & 4096$\times$4096 & 4 & \textcolor{green!50!black}{-61.4} & \textcolor{green!50!black}{-65.6} & \textcolor{green!50!black}{-68.5} & \textcolor{green!50!black}{-61.7} & \textcolor{green!50!black}{-67.7} & \textcolor{green!50!black}{-70.0} & \textcolor{green!50!black}{-69.6} & \textcolor{green!50!black}{-65.0} \\
34 & lobster & 808K & 865K & 2048$\times$4096 & 3 & \textcolor{green!50!black}{-50.6} & \textcolor{green!50!black}{-55.4} & \textcolor{green!50!black}{-57.6} & \textcolor{green!50!black}{-51.5} & \textcolor{green!50!black}{-74.8} & \textcolor{green!50!black}{-77.6} & \textcolor{green!50!black}{-78.5} & \textcolor{green!50!black}{-75.2} \\
35 & luna\_lionfish & 512K & 546K & 2048$\times$4096 & 5 & \textcolor{green!50!black}{-47.5} & \textcolor{green!50!black}{-56.9} & \textcolor{green!50!black}{-62.2} & \textcolor{green!50!black}{-46.9} & \textcolor{green!50!black}{-60.6} & \textcolor{green!50!black}{-65.0} & \textcolor{green!50!black}{-64.2} & \textcolor{green!50!black}{-60.3} \\
36 & cherry & 970K & 1068K & 4096$\times$4096 & 4 & \textcolor{green!50!black}{-31.8} & \textcolor{green!50!black}{-44.2} & \textcolor{green!50!black}{-61.3} & \textcolor{green!50!black}{-36.7} & \textcolor{green!50!black}{-60.7} & \textcolor{green!50!black}{-63.1} & \textcolor{green!50!black}{-60.2} & \textcolor{green!50!black}{-60.5} \\
37 & wiz\_boots & 160K & 300K & 2048$\times$4096 & 18 & \textcolor{green!50!black}{-55.7} & \textcolor{green!50!black}{-55.5} & \textcolor{green!50!black}{-54.1} & \textcolor{green!50!black}{-56.1} & \textcolor{green!50!black}{-14.1} & \textcolor{green!50!black}{-1.7} & \textcolor{green!50!black}{-2.2} & \textcolor{green!50!black}{-13.6} \\
38 & cela\_ruins & 369K & 708K & 4096$\times$4096 & 9 & \textcolor{green!50!black}{-59.9} & \textcolor{green!50!black}{-52.9} & \textcolor{green!50!black}{-57.6} & \textcolor{green!50!black}{-57.7} & \textcolor{red}{4.8} & \textcolor{green!50!black}{-1.0} & \textcolor{red}{10.2} & \textcolor{red}{10.7} \\
39 & police\_station & 1211K & 1539K & 4096$\times$4096 & 5 & \textcolor{green!50!black}{-57.3} & \textcolor{green!50!black}{-32.4} & \textcolor{green!50!black}{-45.4} & \textcolor{green!50!black}{-53.3} & \textcolor{green!50!black}{-52.7} & \textcolor{green!50!black}{-57.9} & \textcolor{green!50!black}{-59.0} & \textcolor{green!50!black}{-51.1} \\
40 & drawing\_room & 791K & 1000K & 4096$\times$4096 & 8 & \textcolor{green!50!black}{--} & \textcolor{green!50!black}{-76.0} & \textcolor{green!50!black}{-71.5} & \textcolor{green!50!black}{-85.2} & \textcolor{green!50!black}{--} & \textcolor{green!50!black}{-39.3} & \textcolor{green!50!black}{-52.6} & \textcolor{green!50!black}{-43.0} \\
\bottomrule
\end{tabular}

}
\end{table*}

\vspace{-0.1cm}
\clearpage

\end{document}